%% file: main.tex
\documentclass[acmlarge]{acmart}
\setcopyright{acmlicensed}
\acmJournal{IMWUT}

\usepackage{amsmath}
\usepackage{algorithm}
\usepackage{algorithmicx}
\usepackage{algpseudocode}
\usepackage{graphicx}
\usepackage{lipsum}
\usepackage{enumerate}
\usepackage{enumitem}
\usepackage{hyperref} % moved before hyperxmp as required

\usepackage{multirow, makecell}
\usepackage{nccmath}
\usepackage{ragged2e}
\usepackage{threeparttable} 
\usepackage{url}
\usepackage{xspace}
\usepackage{subcaption}
\usepackage{bbding}
\usepackage{hyperxmp}
\usepackage{array}
\usepackage{array,tabularx}
\newcolumntype{Y}{>{\centering\arraybackslash}X}                 % 居中自适应ubi
\newcolumntype{B}{>{\bfseries\centering\arraybackslash}X}        % 加粗+居中自适应
\usepackage{array,graphicx}  % array: m{} 居中; graphicx: \resizebox
\newcolumntype{M}[1]{>{\centering\arraybackslash}m{#1}}          % 固定宽度居中
\usepackage{array,tabularx} % 导言区需要
\newcolumntype{M}[1]{>{\centering\arraybackslash}m{#1}} % 固定列宽 + 水平/竖直居中

\AtBeginDocument{%
  \providecommand\BibTeX{{%
    \normalfont B\kern-0.5em{\scshape i\kern-0.25em b}\kern-0.8em\TeX}}}

\newcommand{\ie}{\emph{i.e.},\xspace}

\newcommand{\eg}{\emph{e.g.},\xspace}

\newcommand\figref[1]{Fig.~\ref{#1}}

\newcommand\tabref[1]{Tab.~\ref{#1}}

\newcommand\secref[1]{Sec.~\ref{#1}}

\newcommand{\fakeparagraph}[1]{\noindent\textbf{#1.}}

\newcommand{\sysname}{{\sf UbiEnlight}\xspace}

\ifodd 1

\else

\fi
\definecolor{mypurple}{RGB}{128,0,128}

\algrenewcommand{\algorithmiccomment}[1]{\textcolor{black}{/* #1 */}}

\begin{document}

\title{UbiEnlight: User-Perception-Aware Real-Time Low-Light Video Enhancement on Mobile Devices and Smartglasses}

% \author{Anonymous}\affiliation{%
%     \institution{Anonymous}
%     \country{Anonymous}
% }

\author{Minfan Wang}
\orcid{0009-0002-0046-4828}
\affiliation{%
  \institution{Northwestern Polytechnical University}
  % \department{School of Computer Science}
  % \city{Xi'an}
  \country{China}
}\email{minfanwang@mail.nwpu.edu.cn}

\author{Sicong Liu}\authornote{Corresponding author.}
\orcid{0000-0003-4402-1260}
% \authornote{Corresponding author: Sicong Liu, Junzhao Du.}
\affiliation{%
  \institution{Northwestern Polytechnical University}
  % \department{School of Computer Science}
  % \city{Xi'an}
  \country{China}
}\email{scliu@nwpu.edu.cn}

\author{Teng Li}
\orcid{0009-0007-9033-1944}
\affiliation{%
  \institution{Northwestern Polytechnical University}
  % \city{Xi'an}
  \country{China}
}\email{liteng4735@mail.nwpu.edu.cn}

\author{Zimu Zhou}
\orcid{0000-0002-5457-6967}
\affiliation{%
  \institution{City University of Hong Kong}
  % % \department{School of Data Science}
  % \city{Hong Kong}
  \country{China}
}\email{zimuzhou@cityu.edu.hk}

\author{Sixun He}
\orcid{0009-0003-9823-9772}
\affiliation{%
  \institution{Northwestern Polytechnical University}
  % \city{Xi'an}
  \country{China}
}\email{1735922512@qq.com}

\author{Bin Guo}
\orcid{0000-0001-6097-2467}
\affiliation{%
  \institution{Northwestern Polytechnical University}
  % \city{Xi'an}
  \country{China}
}\email{guob@nwpu.edu.cn}

\author{Junzhao Du}
\orcid{0000-0001-8105-3224}
% \authornotemark[1]
\affiliation{%
  \institution{Xidian University}
  % \city{Xi'an}
  \country{China}
}\email{dujz@xidian.edu.cn}

\author{Zhiwen Yu}
\orcid{0000-0002-9905-3238}
\affiliation{%
  \institution{Harbin Engineering University}
  % \city{Harbin}
  \country{China}
}\email{zhiwenyu@nwpu.edu.cn}

% \begin{abstract}
% Low-light video capture is now common on smartphones and wearables, yet dynamic illumination coupled with camera/object motion produces noise, blur, and temporal flicker that make videos \emph{unreliable and hard to use}, especially posing barrier for users with night blindness.
% Existing methods either enhance frames and enforce post-hoc temporal consistency, or enhance videos directly.
% Low-light and motion degradation is spatiotemporally coupled; thus, we follow video-native enhancement rather than frame-wise enhancement with post-hoc consistency, which often introduces flicker and motion artifacts.
% We present \sysname, the first diffusion-transformer-based on-device system for \emph{real-time, temporally stable} low-light \textit{video} enhancement.
% \sysname builds on two insights: 1) illumination and structure separate cleanly in the \textit{frequency} domain, enabling a \textit{spectrum-guided diffusion transformer} that injects Fourier amplitude/phase priors in latent space to correct illumination while preserving structure without optical flow; and 2) human perceptual loss tolerance varies across contexts; this enables an \textit{adaptive controller} that scales sampling depth, attention reuse, and frame caching at runtime.
% We implement \sysname end-to-end on smartphones and smart glasses.
% Compared to state-of-the-art baselines, \sysname improves temporal stability by up to 50.56\% under dynamic and extreme conditions while reducing latency by up to 99.68\% (18.8\,s$\rightarrow$60\,ms).
% \end{abstract}
\renewcommand{\shortauthors}{Wang et al.}

\begin{abstract}
Low-light video capture is now common on smartphones and wearables, yet dynamic illumination coupled with mobile/werable user motion causes noise, blur, and flicker that make videos \emph{hard to perceive and use}, especially for users with night blindness and near-eye displays.
Prior work enhances frames with post-hoc temporal constraints or optimizes videos directly; however, low-light and motion degradations are \emph{spatiotemporally coupled}, so post-hoc consistency after per-frame restoration often leaves flicker and motion artifacts.
We present \sysname, the first diffusion-transformer-based \emph{on-device} system for \emph{real-time, temporally stable} low-light \textit{video} enhancement.
\sysname builds on two insights: (1) illumination and structure separate more robustly in the \textit{frequency} domain, enabling a spectrum-guided diffusion transformer that injects Fourier amplitude/phase priors to correct illumination while anchoring structure without optical flow; and (2) user perceptual tolerance varies by context, motivating a perception-aware controller that adapts sampling depth, attention reuse, and frame caching at runtime.
We implement \sysname on smartphones and smartglasses, and evaluate it against state-of-the-art baselines.
\sysname improves temporal stability by up to 54.76\% under dynamic/extreme conditions while sustaining real-time on-device performance.
% We present UbiEnlight, an integrated on-device low-light video enhancement system that combines a spectrum-guided diffusion model with a runtime-adaptive controller for stable and efficient mobile/wearable enhancement.
\end{abstract}



% \keywords{Lowlight, mobile environments, mobile computing, ubiquitous computing.}
\setcopyright{cc}
\setcctype{by}
\acmJournal{IMWUT}
\acmYear{2026} \acmVolume{10} \acmNumber{3} \acmArticle{161}
\acmMonth{9} \acmDOI{10.1145/3832024}

\makeatletter

\let\UbiHearoOriginalAuthorsAddresses\@mkauthorsaddresses

\renewcommand{\@mkauthorsaddresses}{%
  \begingroup
    \def\@authornotemark{\space (corresponding author)}%
    \UbiHearoOriginalAuthorsAddresses
  \endgroup
  \unskip.\space
}

\makeatother

\maketitle

\input{body/introduction.tex}
\input{body/motivation}
\input{body/overview}

\input{body/model}

\input{body/system}
\input{body/experiment}
\input{body/related}
\input{body/conclud}
\begin{acks}
This work was partially supported by the National Natural Science Foundation of China (62522215, 62532009, 62472354,62272368) and Key Talent Project of Xidian University (No.QTZX24004). Zimu Zhou's research is supported by CityU APRC grant (No. 9610633).
The authors thank the anonymous reviewers for their constructive feedback that has made the work stronger.
\end{acks}

\newpage
\bibliography{acmart}
\bibliographystyle{ACM-Reference-Format}
\end{document}

%% file: body/introduction.tex
\section{INTRODUCTION}
\label{sec:introduction}

Ubiquitous visual sensing on smartphones and wearables has become a primary interface for \textit{non-expert users} to perceive, record, and interact with the physical world at scale, and is especially important for users with \textit{night blindness} (affecting roughly 1 in 4,000 worldwide), for whom low-light enhancement on phones and smart glasses can serve as a practical assistive layer beyond general-purpose capture.
Video now underpins mass-market uses spanning communication, safety monitoring, assistive perception, and immersive AR/VR (\eg Google Meet~\cite{googlemeet}, Microsoft Teams~\cite{microsoftTeams}, Google Clips~\cite{clipping}, Tesla Sentry~\cite{Tesla2017functionality}, Apple Magnifier~\cite{AppleMagnifier}, Microsoft Seeing AI~\cite{seeai2021evaluation}, Google ARCore~\cite{ARCore2018learn}, Meta~\cite{meta2022ego4d}), across a global smartphone base exceeding 4.3B owners.

Yet, video is routinely captured under uncontrolled lighting, \eg night commuting, dim indoors, or restricted-illumination venues, making \textit{low light} a pervasive challenge on mobile and wearable devices.
This challenge goes beyond aesthetics: 
% in safety-critical settings (night driving assistance, elderly care, home monitoring, accessibility), degraded visibility and temporal instability reduce usability and can break downstream perception.
Beyond visual quality, degraded visibility and temporal instability reduce usability in safety-critical scenarios such as night driving, elderly care, and home monitoring.
Our user study with 53 participants (aged 21$\sim$68, \secref{sec:pre:application}) further confirms that users perceive low-light degradation as a threat to safety, reliability, and usability, motivating a responsive and temporally stable on-device enhancement solution.

The key challenge is that real-world videos captured by non-expert users exhibit \textit{non-stationary, mixed degradations}. 
To increase brightness, mobile cameras often extend exposure time through auto-exposure and ISP adaptation, which inevitably introduces motion blur from camera shake and object motion while leaving residual noise, low contrast, and color distortion. Existing enhancement methods~\cite{lowsurvey} mainly improve brightness and denoising but overlook motion-induced structural degradation; deblurring methods~\cite{2014deblurring} assume well-lit inputs; and existing joint or cascaded approaches~\cite{chen2018Sid,Retinex-Net,KinD} remain brittle and largely rely on synthetic paired data.

Existing low-light video enhancement includes \textit{frame-centric pipelines} that add temporal terms as an post-hoc auxiliary constraint, and \textit{video-native} approaches that optimize spatiotemporal restoration end-to-end.
Our key insight is that low-light degradation and motion/illumination changes are \emph{spatiotemporally coupled}, so enforcing consistency \emph{after} per-frame restoration often leaves residual flicker and motion artifacts.
Accordingly, video-native methods have evolved from i) paired dynamic-video supervision~\cite{smoid,SDSD}, to ii) stronger spatiotemporal guidance for harder motion/lighting~\cite{DID}, and further to iii) unpaired/generative learning~\cite{zhu2024unrolled}.
In parallel, diffusion is a promising generative branch. 
Yet they have a core mismatch: image-oriented enhancers~\cite{2023diffretinex,2023exposurediffusion,2024fourierdiff,2024lightendiffusion} process frames independently and flicker, while video-generation diffusions~\cite{2022videodiffusion,2024opensora,ma2024latte} favor plausibility and can drift or hallucinate structure when repurposed.
Specifically, two gaps remain:

\noindent$\bullet$ \textbf{Challenge \#1: Long-stream stability under mixed illumination-motion shifts.}
In mobile/wearable capture, illumination can change abruptly (often into extreme low light) while user motion induces fast blur and vibration (especially under increased exposure time ).
Thus, we must suppress flicker/exposure oscillation \emph{and} prevent structural/semantic drift.
Yet most frame-centric pipelines, even with flow warping or recurrent consistency~\cite{smoid,DRVsmid,stableLLVe,LVE-S2D}, still depend on correspondence estimation, which degrades under low SNR and blur, yielding artifacts or over-smoothing.
Moreover, temporal stabilization typically adds extra passes (flow/warping/refinement), raising latency and compute, a major cost for on-device, long-stream deployment.
The bottleneck is \emph{preserving coherent illumination and structure when alignment is unreliable}, under tight budgets.

\noindent$\bullet$ \textbf{Challenge \#2: Perception-constrained, resource-aware execution.}
Mobile and wearable users are highly sensitive to \emph{delay} and \emph{temporal artifacts}, yet what they notice and value is \emph{context-dependent}.
For example, during driving/riding/running, \emph{temporal continuity} and low latency often matter more than marginal per-frame fidelity; on near-eye smartglasses, even subtle jitter or exposure drift becomes salient; in static inspection (\eg reading signs), fine detail and color fidelity dominate.
Ubiquitous enhancement therefore demands \emph{runtime} trade-offs that align \emph{compute} (latency/power/memory) with \emph{human-perceptible} gains, rather than a fixed per-frame policy.
However, most backbones run with near-constant per-frame cost~\cite{2022llFlow,2023diffretinex,exposurediffusion}, and existing adaptivity often tunes a knob without explicitly using perceptual thresholds or real-time resource feedback~\cite{adaenlight, chin2019adascale}, leading to wasted compute or unstable outputs under tight budgets.
The bottleneck is \emph{closing the loop} between perceptual constraints and runtime budgets.

To address these gaps, we present \sysname, a user-centric low-light video enhancement system with two integrated modules, enabling real-time and temporally stable enhancement on mobile and wearables.

\noindent$\bullet$ \textbf{First, spectrum-guided diffusion transformer.}
Our key idea is to \emph{anchor} diffusion-based low-light \emph{video} enhancement with frequency-domain priors that remain stable under lighting changes, avoiding brittle pixel-domain decomposition and correspondence-based alignment.
Motivated by dynamic and extreme low light where Retinex-style decomposition becomes ill-posed, we shift to the \emph{frequency domain}.
Using Fourier analysis, we treat amplitude as an illumination \textit{prior} and phase as a structural-semantic \textit{anchor}, and inject both as \textit{conditioning} tokens into a \emph{latent} diffusion transformer throughout denoising.
This converts diffusion into \emph{prior-guided} enhancement: amplitude steers exposure correction while phase constrains structure to prevent drift.
We further stabilize long streams via spectrum-gated temporal reuse, correlated noise, and phase-aware regularization, achieving temporally coherent enhancement \emph{without} optical-flow alignment.

\noindent$\bullet$ \textbf{Second, user-perception-aware dynamic controller.}
Our key idea is to \emph{redefine} diffusion video enhancement as a \emph{perception-budgeted control problem}: beyond a user’s sensitivity thresholds, extra denoising steps or attention computation yield diminishing returns, while instability (flicker/lag) is immediately noticeable and task-dependent.
We therefore prioritize \emph{perceptual stability} under strict latency/energy budgets, explicitly coupling \textit{what we compute} to \textit{what users perceive}.
Leveraging redundancy across frames, attention states, and denoising steps, we expose four coordinated knobs: keyframe caching, attention reuse, sampling-depth scheduling, and resolution scaling.
We then design a closed-loop controller that jointly selects their combination to maximize perceived quality while minimizing energy subject to a real-time latency constraint, supported by a low-overhead profiler that estimates latency/energy/memory online.

% Together, these components allocate compute to \emph{perceptually critical} content (stability over fidelity when needed), achieving real-time streaming with lower energy while avoiding user-visible jitter, flicker, and exposure drift.

\begin{figure}[t]
  \centering
  \includegraphics[width=0.97\textwidth]{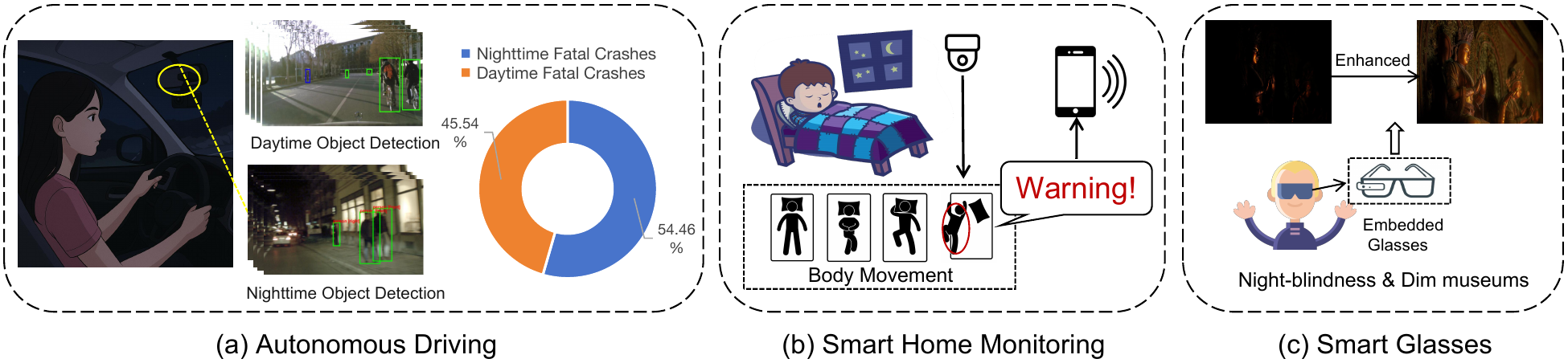}
  \caption{Representative use cases of low-light video enhancement.}
  \vspace{-3mm}
  \label{fig:usecases}
\end{figure}

\fakeparagraph{Representative Applications}
We highlight three scenarios where enhancement is \textit{functionally critical} for users rather than merely aesthetic (\figref{fig:usecases}).
\begin{itemize}
    \item \textbf{Nighttime Driving and Assistance.} 
    Headlight glare and weak dark-region cues reduce visibility at night.
    Image-based enhancement can introduce flicker and unstable object boundaries, undermining situational awareness and user trust.
    \item \textbf{In-Home Nighttime Monitoring.} 
    Low-light cameras are widely used for child and elderly monitoring.
    Brightness oscillation and contour distortion can trigger false alarms and erode long-term reliability.
    \item \textbf{Smartglasses for low-light augmentation (night-blindness and dim museums).}
    For night-blind users in particular, enhancement must be temporally stable, unobtrusive on near-eye displays, and energy-efficient for prolonged use.
    In dim museums/heritage sites with restricted lighting, users rely on mobile or wearable devices to perceive fine details.
\end{itemize}

% We implement \sysname as an end-to-end \textit{on-device} system with optimized inference, caching, and runtime control.
We implement UbiEnlight as a \textbf{joint algorithm-system design} for on-device low-light video enhancement on mobile and wearable platforms. It tightly integrates a spectrum-guided video-native enhancement model with a runtime-adaptive controller, enabling real-time and temporally stable enhancement under dynamic low-light conditions. 
We evaluate it on \textit{four} heterogeneous platforms (phones and smart glasses) and \textit{four} real-world low-light video datasets covering diverse illumination dynamics, motion patterns, and extreme low-light conditions.
Against \textit{nine} state-of-the-art baselines, \sysname cuts end-to-end latency by up to \textit{99.68\%} (\eg \textit{18.81s $\rightarrow$ 60ms}), enabling real-time operation on resource-constrained devices, while delivering \textit{stable} enhancement (improving temporal consistency by up to 54.76\%, fidelity by 44.15\%) under \textit{mixed} motion and \textit{non-stationary} illumination where prior methods flicker or amplify blur.
These gains stem from spectrum-guided conditioning and hierarchical reuse, aligning efficiency with human perception by prioritizing temporal stability and adapting computation online under dynamic and extreme lighting.
Our main contributions are summarized as follows.
\begin{itemize}
    \item To our knowledge, \sysname is the first diffusion-transformer-based real-time low-light video enhancement system designed for non-expert mobile/wearable users under dynamic illumination and mixed user motion.
    \item We introduce a \textit{spectrum-guided} latent diffusion transformer that injects frequency-domain priors (amplitude for illumination, phase for structure) for temporally coherent enhancement, and a user-perception-aware controller that adaptively allocates computation via hybrid reuse to balance quality, latency, and energy.
    \item Extensive experiments show that \sysname achieves real-time performance with improved temporal stability and perceptual quality, being robust to dynamic/extreme conditions and aligning with user-perceived usability.
    % \item \wmf{Overall, UbiEnlight is best understood as a joint algorithm-system design for on-device low-light video enhancement, rather than as only a new enhancement model or only a standalone perceptual control framework.}
\end{itemize}

%% file: body/motivation.tex
\section{BACKGROUND AND OVERVIEW}
\label{sec:pre}

%This section introduces the background and motivates the problem scope and design goals.

\subsection{Ubiquitous Applications of Low-Light Video Enhancement}
\label{sec:pre:application}
Ubiquitous cameras on smartphones and wearables have become a primary interface for users to perceive, record, and interact with the physical world. 
However, these systems often operate in \textit{uncontrolled lighting environments}, where videos are frequently captured under dim illumination (\eg nighttime commuting, dark indoor spaces, and illumination-restricted venues such as museums). 
In such settings, low-light videos suffer from severe noise, motion blur, and suppressed details, directly undermining \textit{user perception}. 
This motivates \textit{low-light video enhancement} as a key enabler for ubiquitous visual sensing applications.

We further conducted a user study with 53 participants (age 21-68), which shows that low-light degradation is not perceived merely as reduced visual quality.
Instead, users strongly associate it with tangible functional consequences, including compromised safety (84.9\%), reduced system reliability (77.4\%), and impaired day-to-day usability (90.6\%).
They also report that low-light enhancement benefits both \textit{human-facing} scenarios (\eg visual communication and media recovery) and \textit{downstream machine perception} tasks (\eg detection, tracking, and scene understanding).

\fakeparagraph{Design Goals}
Informed by these applications and user feedback, we target low-light video enhancement for mobile and wearable platforms with three user-centric objectives:
\begin{itemize}
    \item \textbf{Temporal Continuity}:
    Users are more sensitive to flicker and drift than to minor per-frame artifacts.
    \item \textbf{Real-Time Responsiveness}:
    Enhancement latency should remain imperceptible to support interactive and safety-critical usage.
    \item \textbf{Resource-Aware Sustainability}:
    Continuous enhancement must adapt to battery and resource constraints for long-term operation.
\end{itemize}

\subsection{Limitations of Existing Low-Light Solutions}
\label{sec:pre:limitation}
Despite the demand from ubiquitous applications, low-light video enhancement remains challenging on mobile and wearable platforms. 
In this work, we focus on \textbf{algorithmic enhancement} on commodity RGB cameras, without requiring specialized sensors, auxiliary illumination, or cloud offloading.
This scope is motivated by deployability and privacy: many human-facing and safety-critical scenarios require immediate feedback and local processing, and cannot assume extra hardware or reliable connectivity.

\subsubsection{Limitations of Hardware and Multi-Modal Approaches}
Hardware-based solutions (\eg low-light sensors~\cite{zhang2023cmos,liu2007CMOS}, IR/NIR modules~\cite{han2016nir,luo2010NIR}, or active illumination~\cite{li2007illumination,wei2018nighttime,sharma2021nighttime}) can improve signal quality under challenging lighting conditions, but they often incur higher energy cost and form-factor complexity, typically requiring specialized sensors or custom hardware (\tabref{diff-device}). 
These requirements hinder scalability and limit practicality for commodity ubiquitous devices such as smartphones and wearables.
Multi-modal fusion~\cite{jin2022multi,sugimura2015mult,zhang2024multdeeplight} further improves robustness by combining RGB with depth, thermal, IR, or event sensors. While effective in controlled settings (\tabref{diff-device}), such systems introduce substantial calibration overhead, parallel sensing pipelines, and additional hardware dependencies, complicating end-to-end integration. 
Consequently, they are often incompatible with always-on, and energy-constrained mobile operation, limiting their suitability for pervasive deployment in ubiquitous computing environments.

\begin{table}[t]
\centering
\tiny
\caption{Comparison of existing low-light visual sensing solutions in terms of efficiency (accuracy, energy cost, real-time performance), characteristics, sensors, and specialized hardware.}
\label{diff-device}
{%
\begin{tabular}{|c|c|c|c|c|c|c|c|}
\hline
\multirow{2}{*}{ \textbf{Category}} & \multirow{2}{*}{\textbf{Sensors}} & \multicolumn{3}{c|}{\raisebox{0pt}[4ex][2ex]{\textbf{Efficiency}}} & \multirow{2}{*}{\textbf{\begin{tabular}[c]{@{}c@{}}Suitable \\ applications\end{tabular}}} & \multicolumn{2}{c|}{\textbf{Special demands}} \\ \cline{3-5} \cline{7-8} 
 &  & \textbf{Accuracy} & \textbf{\begin{tabular}[c]{@{}c@{}}Energy cost\end{tabular}} & \textbf{\begin{tabular}[c]{@{}c@{}}Real-time \end{tabular}} &  & \textbf{Extra illuminator} & \textbf{\begin{tabular}[c]{@{}c@{}}Custom hardware\end{tabular}} \\ \hline
% \vspace{0.6cm} 
\multirow[c]{3}{*}{\textbf{\begin{tabular}[c]{@{}c@{}}\\ Special \\ imaging \\ hardware\end{tabular}}} 
& \begin{tabular}[c]{@{}c@{}}CMOS/CCD \\ Low-light sensor\end{tabular} 
& High & Medium & $\checkmark$ 
& \begin{tabular}[c]{@{}c@{}}Industrial camera, \\ General imaging \end{tabular} 
& - 
& \begin{tabular}[c]{@{}c@{}}Industrial/scientific \\ cameras\end{tabular} \\ \cline{2-8} 

& \begin{tabular}[c]{@{}c@{}}Thermal Infrared / \\ NIR camera\end{tabular} 
& Medium & High & $\checkmark$ 
& \begin{tabular}[c]{@{}c@{}} Surveillance/vehicle/UAV \end{tabular} 
& NIR illuminator & IR camera \\ \cline{2-8} 

& \begin{tabular}[c]{@{}c@{}}Radar \end{tabular} 
& High & High & \begin{tabular}[c]{@{}c@{}}$\checkmark$ \\ (low-fps)\end{tabular} 
& \begin{tabular}[c]{@{}c@{}} Automotive/robotics \end{tabular} 
& \begin{tabular}[c]{@{}c@{}}Built-in \\ laser emitter\end{tabular} 
& \begin{tabular}[c]{@{}c@{}}LiDAR (laser, \\ photodetector)\end{tabular} \\ \hline
% \vspace{0.6cm} 
\multirow[c]{3}{*}{\textbf{\begin{tabular}[c]{@{}c@{}} \\ Multi-modal \\ fusion \\ systems\end{tabular}}} 
& \begin{tabular}[c]{@{}c@{}}Event camera + RGB\end{tabular} 
& High & Low & $\times$ 
& \begin{tabular}[c]{@{}c@{}}Robotics/UAV/AR\end{tabular} 
& - & dual cameras \\ \cline{2-8} 

& \begin{tabular}[c]{@{}c@{}}RGB+NIR \\ fusion camera\end{tabular} 
& High & Medium & $\times$ 
& \begin{tabular}[c]{@{}c@{}} Industrial camera/inspection\end{tabular} 
& NIR illuminator & dual sensors \\ \cline{2-8} 

& \begin{tabular}[c]{@{}c@{}}Thermal+visible \\ fusion camera\end{tabular} 
& High & Medium-High & $\times$ 
& \begin{tabular}[c]{@{}c@{}}Automotive/security \end{tabular} 
& - & dual cameras \\ \hline
\end{tabular}%
}
\end{table}

\subsubsection{Limitations of Existing Algorithmic Enhancement}

Algorithmic methods avoid extra hardware, but current techniques still fail to meet the goals in \secref{sec:pre:application}, mainly due to two coupled limitations.
\begin{itemize}
    \item \textbf{Temporal instability under real-world dynamics.}
    Many low-light enhancers are designed for images~\cite{llnet,RRDNet,Retinex-Net,ExCNet} and applied to videos frame by frame, producing exposure jitter and flicker that are highly salient to users and disruptive to downstream perception.
    Video-specific methods rely on optical flow~\cite{sun2018pwcnet} or recurrent propagation~\cite{stableLLVe}, but these mechanisms become fragile under extremely low signal-to-noise ratios and severe motion blur.
    As a result, existing methods struggle to deliver the \textbf{temporal continuity} required by ubiquitous applications.
    \item \textbf{Prohibitive and inflexible computation on ubiquitous devices.}
    High-quality enhancement increasingly relies on heavy neural models and iterative generative pipelines.
    Diffusion-based enhancement~\cite{2023diffretinex,2024fourierdiff, wavelet-baseddiff} is promising for perceptual fidelity, yet iterative sampling and repeated attention computation incur substantial latency and energy overhead.
    Moreover, ubiquitous deployment is inherently dynamic: device resources fluctuate, and video content exhibits time-varying user/target \textit{motion} and \textit{illumination} changes.
    A fixed configuration cannot sustain a stable quality-latency-energy trade-off across scenes and devices, conflicting with the need for \textbf{real-time responsiveness} and \textbf{resource-aware sustainability}.
\end{itemize}

\fakeparagraph{Takeaway}
These limitations call for a new design that \textit{(i)} enforces temporally stable enhancement under low-light dynamics and \textit{(ii)} adapts computation to runtime resource and scene variations, enabling real-time and sustainable operation on ubiquitous devices.

%% file: body/overview.tex
\subsection{Solution Overview}
\label{sec:pre:overview}

\begin{figure*}[t]
  \centering
  \includegraphics[width=0.8\linewidth]{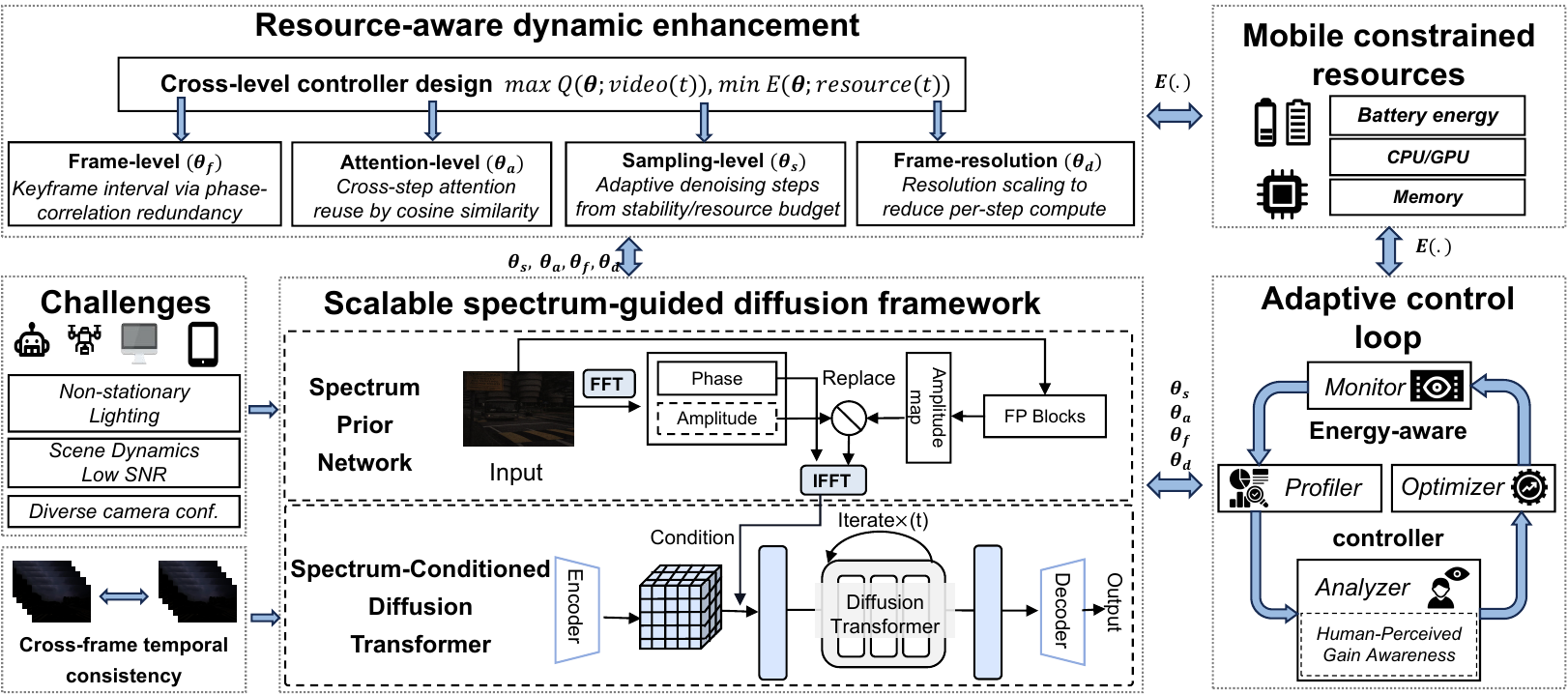}
  \vspace{-1mm}
  \caption{{Workflow of \sysname}}
  \vspace{-3mm}
  \label{fig:overview}
\end{figure*}
% zai loop limianjiayigemokuai humanperceived gain awraeness;
% intro-challenge
% 10 
To meet these requirements, we present \sysname, a low-light video enhancement system that delivers \textbf{temporally stable} enhancement with \textbf{real-time} and \textbf{resource-aware} execution on ubiquitous devices.
\sysname is guided by two insights that translate the above goals into system designs.
\begin{itemize}
    \item \textbf{Spectrum priors provide a stable anchor for illumination correction and structure preservation.}
    Pixel-space appearances are highly unstable in low light.
    This makes it difficult for image-based models to maintain consistent structure when applied to videos, and it also makes flow-based alignment unreliable.
    Instead, \sysname leverages a more robust representation in the frequency domain.
    Fourier \textit{amplitude} reflects illumination and energy distribution, while \textit{phase} encodes geometric layout and structural contours.
    Importantly, phase tends to remain stable under illumination changes, making it a natural \textbf{structural anchor} for temporally coherent enhancement.
    This motivates a \textbf{spectrum-guided diffusion model} (\secref{sec:model}), which injects spectrum priors into diffusion denoising to brighten low-light content while constraining structure evolution.
    \item \textbf{Human-perceived gains saturate, enabling adaptive computation through reuse and skipping.}
    While diffusion models can significantly improve perceptual quality, running a fixed number of denoising steps with full attention computation is often prohibitive on mobile and wearable devices.
    Meanwhile, resource availability and video content vary over time (\eg background workloads, thermal throttling, static vs. fast-motion segments).
    Since users prioritize temporal continuity and responsiveness, extra computation beyond perceptual thresholds often yields diminishing returns.
    This motivates \textbf{resource-aware dynamic control} (\secref{sec:system}) that adapts computation at runtime to balance enhancement quality, latency, and energy.
\end{itemize}

\fakeparagraph{Architecture}
Based on these insights, \sysname integrates two tightly coupled components (\figref{fig:overview}). 
% \wmf{Accordingly, the main contribution of \sysname is the integrated system design itself, while the spectrum-guided model and the adaptive controller serve as its two complementary technical pillars.}
First, its \textbf{spectrum-guided diffusion model} (\secref{sec:model}) predicts amplitude-based illumination guidance and uses phase as a structural anchor to guide denoising, suppressing flicker and geometry drift without relying on motion estimation.
Second, it builds a lightweight \textbf{closed-loop control stack} (\secref{sec:system}) that exposes efficiency knobs via hierarchical reuse and selective skipping across \textit{frames}, \textit{denoising steps}, and \textit{attention computation}.
A runtime profiler and controller then select the configuration to satisfy real-time constraints under fluctuating resources, while preserving user-perceived quality.

%% file: body/model.tex
\section{SPECTRUM-GUIDED DIFFUSION MODEL FOR LOW-LIGHT VIDEO ENHANCEMENT}   
\label{sec:model}
This section presents the spectrum-guided diffusion model of \sysname for low-light video enhancement. 
Our design is motivated by two requirements: \textit{(i) low-light enhancement} should increase illumination without distorting structure; and \textit{(ii) video enhancement} must be temporally stable. 
We build our model upon diffusion transformers ~\cite{2023DIT}given their strong performance on image restoration and video generation.
However, existing diffusion-based low-light enhancers~\cite{2023diffretinex,2023exposurediffusion,2024fourierdiff,2024lightendiffusion} are developed for \textit{images} and process frames independently when applied to videos, leading to temporal inconsistency. 
Meanwhile, diffusion models designed for \textit{video generation} ~\cite{2022videodiffusion,2024opensora,ma2024latte} emphasize visual plausibility and lack explicit mechanisms to anchor structure, making them prone to structural drift or hallucinated details when repurposed for video enhancement. 
To meet these requirements, we introduce a \textit{spectrum prior network} that provides amplitude-based illumination priors with phase as a structural anchor (\secref{sec:model:spectrum}), and a \textit{spectrum-conditioned diffusion transformer} that injects these priors to guide denoising and stabilize temporal behavior (\secref{sec:model:diffusion}).

\subsection{Spectrum Prior Network}
\label{sec:model:spectrum}
\begin{figure*}[t]
  \centering
  \includegraphics[width=0.9\linewidth]{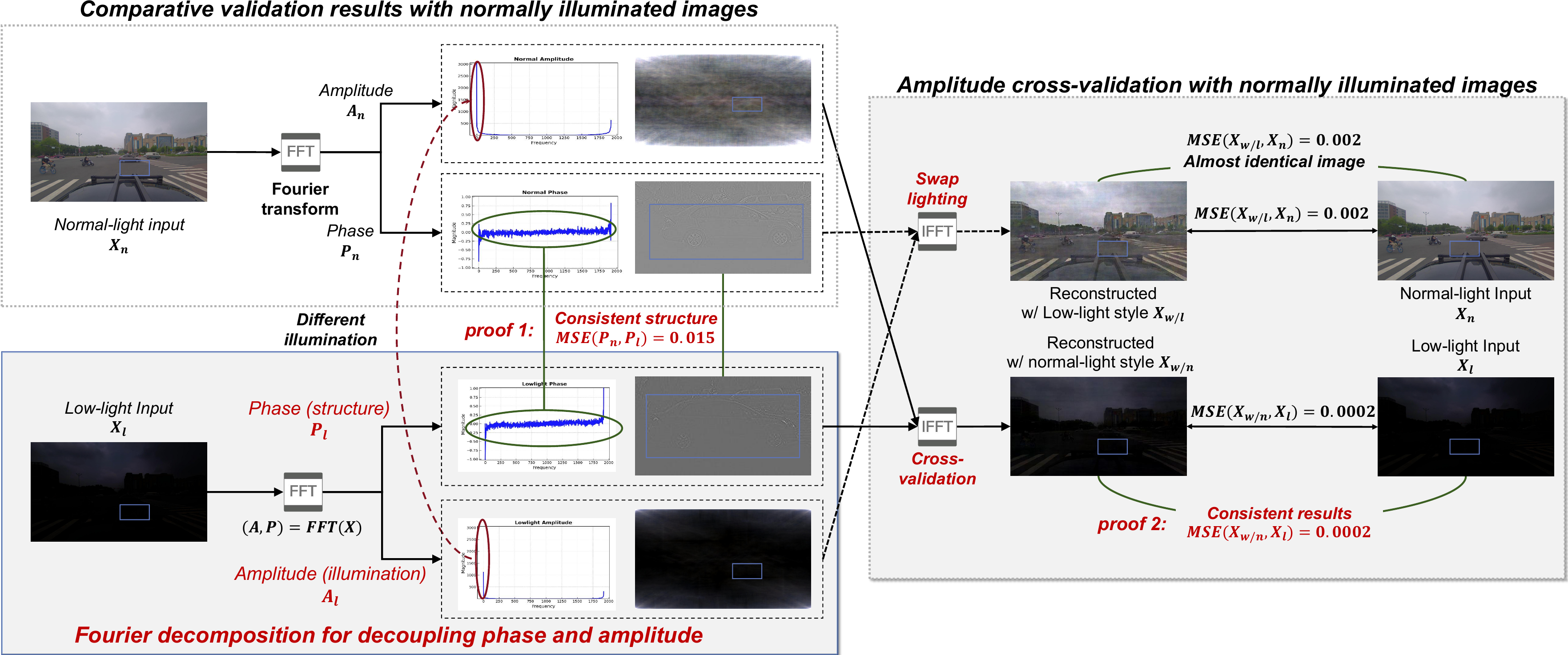}
  \caption{Illustration of Fourier-based decomposition of illumination and structure.}
  \label{fig:exchange}
\end{figure*}

This subsection introduces the \textit{Spectrum Prior Network (SP)}, a lightweight module that extracts spectrum-based priors for diffusion denoising.
SP is motivated by a simple observation: in low-light videos, illumination degradation mainly reduces signal energy, while the underlying scene layout is comparatively stable (\figref{fig:exchange}).
Accordingly, SP predicts an \textit{illumination modulation prior} in the \textit{amplitude} spectrum and retains the \textit{phase} spectrum as a \textit{structural anchor}.
These priors are projected into conditioning tokens and injected into the diffusion transformer (\secref{sec:model:diffusion}) to guide both enhancement quality and temporal stability.

\fakeparagraph{Spectrum Decomposition}
Given an input frame $x$, we compute a luminance proxy $Y=\psi(x)$ (\eg the $Y$ channel) and apply a Fourier transform:
\begin{equation}\label{eq:freq_decomp_keep_v2}
\mathbf{S}=\mathcal{F}(Y),\qquad 
\mathbf{A}_{in}=|\mathbf{S}|,\qquad 
\mathbf{P}_{in}=\angle\mathbf{S},
\end{equation}
where $\mathcal{F}(\cdot)$ denotes the 2D Fourier transform, and $\mathbf{A}_{in}$ and $\mathbf{P}_{in}$ are the amplitude and phase spectra.

\fakeparagraph{Network Design}
We adopt the amplitude-transform formulation (i.e., predicting a transform map rather than directly regressing Fourier amplitudes), which has been shown to better preserve amplitude structure and avoid detail/color degradation under low-light conditions~\cite{2023fourllie}.
Specifically, SP predicts a spatially varying \textit{amplitude transform map} $\mathbf{M}$ that regulates how much illumination should be amplified.
We implement SP as a lightweight network with four Fourier Processing (FP) blocks and skip connections~\cite{2023fourllie}.
Each FP block follows an FFT-process-IFFT routine: intermediate features are modulated in the frequency domain using lightweight $1\times1$ convolutions on the amplitude and phase branches, and then mapped back to the spatial domain via a $3\times3$ convolution with residual connections.
A sigmoid layer constrains $\mathbf{M}$ to $(0,1)$ to avoid excessive amplification and improve stability.

\fakeparagraph{Amplitude Modulation and Phase Preservation}
SP performs illumination correction by modulating the amplitude spectrum while keeping the phase fixed:
\begin{equation}
\mathbf{A}_{out}=\mathbf{A}_{in}\odot\frac{1}{\mathbf{M}+\varepsilon},
\qquad
Y^{(s1)}=\mathcal{F}^{-1}\!\Big(\mathbf{A}_{out}\odot e^{j\mathbf{P}_{in}}\Big),
\label{eq:amp_adjust_keep_v2}
\end{equation}
where $\odot$ denotes element-wise multiplication, $\varepsilon$ is a small constant for numerical stability, and $\mathcal{F}^{-1}(\cdot)$ is the inverse Fourier transform.
This yields a structure-faithful luminance estimate $Y^{(s1)}$, which serves as an effective prior for subsequent diffusion denoising.

SP is supervised using an amplitude alignment objective against the normal-light ground truth:
\begin{equation}
\mathcal{L}_{\mathrm{freq}}=
\left\|
\left|\mathcal{F}(Y^{(s1)})\right|
-
\left|\mathcal{F}(Y_{\mathrm{gt}})\right|
\right\|_2^2,
\label{eq:freq_loss_keep_v2}
\end{equation}
where $Y_{\mathrm{gt}}=\psi(x_{\mathrm{gt}})$ is the luminance proxy of the ground-truth frame.
Finally, SP outputs are projected into conditioning tokens $c=\Pi(\mathbf{M},\mathbf{P}_{in})$ using a lightweight projector $\Pi(\cdot)$, and injected into the diffusion transformer for spectrum-conditioned denoising.

\begin{figure*}[t]
  \centering
  \includegraphics[width=0.7\linewidth]{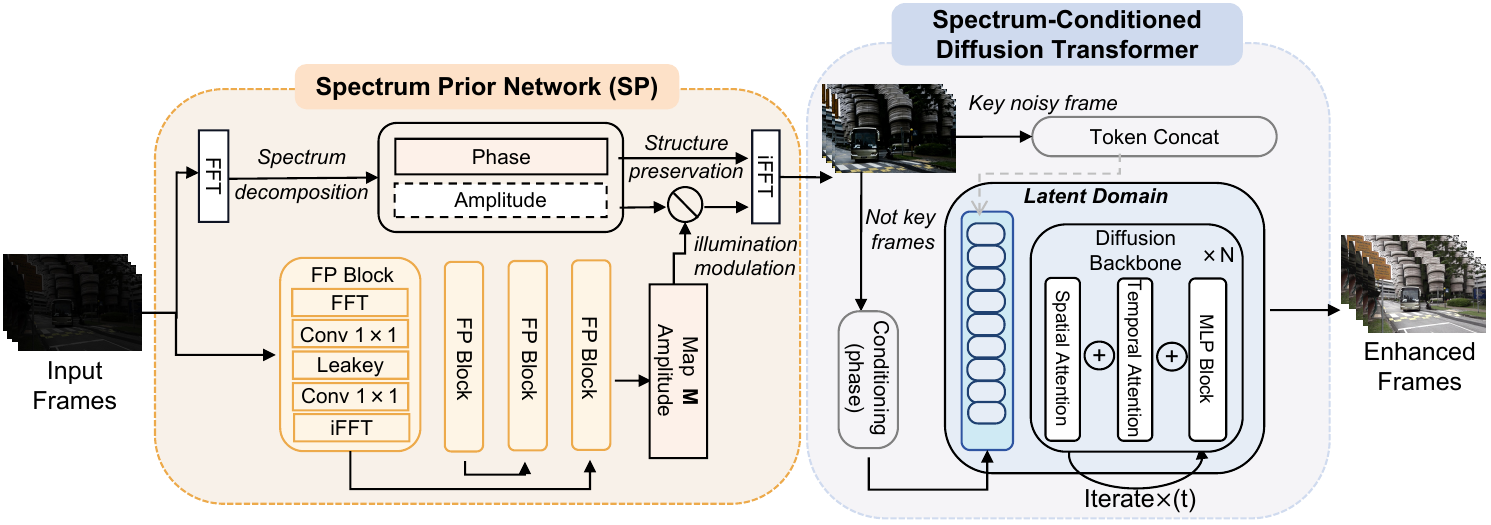}
  % \vspace{-3mm}
  \caption{{Architecture of the Spectrum-guided Diffusion Transformer model for lowlight enhancement.}
  }
  \vspace{-3mm}
  \label{fig:model}
\end{figure*}

\subsection{Spectrum-Conditioned Diffusion Transformer}
\label{sec:model:diffusion}
This subsection describes how we use the spectrum priors from SP to guide diffusion denoising (\figref{fig:model}) for \textit{temporally stable} low-light video enhancement.
In particular, the priors are used in two complementary ways:
\textit{(i)} they condition the denoising backbone (\secref{sec:model:diffusion:backbone}) to improve illumination correction and structure preservation; and
\textit{(ii)} they provide signals to stabilize temporal behavior (\secref{sec:model:diffusion:temporal}) by regulating reuse and suppressing diffusion-induced randomness.

\subsubsection{Diffusion Backbone}
\label{sec:model:diffusion:backbone}
We adopt a DiT-style diffusion backbone~\cite{2023DIT,chen2023pixart} for reverse denoising in the \textit{latent domain},
where input video frames are first encoded into compact spatial tokens using a lightweight encoder,
following the standard latent diffusion paradigm.
This design reduces spatial resolution by 4-8$\times$, substantially lowering computational and memory overhead on ubiquitous devices.
The backbone follows a spatio-temporal Transformer design, and each denoising block is composed of three functional components:
each denoising block contains three components:
\textit{(i) spatial self-attention} to model intra-frame structures and textures;
\textit{(ii) temporal self-attention} to propagate reliable information across neighboring frames and suppress frame-wise stochastic variations; and
\textit{(iii) Spectrum-guided conditional diffusion } that injects SP-derived conditioning tokens $c$ into denoising process by encoding spectrum-derived representations as latent conditioning signals, which are incorporated into the diffusion model to steer the denoising trajectory toward frequency-consistent and temporally stable video enhancement results.
Unlike purely frame-wise diffusion, temporal attention enables the model to reuse latent representations from adjacent frames as an implicit temporal prior, which improves stability under motion and reduces diffusion-induced flicker.
Meanwhile, spectrum-conditioned cross-attention decouples illumination control from structural generation by explicitly providing amplitude and phase cues at each denoising step.
By applying spectral conditioning consistently throughout the reverse diffusion trajectory, the model enhances visibility while preserving scene layout and temporal coherence, which is critical for low-light video enhancement in mobile and wearable scenarios.

\subsubsection{Temporal Stabilization Mechanisms}
\label{sec:model:diffusion:temporal}
Diffusion-based video enhancement can still exhibit temporal artifacts due to \textit{unstable propagation}, \textit{stochastic sampling}, and \textit{long-horizon drift}.
Building on the spectrum-conditioned backbone, we introduce three lightweight stabilization mechanisms that address these issues.
\begin{itemize}
    \item \textbf{Stabilizing Propagation: Spectrum-Gated Temporal Reuse.}
    Temporal attention enables frame-to-frame propagation, but naively reusing past information can be harmful when illumination changes abruptly or motion is fast.
    We therefore \textit{adapt the reuse strength} using spectrum cues.
    Specifically, we observe that the \textit{phase spectrum} is closely related to scene structure and is often more stable across frames, whereas the \textit{amplitude spectrum} varies more with illumination.
    Based on amplitude/phase consistency, we compute a compact similarity score $s_i$ and use it to modulate a temporal reuse gate:
    \begin{equation}\label{eq:spectrum_gate_v2}
        \alpha_i=\mathrm{clip} \, \big(\gamma\, s_i,\,0,\,1\big),
    \end{equation}
    where $\gamma$ is a scaling factor and $\alpha_i\in[0,1]$ controls how strongly temporal information is reused.
    When adjacent frames have similar spectra, $\alpha_i$ increases, encouraging stronger reuse.
    \item \textbf{Stabilizing Sampling: Correlated Noise across Frames.}
    Even with controlled propagation, diffusion sampling itself introduces randomness.
    If each frame samples noise independently, small sampling differences can accumulate into perceptible flicker.
    To suppress such frame-wise jitter without optical-flow alignment, we adopt \textit{correlated} noise sampling.
    At denoising step $t$, the noise for frame $i$ is generated as:
    \begin{equation}\label{eq:correlated_noise_simple}
        \epsilon_{i,t}=\rho\,\epsilon_{i-1,t}+\sqrt{1-\rho^2}\,\eta_{i,t},
    \end{equation}
    where $\eta_{i,t}\sim\mathcal{N}(0,I)$ is newly sampled Gaussian noise and $\rho\in[0,1)$ controls the correlation strength.
    A larger $\rho$ increases stochastic consistency across adjacent frames and reduces flicker, while the injected noise maintains diversity and prevents over-smoothing.
    \item \textbf{Stabilizing Structure: Phase-Aware Temporal Stabilization.}
    Over long sequences, stochastic denoising can still accumulate small inconsistencies and introduce \textit{geometry drift}.
    To further stabilize structure, we introduce phase-aware temporal stabilization, which explicitly enforces \textit{continuity of structural evolution} across frames.
    \begin{itemize}
        \item 
        For consecutive frames, we compute luminance proxies $(Y_{i-1},Y_i)$ and their phase spectra $(P_{i-1},P_i)$.
        We characterize inter-frame structural evolution using a phase-change descriptor:
        \begin{equation}
            \Delta P_i=P_i-P_{i-1},\qquad w_k=A_{i-1}(k)^{\beta},
        \end{equation}
        where $k$ indexes frequency bins and $w_k$ suppresses noise-dominated components via amplitude-aware weighting.
        The exponent $\beta>0$ controls the sensitivity to reliable frequency components;
        $\Delta P_i$ captures motion-consistent structural changes while filtering enhancement-induced artifacts.
        \item 
        Rather than being injected through an explicit projection or concatenation, the phase-change cue influences temporal attention by \emph{constraining the set of plausible contextual representations} reused across frames. Specifically, correlated noise sampling aligns the diffusion trajectories of adjacent frames, while cached latent tokens from the previous frame already encode spectrum-guided structural priors from Stage~1. Under this setting, temporal attention operates over a restricted and structurally consistent context space. The phase-change descriptor $\Delta P_i$ thus serves as a criterion for distinguishing motion-consistent structural evolution from stochastic sampling variations, encouraging the model to reuse phase-consistent latent structures while suppressing diffusion-induced jitter.
        From a modeling perspective, this effect can be abstractly described as current-frame latent tokens attending to prior-frame context under phase-consistency constraints:
        \begin{equation}
        \mathbf{z}'_i \leftarrow 
        \mathrm{Attn}_{\Delta P_i}\!\big(Q(\mathbf{z}_i), K(\mathbf{z}_{i-1}), V(\mathbf{z}_{i-1})\big),
        \end{equation}
        where $\Delta P_i$ does not act as an explicit conditioning input, but modulates the attention behavior by favoring reuse of phase-consistent structures while suppressing stochastic variations introduced by diffusion sampling.
        This encourages the model to reuse reliable \textit{structural} evidence, distinguish \textit{motion-induced} phase changes from diffusion randomness, and stabilize geometry without optical-flow alignment.
        \item 
        We further enforce inter-frame consistency with a regularization loss:
        \begin{equation}
            \mathcal{L}_{phase}=\sum_i\left(1-\mathrm{Coh}(P_{\hat{Y}_i},P_{\hat{Y}_{i-1}})\right),\qquad \mathrm{Coh}(P_a,P_b)=\frac{\left|\sum_k w_k e^{j(P_a(k)-P_b(k))}\right|}{\sum_k w_k},
        \end{equation}
        which penalizes abnormal phase deviations while preserving legitimate motion.
        As a result, the enhanced video exhibits smooth and physically plausible structural evolution.
    \end{itemize}
\end{itemize}

All components of the diffusion transformer, including the backbone and the temporal stabilization mechanisms, operate in the \textit{latent domain} for efficiency.
Operating in the latent space reduces spatial resolution by 8$\times$, lowering latency and memory usage (\secref{sec:model:diffusion:backbone}).
Finally, beyond these architectural choices, we further design explicit compute reuse mechanisms to boost real-time performance on resource-limited devices, as explained next.

%% file: body/system.tex
\section{USER-PERCEPTION-AWARE DYNAMIC CONTROL FOR REAL-TIME VIDEO ENHANCEMENT}
\label{sec:system}
This section presents the user-perception-aware dynamic control stack of \sysname for real-time low-light video enhancement.
While our spectrum-guided diffusion model (\secref{sec:model}) improves perceptual quality, iterative denoising and repeated attention make it hard to meet mobile/wearable latency and energy budgets.
To address this, \sysname adopts a lightweight closed-loop design with three components:
a \textbf{hierarchical reuse and skipping strategy} (\secref{sec:param}) that exploits redundancy across frames, attention, and denoising steps;
a \textbf{resource-aware controller} (\secref{sec:Controller}) that meets a real-time latency budget while balancing quality and energy; and
a \textbf{runtime cost profiler} (\secref{sec:Profiler}) that estimates resource usage with minimal overhead.

\begin{figure*}[t]
    \centering
    % ===== (a) 帧间相似度矩阵 =====step-wise redundancy, frame-wise redundancy, andattention-map redundancy
    \subfloat[frame-wise redundancy]{
        \includegraphics[height=0.18\textwidth]{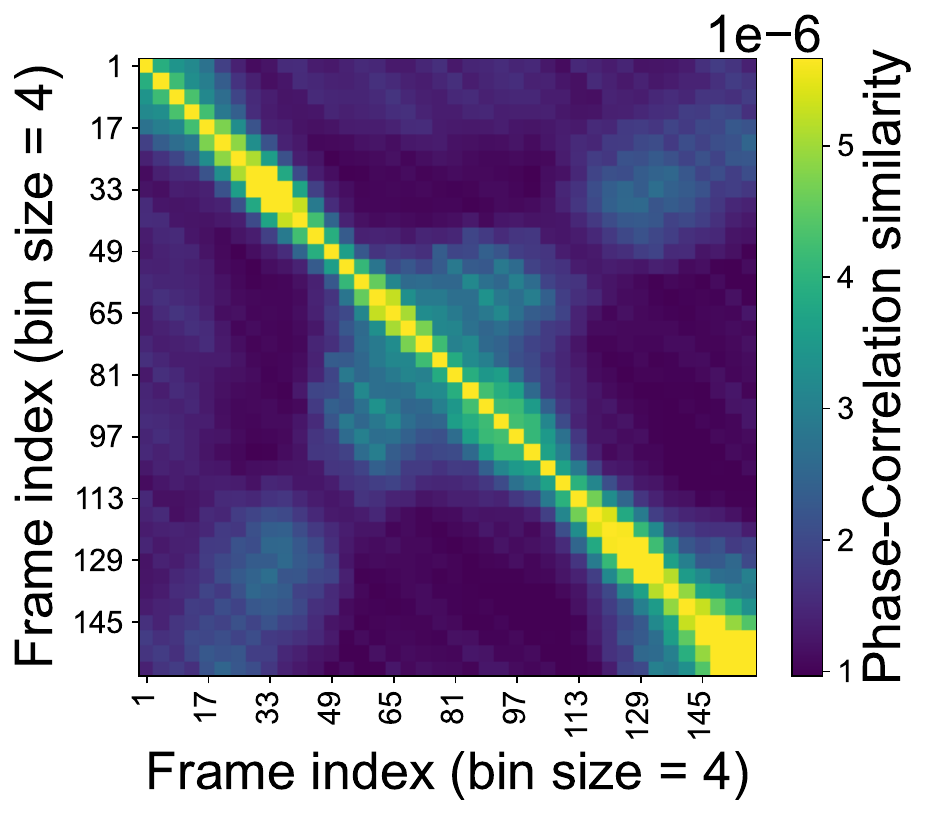}
        \label{fig:frame_sim}
    }
    \hspace{4mm}
    % ===== (b) 注意力层间相似度矩阵 =====
    \subfloat[attention redundancy]{
        \includegraphics[height=0.18\textwidth]{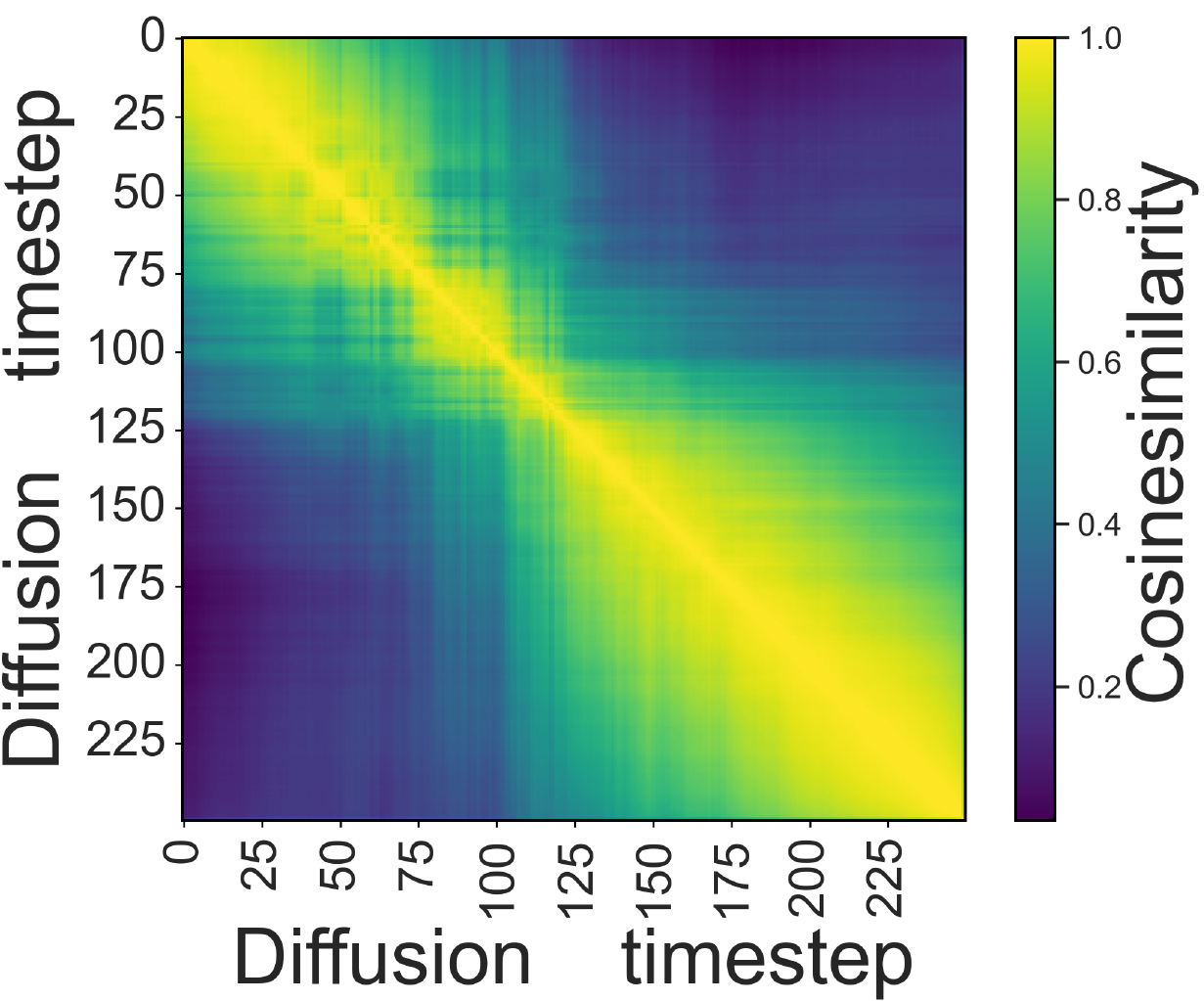}
        \label{fig:attn_corr}
    }
    \hspace{4mm}
    % ===== (c) 不同步数的时延曲线 =====
    \subfloat[mlp redundancy]{
        \includegraphics[height=0.18\textwidth]{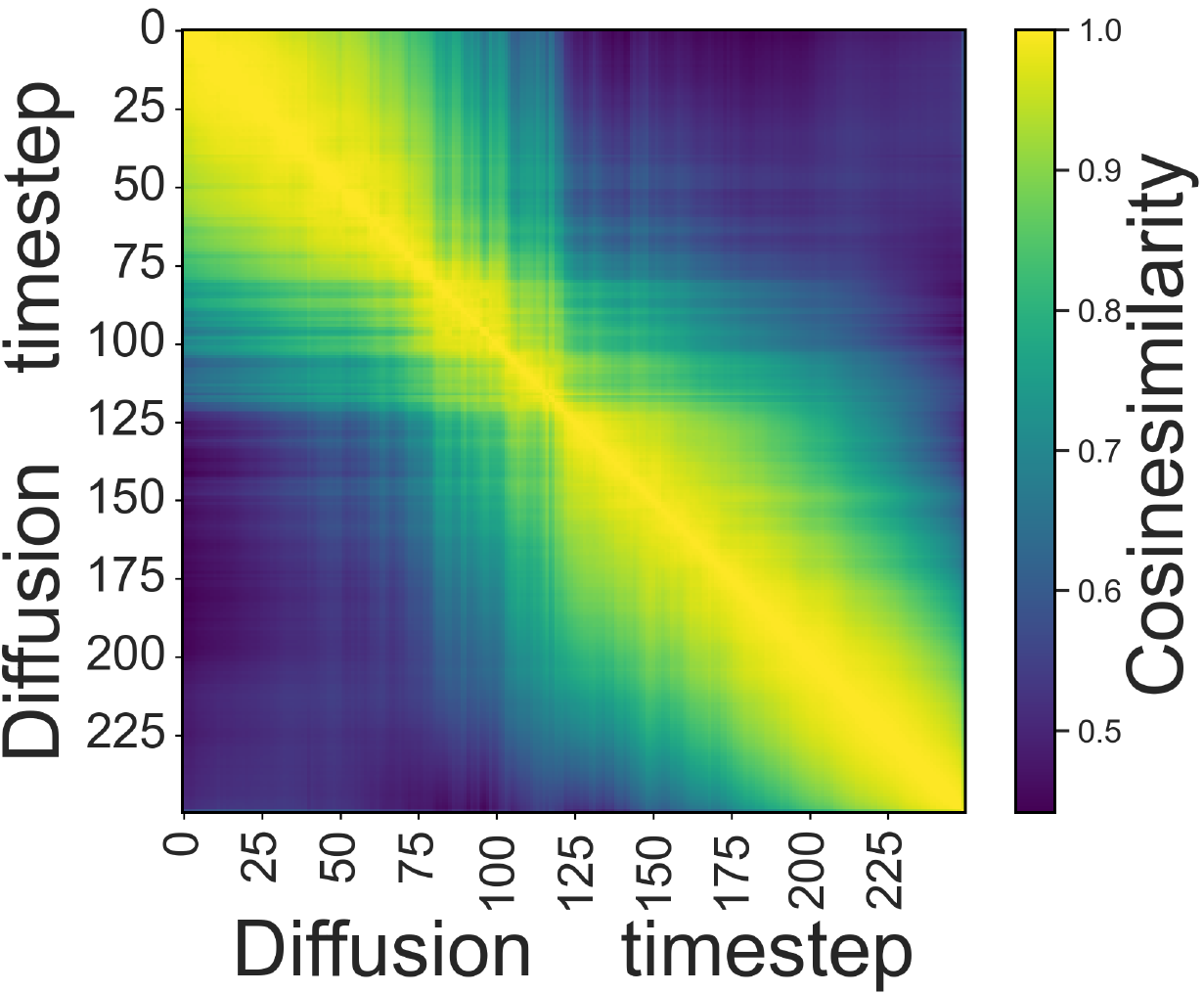}
        \label{fig:latency_steps}
    }
    % ===== Caption =====
    \vspace{-4pt}
    \caption{Temporal redundancy and efficiency characteristics in diffusion-based video enhancement.
}
    \vspace{-4mm}
    \label{fig:multi_analysis}
\end{figure*}

\begin{figure*}[t]
  \centering
  \includegraphics[width=0.8\linewidth]{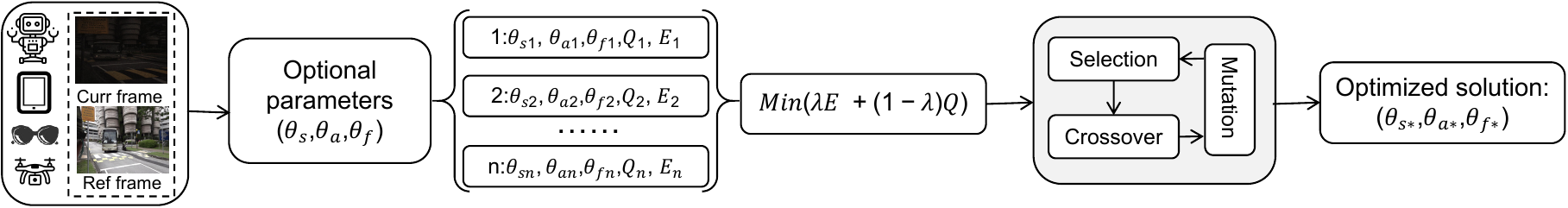}
  \vspace{-3mm}
  \caption{{Workflow of the optimizer in the resource-aware dynamic controller.}}
  \vspace{-2mm}
  \label{fig:controller}
\end{figure*}

\subsection{Hierarchical Reuse and Skipping Strategies}
\label{sec:param}
Diffusion-based low-light video enhancement is computationally expensive, as it iteratively denoises each frame and repeatedly computes transformer attention.
In mobile/wearable use, users are often more sensitive to \emph{flicker and delay} than marginal per-frame fidelity, especially during driving/riding/running or on near-eye displays.
And videos exhibit redundancy at three levels:
\textbf{frame-wise redundancy} across adjacent frames with similar structure/illumination; \textbf{attention redundancy} when attention states change slowly across neighboring denoising steps; and \textbf{sampling redundancy} where enhancing every frame is unnecessary under \textit{mild motion}.
This motivates \sysname to trade \emph{redundant computation} for \emph{perceptual stability} and \textit{real-time deployment} by exploiting four coordinated adaptation mechanisms.

% To exploit these redundancies for real-time deployment, \sysname integrates four coordinated adaptation mechanisms:
% \textbf{keyframe-based redundancy exploitation} controls the frame enhancement rate via $\theta_f$ by selectively enhancing structurally representative frames while reusing cached latent context for intermediate frames;
% \textbf{attention-level reuse} avoids redundant attention computation across neighboring denoising steps by reusing step-wise attention outputs when latent evolution remains stable, governed by $\theta_a$;
% \textbf{sampling-step adaptation} directly regulates the denoising depth per enhanced keyframe through $\theta_s$, enabling fine-grained trade-offs between visual refinement and latency; and
% \textbf{frame-resolution adaptation} scales the inference resolution via $\theta_d$ to proportionally adjust per-step computation under dynamic latency and energy budgets.
% To exploit these redundancies for real-time deployment, \sysname integrates three corresponding strategies \textbf{frame-level feature caching} reuses latent representations and key-value states across frames; \textbf{attention-level reuse} avoids redundant attention computation across denoising steps; and \textbf{keyframe-based sampling} enhances sparse keyframes while propagating intermediate frames via lightweight motion-guided warping.

\begin{figure*}[t]
    \centering
    % ===== Row 1 =====
    \subfloat[Frame-level Feature Caching (\(\theta_f\))]{
        \includegraphics[width=0.41\textwidth]{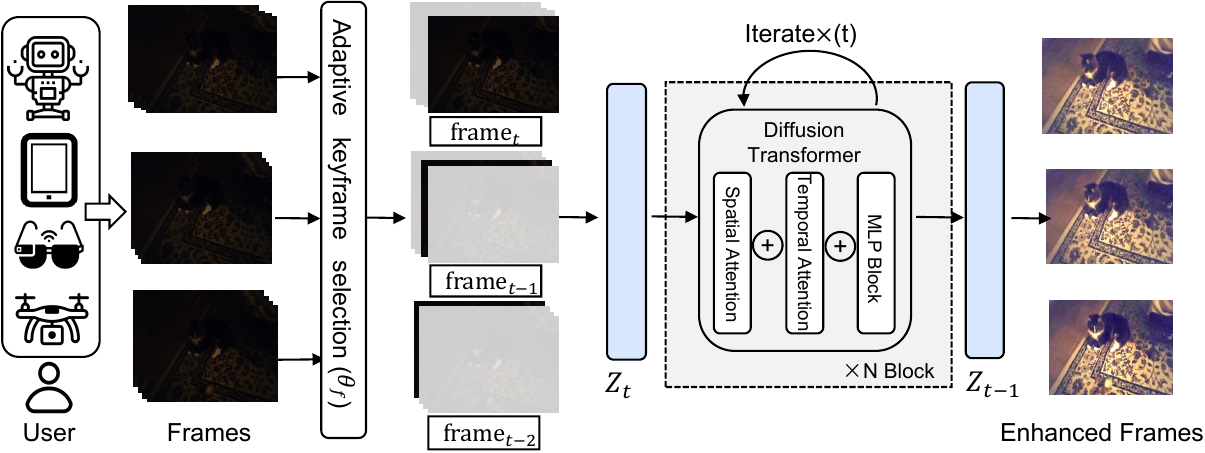}
        \label{fig:frame_cache}
    }
    \hspace{4mm}
    \subfloat[Attention-level Temporal Reuse (\(\theta_a\))]{
        \includegraphics[width=0.33\textwidth]{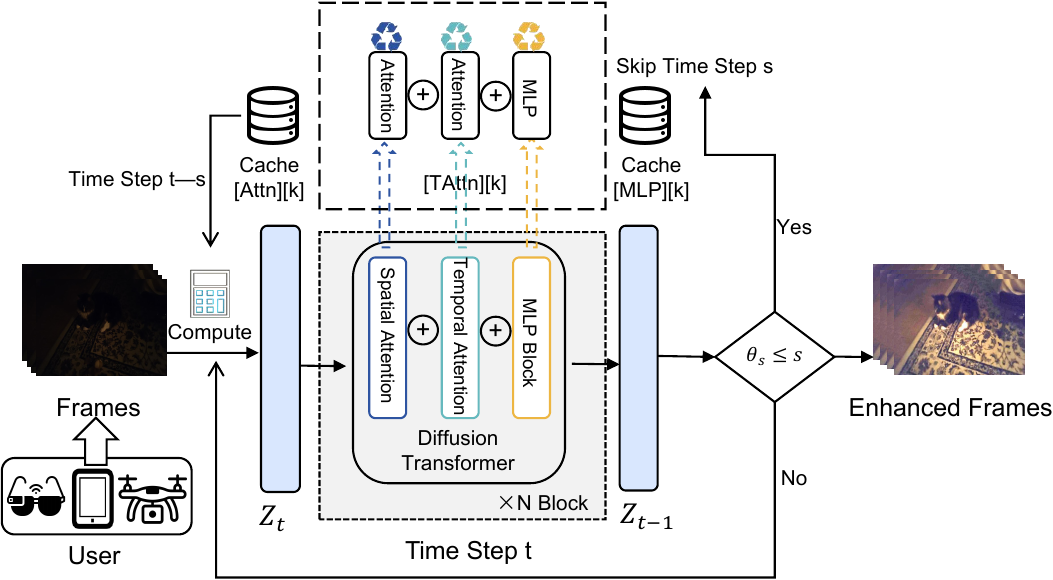}
        \label{fig:attn_reuse}
    }

    \vspace{4pt} % 控制两行间距（可调）
    % ===== Row 2 =====
    \subfloat[Sampling-step Adaptation (\(\theta_s\))]{
        \includegraphics[height=2.5cm]{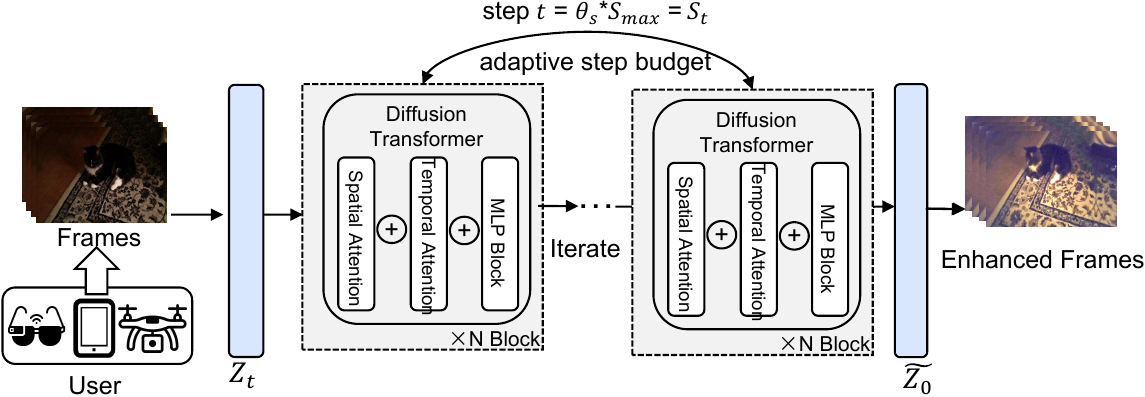}
        \label{fig:step_adapt}
    }\hspace{4mm}
    \subfloat[Frame-resolution Adaptation (\(\theta_d\))]{
        \includegraphics[height=2.5cm]{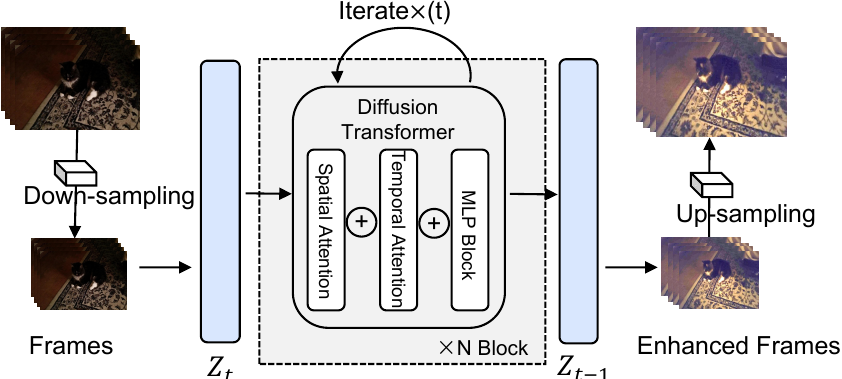}
        \label{fig:res_adapt}
    }
    \vspace{-4pt}
    \caption{Illustration of hierarchical reuse and skipping mechanisms in the proposed runtime scaling strategy.}
    \vspace{-6pt}
    \label{fig:reuse_mechanisms}
\end{figure*}

% \begin{figure*}[t]
%     \centering

%     % ================= Row 1 (same height, full text width) =================
%     \subfloat[Frame-level Feature Caching (\(\theta_f\))]{
%         \includegraphics[height=2.7cm]{figure/frame-level.pdf}
%         \label{fig:frame_cache}
%     }\hfill
%     \subfloat[Attention-level Temporal Reuse (\(\theta_a\))]{
%         \includegraphics[height=3cm]{figure/atteion-level.pdf}
%         \label{fig:attn_reuse}
%     }

%     \vspace{4pt}

%     % ================= Row 2 (same height, full text width) =================
%     \subfloat[Sampling-step Adaptation (\(\theta_s\))]{
%         \includegraphics[height=2.7cm]{figure/step-reuse.pdf}
%         \label{fig:step_adapt}
%     }\hfill
%     \subfloat[Frame-resolution Adaptation (\(\theta_d\))]{
%         \includegraphics[height=2.7cm]{figure/dowmsample.pdf}
%         \label{fig:res_adapt}
%     }

%     \vspace{-4pt}
%     \caption{Illustration of hierarchical reuse and skipping mechanisms in the proposed runtime scaling strategy.}
%     \vspace{-6pt}
%     \label{fig:reuse_mechanisms}
% \end{figure*}

 \textbf{Frame-Level Redundancy Exploitation via Keyframe Selection ($\theta_f$)}.
    Diffusion-based video enhancement applies multi-step denoising to every frame, making latency proportional to the number of processed frames. However, adjacent mobile video frames exhibit strong temporal redundancy. As shown in Fig.~\ref{fig:frame_sim}, over 70\% of adjacent frame pairs have phase-correlation similarity above 0.9, indicating largely consistent amplitude--phase structures under moderate motion and illumination changes.
    To exploit this redundancy, \sysname enhances only selected \emph{keyframes} and reuses cached latent states (\eg features and attention context) for intermediate frames, avoiding explicit motion estimation. Illumination-related amplitude components are inherited from the latest keyframe, while fine-grained variations are adaptively refined. The keyframe interval is determined by the phase-correlation similarity $s_t\!\in\![0,1]$ between the current frame and the latest keyframe:
    \begin{equation}
    K_t = \mathrm{clip}\!\left(\left\lfloor \frac{\kappa}{1-s_t+\epsilon}\right\rfloor,\, K_{\min},\, K_{\max}\right),
    \end{equation}
    where $\kappa$ is a scaling factor, $\epsilon$ prevents division by zero, and $K_{\min}$/$K_{\max}$ bound the interval for responsiveness and stability (Fig.~\ref{fig:frame_cache}). 
    Higher similarity results in a longer reuse interval, whereas lower similarity triggers earlier keyframe updates, reducing redundant computation while preserving temporal consistency.

 \textbf{Attention-Level Reuse across Denoising Steps ($\theta_a$)}.
    Within diffusion transformers, neighboring denoising steps often operate on highly correlated latent states, making attention computation partially redundant as shown in Fig.\ref{fig:attn_corr} and Fig.\ref{fig:latency_steps}.
    Within diffusion transformers, neighboring denoising steps often operate on highly correlated latent states, making attention computation partially redundant .
    Hence, \sysname reuses cached attention outputs when they remain sufficiently similar across steps.
    Let $O_{s}^{\mathrm{attn}}$ be the multi-head attention output at denoising step $s$, and $O_{s'}^{\mathrm{attn}}$ the cached output from the most recent step $s' < s$.
    We compute cosine similarity $S_s = \frac{\left\langle O_{s}^{\mathrm{attn}},\, O_{s'}^{\mathrm{attn}} \right\rangle}{\left\| O_{s}^{\mathrm{attn}} \right\|_2 \cdot \left\| O_{s'}^{\mathrm{attn}} \right\|_2}$.
    If $S_s > \tau_{\mathrm{attn}}(\theta_a)$, \sysname reuses $O_{s'}^{\mathrm{attn}}$; otherwise, it recomputes attention and updates the cache(seeing \figref{fig:attn_reuse}).
    A larger $\theta_a$ promotes reuse for efficiency, while a smaller $\theta_a$ enforces more frequent recomputation to preserve temporal stability.

 \textbf{Sampling-Step Adaptation via Stability-Aware Denoising Scheduling ($\theta_s$)}.
    Diffusion inference remains the primary latency bottleneck, because the denoising network is executed repeatedly over a step schedule $\{t_1, \dots, t_S\}$, making runtime approximately proportional to the number of sampling steps $S$. To provide fine-grained control of per-keyframe computation under dynamic resource constraints, \textit{sysname} introduces a sampling-step knob $\theta_s$ that directly determines the step budget used for each enhanced keyframe.
    Mechanistically, $\theta_s$ maps the current resource state (\eg latency/energy budget) and the estimated input difficulty (\eg darkness level or structural change) to a step count:
    \begin{equation}
    S_t = \mathrm{clip} \left( \left\lfloor \theta_s \cdot S_{\max} \right\rfloor, S_{\min}, S_{\max} \right)
    \end{equation}
    where $S_t$ is the number of denoising steps executed for keyframe $t$, and $[S_{\min}, S_{\max}]$ bounds the feasible range. 
    When energy is sufficient or latency constraints are relaxed, $\theta_s$ increases, allocating more denoising steps to better refine high-frequency details and reduce residual noise. Under tight latency or energy budgets, $\theta_s$ decreases, shortening the denoising trajectory and reducing compute(seeing Fig.~\ref{fig:step_adapt}), while relying on (i) the Stage-1 spectrum-guided prior and (ii) temporal stabilization (e.g., correlated noise) to limit flicker and maintain coherent structure. This sampling-step adaptation complements keyframe selection: keyframe selection reduces the number of frames requiring diffusion, whereas $\theta_s$ controls the per-keyframe denoising depth, jointly balancing quality, latency, and energy for streaming deployment.
 
 \textbf{Resolution Scaling ($\theta_d$)}.
    In addition to reducing denoising steps, \sysname adapts the input resolution to directly control per-frame computation.
    Specifically, $\theta_d \in (0,1]$ denotes a down-sampling factor applied to each incoming frame before enhancement(seeing Fig.~\ref{fig:reuse_mechanisms}); \sysname performs enhancement on the resized frames (aligned to the model stride) and then upsamples the result back to the target resolution.

\subsection{Resource-Aware Controller}
\label{sec:Controller}
We design a controller that dynamically coordinates the four efficiency knobs in \secref{sec:param}, frame-level caching ($\theta_f$), attention reuse ($\theta_a$), and sampling-step adaptation ($\theta_s$), and frame-resolution adaptation($\theta_d$) to maintain real-time enhancement under time-varying scene dynamics and device budgets.

\fakeparagraph{Formulation} 
At each time $t$, given the incoming stream $\mathrm{video}(t)$ and measured resource state $\mathrm{resource}(t)$, the controller selects a configuration
$\boldsymbol{\theta}=(\theta_s,\theta_a,\theta_f,\theta_d)$ to balance enhancement quality, energy cost, and latency.
We model this as a constrained multi-objective control problem:
\begin{align}
\begin{aligned}
\max_{\boldsymbol{\theta}} \quad & Q\!\left(\boldsymbol{\theta};\mathrm{video}(t)\right), \qquad
\min_{\boldsymbol{\theta}} \quad E\!\left(\boldsymbol{\theta};\mathrm{resource}(t)\right), \\
\text{s.t.} \quad & T\!\left(\boldsymbol{\theta};\mathrm{resource}(t)\right) \le T_{\max},
\end{aligned}
\end{align}
where $Q(\cdot)$ is the resulting video quality, $E(\cdot)$ is the energy cost, and $T(\cdot)$ is the per-frame latency.
This formulation captures the trade-off between perceptual quality and resource expenditure while enforcing real-time constraints.

\fakeparagraph{Optimizer} 
We adopt a two-stage optimization strategy consisting of offline profiling and online adaptation.
\textit{i) Offline.} We construct a Pareto-optimal configuration library that maps runtime knobs $\boldsymbol{\theta}=(\theta_s,\theta_a,\theta_f,\theta_d)$ to their measured quality, energy, and latency $(Q,E,T)$. 
Here, $\theta_s$, $\theta_a$, $\theta_f$, and $\theta_d$ control the sampling-step budget, attention reuse, keyframe interval, and input resolution, respectively. 
The decision space is intentionally limited, with $\theta_s \in \{2,4...18,20\}$,$\theta_a \in \{1,1/2...1/5\}$, $\theta_f \in \{0,1/5,1/3,1/2,1\}$, and $\theta_d \in \{1,1/2,1/3\}$. For each candidate configuration, we measure the corresponding enhancement quality, energy cost, and per-frame latency on representative low-light videos and target hardware profiles. 
We profile all candidate configurations on representative low-light videos and target devices, discard those violating the latency constraint ($T>300$ ms), and retain only Pareto-optimal configurations.
The resulting library typically contains 35$\sim$45 configurations per platform.
\textit{ii) Online.} The controller monitors device status and lightweight scene indicators, and selects a runtime configuration from the Pareto library according to a preference weight $\lambda\!\in\![0,1]$, which balances enhancement quality and efficiency:
\begin{equation}
\boldsymbol{\theta}^\star = \arg\max_{\boldsymbol{\theta}\in\mathcal{L}}
\; U(\boldsymbol{\theta})
\quad \text{s.t.}\quad T(\boldsymbol{\theta}) \le T_{\max},
\qquad
U(\boldsymbol{\theta})=(1-\lambda)\,Q(\boldsymbol{\theta})-\lambda\,E(\boldsymbol{\theta}),
\end{equation}
where $\mathcal{L}$ denotes the offline Pareto library. 
As scene dynamics or resource availability change, the controller updates $\lambda$ and switches configurations accordingly, maintaining a balanced tradeoff among quality, latency, and energy during real-time enhancement.

\subsection{Runtime Cost Profiler}
\label{sec:Profiler}
To support online control, \sysname includes a lightweight profiler that estimates the \textbf{energy} and  \textbf{latency} cost of a candidate configuration $\boldsymbol{\theta}=(\theta_s,\theta_a,\theta_f,\theta_d)$ without invasive instrumentation.
The profiler combines an analytic layer-level cost model with a small set of runtime signals, and provides cost feedback to the controller in \secref{sec:Controller}.

% \todo{update and condense}
\textit{Energy Cost}.
Extending AdaEnlight~\cite{adaenlight} to diffusion transformers for videos, we model layer-level computation and memory behavior to guide perception-aware runtime control. 
The energy for layer $l$ is decomposed as:
\begin{equation}
E_l = \delta_C \cdot C_l + \epsilon \cdot \delta_{\text{cache}} \cdot M_l + (1 - \epsilon) \cdot \delta_{\text{DRAM}} \cdot M_l + M_l \cdot \delta_{\text{SM}}
\end{equation}
where $C_l$ is the number of MAC operations, $M_l$ is the memory traffic, and $(\delta_C, \delta_{\text{cache}}, \delta_{\text{DRAM}}, \delta_{\text{SM}})$ are device-calibrated energy costs for computation, cache, DRAM, and on-chip memory. $\epsilon$ represents the cache-hit rate, obtained offline through micro-benchmarks.
Key components include:
i) MAC operations $C_l$: computes offline based on the layer's architecture. 
For a diffusion transformer with windowed attention, $C_l$ depends on QKV projections, attention multiplication, and FFN operations.
ii) Memory accesses $M_l$: estimates offline based on dataflow in attention and FFN layers. Memory access includes parameter loads, activation reads/writes, and intermediate buffers.
iii) Cache-hit ratio $\epsilon$: measures online to reflect real-time cache performance, updated every few frames.
The total energy cost for enhancing a video with $n_{\mathrm{frame}}$ frames under adaptive hyperparameters $\theta_f,\theta_a,\theta_d,\theta_s$ is:
\begin{equation}
E_{\mathrm{enhancer}}
=
\sum_{f=1}^{n_{\mathrm{frame}}}
\sum_{t=1}^{S(\theta_s)}
\sum_{l=1}^{L}
\theta_f(f)\,
\theta_a(l)\,
E_l\big(\theta_d^{f,t}\big),
\label{eq:video_energy}
\end{equation}
where $\theta_f(f)$ and $\theta_a(l)$ are binary indicators for frame $f$ and layer $l$, and $S(\theta_s)$ is the number of diffusion steps. This allows the controller to assess the impact of skipping steps and frame down-sampling on energy.

\paragraph{Memory Access Modeling.}

We model memory access mainly as a runtime  constraint, since memory is not the active bottleneck in our deployment.
For each layer $l$, we estimate memory accesses $M_l$ from the attention and FFN dataflow, and aggregate the per-step cost as $M_{\mathrm{step}}=\sum_l \theta_l M_l$.
We use this metric to monitor memory pressure during execution.
Across tested platforms and configurations, memory usage stays stable (peak $<1.7$\,GB, average $\sim1.5$\,GB), making latency and energy the dominant constraints.

\paragraph{Resource Cost}
We further define a composite resource cost to balance energy, computation, and memory:
\begin{equation}
R(\theta_s,\theta_a,\theta_f,\theta_d)
=
\beta_E \,\widehat{E}_{\mathrm{enhancer}}
+
\beta_C \,\widehat{C}_{\mathrm{total}}
+
\beta_M \,\widehat{M}_{\mathrm{total}},
\label{eq:resource_cost}
\end{equation}
where $\widehat{E}{\mathrm{enhancer}}$, $\widehat{C}{\mathrm{total}}$, and $\widehat{M}_{\mathrm{total}}$ are normalized costs for energy, computation, and memory, and $\beta_E$, $\beta_C$, and $\beta_M$ are tunable weights reflecting resource priorities for the platform.

\subsubsection{Profiler Workflow}
\label{sec:Profiler:Workflow}
Accurate, low-overhead performance profiling is essential for adaptive control on mobile platforms. 
\sysname uses a two-stage workflow that combines offline calibration with online integration to bridge hardware-level resource cost modeling and runtime adaptation.
\textit{Offline Calibration and Modeling}.
We derive platform-specific energy coefficients ($\delta_C, \delta_{\text{cache}}, \delta_{\text{DRAM}}, \delta_{\text{SM}}$) and peak throughput baselines through hardware profiling. 
By correlating power measurements with micro-benchmarks, we normalize memory access costs relative to MAC operations, creating a static energy profile independent of the model architecture.
\textit{Online Runtime Monitoring and Integration}.
During inference, the profiler monitors lightweight signals (\eg cache-hit rate) and the current configuration $\Theta$.
By combining analytic models with runtime feedback, it estimates energy and latency \emph{without disrupting user interaction}.
This feedback closes the loop with the controller, enabling continuous adaptation to \textit{scene dynamics}, \textit{device state}, and \textit{user-perceived responsiveness}.

\subsection{Adaptive Control Loop}
\label{sec:Loop}
\sysname employs a \textit{human-centric adaptive control loop} to balance perceived visual quality, energy efficiency, and responsiveness under dynamic usage conditions (\figref{fig:controller}).
The controller periodically (\eg every second) monitors device resources (\eg battery, compute load, memory, and thermal headroom) together with profiler-estimated latency and energy of the current configuration.
When resource budgets or scene dynamics change, it re-optimizes the runtime knobs $(\theta_s,\theta_a,\theta_f,\theta_d)$, covering sampling depth, attention reuse, frame caching, and frame-resolution adaptation, using a Pareto-based multi-objective formulation that maximizes perceived quality under latency and energy constraints.
The selected configuration is applied only after throughput and thermal safety checks, enabling continuous adaptation without disrupting user interaction.
When the input frame rate exceeds the achievable enhancement throughput, the controller favors bounded-latency enhancement by adaptively selecting keyframes and reuse configurations, instead of executing full diffusion on every frame.

%% file: body/experiment.tex
\section{EXPERIMENT}
\label{sec:experiment}

\subsection{Settings}

\textbf{Implementation}.
We implement \sysname in PyTorch and train all models on a workstation with an NVIDIA GeForce RTX 3090 GPU (24GB VRAM, CUDA 12.6). We deploy \sysname on four representative platforms: Google Pixel~7 (Android, 8GB RAM; $Dev_1$), NVIDIA Jetson AGX Orin (Linux, 64GB RAM; $Dev_2$), Xiaomi~17 Ultra (Android, 16GB RAM; $Dev_3$), and XREAL One Pro AR glasses paired with a smartphone for computation ($Dev_4$).
Deployment uses platform-specific runtimes: PyTorch Mobile (.ptl) on Android devices, CUDA-accelerated PyTorch on Jetson, and a tethered pipeline for XREAL One Pro, where captured frames are streamed to the paired smartphone for enhancement and then returned for display.
Our diffusion backbone is built on STDiT and follows the standard DDPM reverse denoising process. The adaptive controller adjusts only the denoising-step budget, while the sampling algorithm remains unchanged. For reproducibility, we fix the random seed and deterministically initialize the diffusion noise under each runtime configuration.
Across all platforms, videos are captured at 20 FPS, resized to \(270\times480\), and processed frame by frame. For platforms with multiple compute backends (\eg CPU/GPU on Jetson), inference is pinned to a single backend (GPU-only) to avoid synchronization overhead and redundant memory transfers.

\textbf{Datasets}.
We evaluate \sysname on both public and self-collected low-light datasets, covering paired/unpaired supervision and diverse camera/object motion patterns \tabref{tab:datasets} lists the details of these datasets.

\begin{itemize}
    % \item \textbf{Public datasets.} We use two paired low-light frame datasets, LSRW~\cite{LSRW} and LOL-v2~\cite{Retinex-Net}, and four low-light video datasets, DID~\cite{DID}, SDSD~\cite{SDSD}, DRV~\cite{DRVsmid}, and LoLi-Phone~\cite{LoLi-Phone}. LSRW/LOL-v2 provide large-scale paired frames with realistic illumination degradations, and are used to learn single-frame priors for exposure correction, color restoration, and denoising. DID is a paired video dataset with motion-induced degradations, and is used to adapt \sysname to temporal dynamics and motion-aware enhancement. 
    % \wmf{All training frames are resized to a fixed resolution; videos are sampled into clips of length \(T=xx\). We train with Adam (lr \(1\times10^{-4}\), batch size 16). The remaining public datasets are used only for testing: SDSD and DRV are paired static-scene videos for full-reference metrics, while LoLi-Phone is an unpaired real-world mobile dataset used to assess perceptual quality and temporal consistency in streaming scenarios.}
    \item \textbf{Public datasets.} We use two paired low-light frame datasets, LSRW~\cite{LSRW} and LOL-v2~\cite{Retinex-Net}, and four low-light video datasets, DID~\cite{DID}, SDSD~\cite{SDSD}, DRV~\cite{DRVsmid}, and LoLi-Phone~\cite{LoLi-Phone}. LSRW/LOL-v2 provide large-scale paired frames with realistic illumination degradations, and are used to learn single-frame priors for exposure correction, color restoration, and denoising. DID is a paired video dataset which can be used to adapt \sysname to temporal dynamics and motion-aware enhancement.  SDSD and DRV are used only for full-reference testing, while LoLi-Phone is used only for real-world streaming evaluation.

    \item \textbf{Training pipeline.} \sysname is trained in two stages. In Stage~1, we train the spectrum-prior branch on paired low/normal-light image datasets as an optional warm-start. Input images are resized/cropped to \(256\times256\). We train with Adam (lr \(1\times10^{-4}\), batch size 8). In Stage~2, we train the video enhancement model on fixed-length clips of \(T=24\) at \(256\times256\). We train with AdamW (backbone lr \(1\times10^{-5}\)). During Stage~2, the diffusion objective and the auxiliary temporal/frequency/exposure consistency terms are optimized jointly within a unified objective. It supports both paired and unpaired video training.

    \item \textbf{Self-collected UbiVideo dataset.}
    We collect \textit{UbiVideo} to evaluate robustness under realistic mobile capture behaviors.
    UbiVideo contains four scenarios: (A) moving camera + moving objects, (B) moving camera + static objects, (C) static camera + moving objects, and (D) static camera + static objects.
    Each includes 30 clips recorded on a Google Pixel~7 (\(120\) videos total), each \(>\)30\,s, captured in real low-light environments with varying exposure and motion.
    We primarily use UbiVideo to assess temporal stability and perceptual consistency under dynamic camera/object motion during continuous on-device enhancement.
\end{itemize}

\begin{table}[t]
\centering
\tiny
\caption{Overview of low-light frame and video datasets (including our self-collected UbiVideo).}
\label{tab:datasets}
\resizebox{0.8\textwidth}{!}{%
\begin{tabular}{|c|l|l|l|}
\hline
\textbf{Datasets} & \multicolumn{1}{c|}{\textbf{Data type}} & \multicolumn{1}{c|}{\textbf{Sample number}} & \multicolumn{1}{c|}{\textbf{Description}} \\ \hline
\textbf{LSRW~\cite{LSRW}} & Paired frames & 5,650 pairs & Real-world paired low/normal-light frames. \\ \hline
\textbf{LOL-v2~\cite{Retinex-Net}} & Paired frames & 1,789 pairs & Paired low-light frames (Real + Synthetic). \\ \hline
\textbf{DID~\cite{DID}} & Paired videos & 413 videos / 41,038 frames & Real-world paired low/normal-light videos. \\ \hline
\textbf{SDSD~\cite{SDSD}} & Paired videos & 150 videos & Paired low/normal-light videos for dynamic scenes. \\ \hline
\textbf{DRV~\cite{DRVsmid}} & Paired RAW videos & 202 videos (up to 110 frames) & Static RAW Bayer sequences with long-exposure GT. \\ \hline
\textbf{LoLi-Phone~\cite{LoLi-Phone}} & Unpaired videos & 120 videos / 45,148 frames & Smartphone-captured low-light videos. \\ \hline
\textbf{UbiVideo } & Unpaired videos & 120 videos & Self-collected mobile low-light videos across four motion scenarios. \\ \hline
\end{tabular}%
\vspace{-3mm}
}
\end{table}

\textbf{Baselines}.
We compare \sysname with twelve representative low-light enhancement methods.

\textit{Frame-wise (single-image) Baselines.}
They process videos frame-by-frame, providing strong per-frame quality/efficiency references but often inducing temporal flicker/jitter due to independent optimization.
\begin{itemize}
    \item \textbf{Zero-DCE++}~\cite{zero-dce++}: A \textbf{state-of-the-art} curve-based method that estimates \textbf{image-adaptive exposure curves} with (little or) no paired supervision, enabling lightweight deployment.
    \item \textbf{LLFlow}~\cite{2022llFlow}: A \textbf{learning-based single-image} generative model that enhances low-light images by modeling their conditional distribution, applied \textbf{frame-by-frame} on videos.
    \item \textbf{URetinex-Net}~\cite{uretinex}: A \textbf{learning-based single-image} Retinex network estimating illumination and reflectance via end-to-end architecture, applied frame-by-frame.
    \item \textbf{Diff-Retinex}~\cite{2023diffretinex}: An image-oriented diffusion model combining generative refinement with illumination-reflectance decomposition; it delivers \textbf{strong perceptual quality} but requires \textbf{iterative sampling} when used \textbf{frame-by-frame}.
    \item \textbf{FourierDiff}~\cite{2024fourierdiff}: A \textbf{more recent} frequency-domain diffusion model leveraging Fourier priors; it offers \textbf{strong perceptual quality} but remains costly due to iterative sampling when applied \textbf{frame-by-frame}.
\end{itemize}

\textit{Frame-centric enhancement + post-hoc temporal modeling Baselines.}
They follow frame-centric pipelines augmented with explicit temporal constraints (e.g., flow-based warping, recurrent propagation, or short-term temporal modules) to suppress flicker.
\begin{itemize}
    \item \textbf{MBLLEN}~\cite{MBLLEN}: A \textbf{classic} multi-branch network for video enhancement, fusing multi-scale features to recover details while maintaining temporal consistency (a representative \textbf{short-term temporal modules} design).
    \item \textbf{StableLLVE}~\cite{stableLLVe}: A \textbf{state-of-the-art} model with temporal regularization that suppresses flickering and maintains motion coherence via \textbf{inter-frame consistency constraints}; it can be fragile when alignment becomes unreliable.
    \item \textbf{FastLLVE}~\cite{fastllve}: A lightweight CNN optimized for real-time inference on commodity devices, balancing enhancement quality and computational efficiency with \textbf{short-term temporal modules}.
    \item \textbf{AdaEnlight}~\cite{adaenlight}: An adaptive high-fidelity framework designed for real-time mobile video enhancement, leveraging lightweight architectures and content-aware adaptation under resource constraints (with \textbf{short-term temporal modules} for stabilization).
\end{itemize}

\textit{Video-native Baselines.}
They optimize videos directly with video-centric data, models, and objectives, typically relying on paired dynamic-video supervision.
\begin{itemize}
    \item \textbf{SDSDNet}~\cite{SDSD}: A \textbf{more recent} static-to-dynamic supervised framework trained with \textbf{paired dynamic-video supervision} on paired low/normal-light video data, using spatiotemporal feature fusion for dynamic scene enhancement.
    \item \textbf{RetinexMCNet}~\cite{retinexmcnet}: A \textbf{recent} Retinex-based supervised LLVE method that combines illumination-reflectance enhancement with a memory controller for temporal consistency.
    \item \textbf{BRVE}~\cite{BRVE}: A  \textbf{recent} binarized raw-domain LLVE method that uses distribution-aware binary convolutions and spatial-temporal shift operations for efficient temporal enhancement.
    \item \textbf{UbiEnlight W/O\_Align}: A variant of UbiEnlight without the second-stage Diffusion Transformer, reported to isolate the contribution of our video-native design.
    \item \textbf{UbiEnlight W/O\_Spectrum Prior}: A variant of UbiEnlight that removes the amplitude and phase spectrum priors, reported to isolate the contribution of our spectrum-guided conditioning design.
\end{itemize}

\begin{table}[t]
\centering
\tiny 
\setlength{\tabcolsep}{6pt}
\caption{Capability checklist under two hardest real-world settings using usability-aware, no-reference constraints.
A method is marked $\checkmark$ only if it satisfies the threshold for that constraint.}
\label{tab:hardcases_checklist}
\begin{tabular}{|c|c|c|c|c|c|c|}
\hline
\multirow{2}{*}{\makecell[c]{\textbf{Hard Case}}} &
\multirow{2}{*}{\makecell[c]{\textbf{Constraint}}} &
\multicolumn{5}{c|}{\textbf{Methods}} \\
\cline{3-7}
& & \textbf{Ours} & \textbf{SDSDNet} & \textbf{MBLLEN} & \textbf{StableLLVE} & \textbf{FastLLVE} \\
\hline
\multirow{3}{*}{\makecell[c]{Dynamic illumination\\+ intense motion}}
& \makecell[c]{Visibility (VPR@0.15 $>$ 0.9)}
& $\checkmark$(0.99) & $\times$(0.58) & $\times$(0.76) & $\checkmark$(0.99) & $\checkmark$(0.91)\\
\cline{2-7}
& \makecell[c]{Detail (MGE $>$ 0.07)}
& $\checkmark$(0.11) & $\checkmark$(0.07) & $\times$(0.04) & $\times$(0.05) & $\times$(0.03) \\
\cline{2-7}
& \makecell[c]{Stability (MF@0.15 $<$ 0.05)}
& $\checkmark$(0.02)& $\times$(0.13) & $\times$(0.06) & $\times$(0.07) & $\times$(0.09) \\
\hline
\multirow{3}{*}{\makecell[c]{Extreme low-light\\+ intense motion}}
& \makecell[c]{Visibility (VPR@0.15 $>$ 0.9)}
& $\checkmark$(0.99) & $\times$(0.53) & $\checkmark$(0.99)& $\checkmark$(0.99) & $\times$(0.48) \\
\cline{2-7}
& \makecell[c]{Detail (MGE $>$ 0.07)}
& $\checkmark$(0.10) & $\times$(0.06) & $\times$(0.03) & $\times$(0.05) & $\checkmark$(0.07)\\
\cline{2-7}
& \makecell[c]{Stability (MF@0.15 $<$ 0.05)}
& $\checkmark$(0.01) & $\times$(0.11) & $\checkmark$(0.03) & $\checkmark$(0.04) & $\times$(0.10) \\
\hline
\end{tabular}

\vspace{2pt}
{\tiny \textit{Notes:} VPR$_\tau=\frac{1}{|\Omega|}\sum_p\mathbf{1}(Y>\tau)$;
MGE$=\frac{1}{|\Omega|}\sum_p\|\nabla Y\|$;
MF$_\tau=\mathbb{E}_t[|Y_t-Y_{t-1}|\mid Y_t>\tau \vee Y_{t-1}>\tau]$.
where $\Omega$ denotes the pixel domain, $|\Omega|$ its size, $p$ a pixel index, $Y_t$ luminance at time $t$, $\nabla Y$ spatial gradients, $\tau$ a fixed visibility threshold, and $\mathbb{E}_t[\cdot]$ temporal expectation.}
\end{table}

\begin{figure}[t]
    \centering
    \subfloat[Quality-latency trade-off on SDSD]{
        \includegraphics[width=0.48\textwidth]{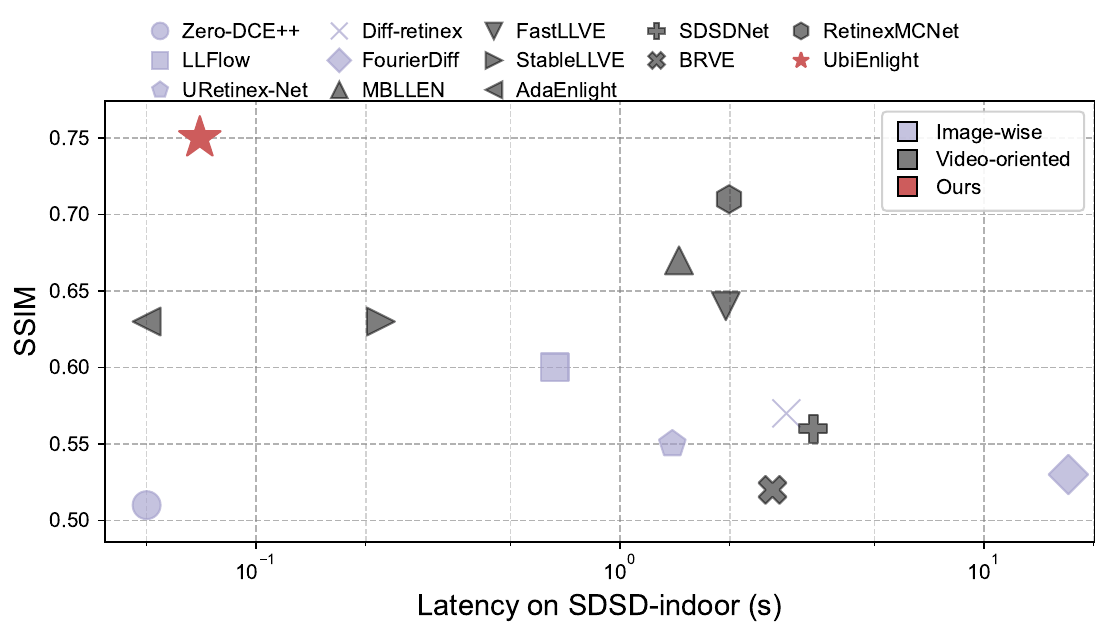}
        \label{fig:sdsd_indoor_latency_ssim}
        \vspace{-5pt}
    }
    \subfloat[Quality-latency usage trade-off on DRV]{
        \includegraphics[width=0.48\textwidth]{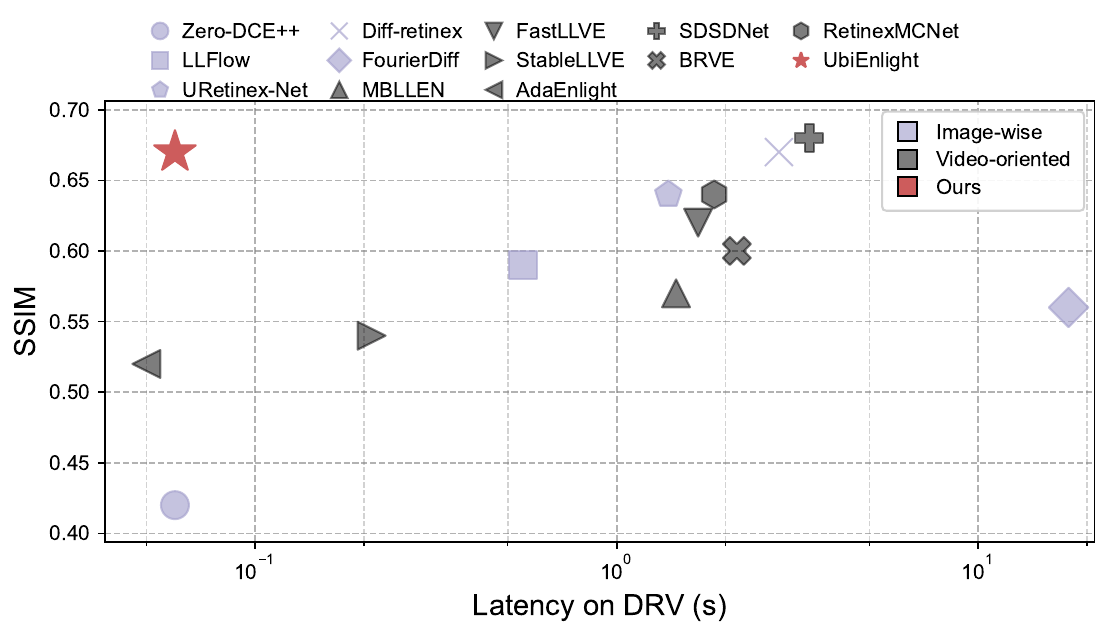}
        \label{fig:drv_latency_psnr}
        \vspace{-5pt}
    }
    \vspace{-3pt}
    \caption{
    Performance comparison of existing representative low-light video enhancement methods.}
    \vspace{-5mm}
    \label{fig:efficiency_tradeoff}
     % \vspace{-2mm}
\end{figure}

\begin{table}[t]
\centering
\scriptsize
\setlength{\tabcolsep}{3.2pt}
\caption{Performance comparison of \sysname with representative baselines in terms of visual quality (PSNR/SSIM, $\uparrow$), temporal consistency (TSSIM, $\uparrow$), and per-frame latency (s, $\downarrow$). We also report whether a method requires paired supervision and extra motion modeling.}
\label{tab:quality}
\resizebox{\textwidth}{!}{%
\begin{tabular}{|c|l|c|c|c|c|c|c|c|c|c|}
\hline
\multicolumn{2}{|c|}{\multirow{2}{*}{\textbf{Methods}}} &
\multirow{2}{*}{\makecell[c]{\textbf{No paired video}\\\textbf{supervision}}} &
\multirow{2}{*}{\makecell[c]{\textbf{No extra motion}\\\textbf{modeling}}} &
\multirow{2}{*}{\textbf{TSSIM} $\uparrow$} &
\multicolumn{3}{c|}{\textbf{SDSD-indoor~\cite{SDSD}}} &
\multicolumn{3}{c|}{\textbf{DRV~\cite{DRVsmid}}} \\
\cline{6-11}
\multicolumn{2}{|c|}{} &  &  &  &
\textbf{PSNR} $\uparrow$ & \textbf{SSIM} $\uparrow$ & \textbf{Latency (s)} $\downarrow$ &
\textbf{PSNR} $\uparrow$ & \textbf{SSIM} $\uparrow$ & \textbf{Latency (s)} $\downarrow$ \\
\hline

\multirow{5}{*}{\makecell[c]{\textbf{Frame-wise}\\\textbf{(single-image)}\\\textbf{methods}}} &
\textbf{Zero-DCE++}~\cite{zero-dce++} &
$\checkmark$ & $\checkmark$ & -- &
14.54 & 0.51 & 0.05 &
14.47 & 0.42 & 0.06 \\
\cline{2-11}

& \textbf{LLFlow}~\cite{2022llFlow} & $\checkmark$ & $\checkmark$ & -- &
16.13 & 0.60 & 0.66 &
14.06 & 0.59 & 0.55 \\
\cline{2-11}

& \textbf{URetinex-Net}~\cite{uretinex} &
 $\checkmark$ & $\checkmark$ & -- &
15.00 & 0.55 & 1.39 &
15.43 & 0.64 & 1.39 \\
\cline{2-11}

& \textbf{Diff-retinex}~\cite{2023diffretinex} &
 $\checkmark$ & $\checkmark$ & -- &
14.36 & 0.57 & 2.86 &
16.40 & 0.67 & 2.81 \\
\cline{2-11}

& \textbf{FourierDiff}~\cite{2024fourierdiff} &
$\checkmark$ & $\checkmark$ & -- &
14.18 & 0.53 & 17.02 &
14.77 & 0.56 & 17.78 \\
\hline

\multirow{4}{*}{\makecell[c]{\textbf{Frame-wise +}\\\textbf{post-hoc temporal}\\\textbf{modeling}}} &
\textbf{MBLLEN}~\cite{MBLLEN} &
$\times$ & $\checkmark$ & 0.79 &
14.52 & 0.67 & 1.45 &
14.90 & 0.57 & 1.46 \\
\cline{2-11}

& \textbf{FastLLVE}~\cite{fastllve} &
$\times$ & $\checkmark$ & 0.68 &
14.00 & 0.64 & 1.95 &
18.05 & 0.62 & 1.68 \\
\cline{2-11}

& \textbf{StableLLVE}~\cite{stableLLVe} &
$\times$ & $\times$ & 0.75 &
15.33 & 0.63 & 0.22 &
11.78 & 0.54 & 0.21 \\
\cline{2-11}

& \textbf{AdaEnlight}~\cite{adaenlight} &
$\checkmark$ & $\times$ & 0.79 &
15.52 & 0.63 & 0.05 &
15.23 & 0.52 & 0.05 \\
\hline

\multirow{4}{*}{\makecell[c]{\textbf{Video-native}\\\textbf{methods}}} &
\textbf{SDSDNet}~\cite{SDSD} &
$\times$ & $\checkmark$ & 0.78 &
21.26 & 0.56 & 3.39 &
17.41 & 0.68 & 3.41 \\
\cline{2-11}

& \textbf{BRVE}~\cite{BRVE} &
$\times$ & $\checkmark$ & 0.76 &
15.86 & 0.52 & 2.62 &
17.12 & 0.60 & 2.15 \\
\cline{2-11}

& \textbf{RetinexMCNet}~\cite{retinexmcnet} &
$\times$ & $\checkmark$ & 0.81 &
20.84 & 0.71 & 1.99 &
18.62 & 0.64 & 1.86 \\
\cline{2-11}

& \textbf{UbiEnlight} &
$\checkmark$ & $\checkmark$ & \textbf{0.86} &
16.22 & 0.75 & 0.07 &
19.34 & 0.67 & 0.06 \\
\hline
\end{tabular}%
}
\end{table}

% BRVE~\cite{brve2024} 
% & $\times$ 
% & $\checkmark$ 
% & 0.76 
% & 15.86 
% & 0.52 
% & 2.62 
% & 17.12 
% & 0.60 
% & 2.15 \\

% RetinexMCNet 
% & $\times$ 
% & $\checkmark$ 
% & 0.81 
% & 20.84 
% & 0.71 
% & 1.99 
% & 18.62 
% & 0.64 
% & 1.86 \\

\subsection{Performance Comparison}
We evaluate \sysname on the SDSD-indoor and DRV benchmarks against frame-wise, temporal-aware, and recent video-native methods (e.g., BRVE and RetinexMCNet) under a unified deployment-oriented protocol. Besides quality and efficiency, we compare whether methods require paired supervision or explicit motion modeling (\eg optical flow). Metrics include PSNR/SSIM, TSSIM, and per-frame latency on Jetson AGX Orin, together with no-reference usability metrics (VPR@$\tau$, MGE, and MF@$\tau$) for challenging scenarios.

Results are summarized in \tabref{tab:quality}. 
\textit{First,} \sysname achieves an excellent quality-efficiency trade-off, running at only $0.06$--$0.07$\,s/frame while maintaining competitive reconstruction quality (\figref{fig:efficiency_tradeoff}). It achieves the best SSIM (0.75) on SDSD-indoor, strong PSNR (19.34) on DRV, and the highest temporal stability (TSSIM 0.86), indicating robust structural recovery with reduced flicker.
PSNR measures pixel-wise fidelity to paired references, whereas SSIM emphasizes structural similarity. 
Since \sysname is designed for spectrum-guided structural preservation and temporal consistency rather than strict pixel-level reconstruction, it achieves stronger SSIM/TSSIM while showing slightly lower PSNR than fully supervised video methods trained with paired ground truth.
\textit{Second,} \sysname achieves this quality--stability trade-off without paired video supervision in the core enhancement stage or explicit motion modeling, unlike several video-native baselines. This avoids costly alignment and reduces dependence on scarce paired video data, making \sysname more practical for mobile deployment.
Compared with high-quality diffusion baselines, \sysname is over $20\times$ faster than FourierDiff and over $25\times$ faster than SDSDNet on DRV. 
Compared with real-time methods, it improves PSNR by 4.87 over Zero-DCE++ while maintaining comparable latency.
We further evaluate two challenging scenarios: (i) dynamic illumination with intense user motion and (ii) extreme low light with intense user motion. As shown in \tabref{tab:hardcases_checklist}, \sysname is the only method satisfying all usability criteria (visibility, detail recovery, and temporal stability) in both scenarios, demonstrating reliable real-time enhancement under severe conditions.

\textbf{Summary.} \sysname offers the best overall balance of visual quality, temporal stability, efficiency, and deployability for real-time mobile and wearable video enhancement.

\begin{figure*}[t]
    \centering

    % -------------------- Row 1: NIQE (lower is better) --------------------
    \begin{subfigure}[b]{0.24\textwidth}
        \centering
        \includegraphics[width=\textwidth]{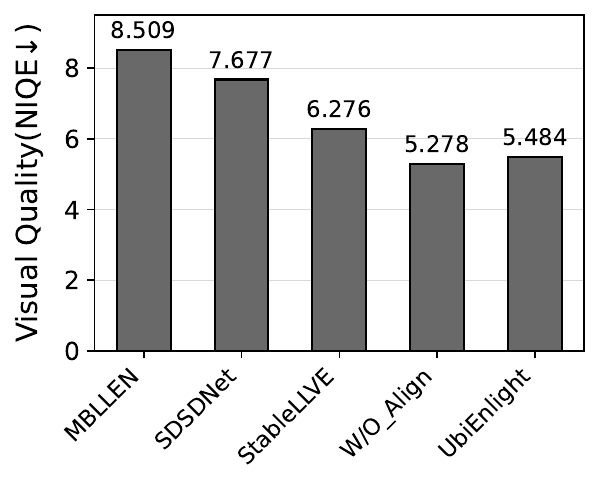}
        \caption{Moving cam, moving obj}
        \label{fig:mm_niqe}
    \end{subfigure}\hfill
    \begin{subfigure}[b]{0.24\textwidth}
        \centering
        \includegraphics[width=\textwidth]{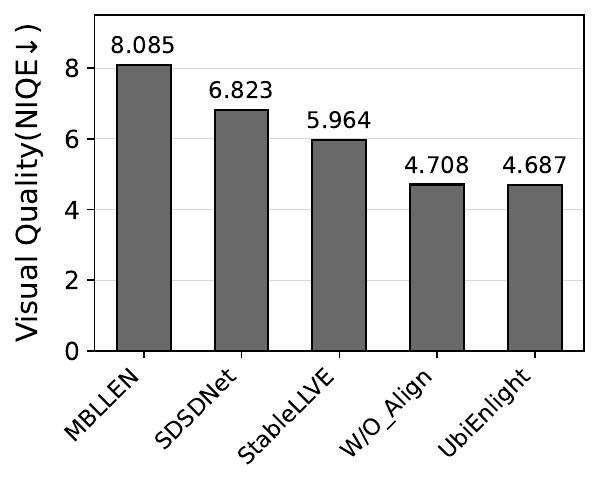}
        \caption{Moving cam, static obj}
        \label{fig:ms_niqe}
    \end{subfigure}\hfill
    \begin{subfigure}[b]{0.24\textwidth}
        \centering
        \includegraphics[width=\textwidth]{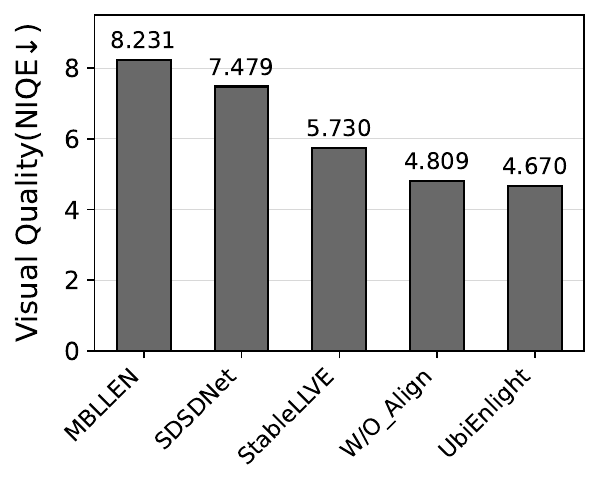}
        \caption{Static cam, moving obj}
        \label{fig:sm_niqe}
    \end{subfigure}\hfill
    \begin{subfigure}[b]{0.24\textwidth}
        \centering
        \includegraphics[width=\textwidth]{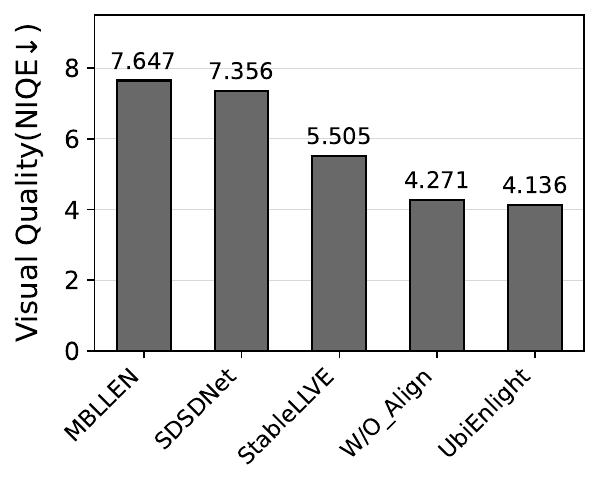}
        \caption{Static cam, static obj}
        \label{fig:ss_niqe}
    \end{subfigure}

    \vspace{0.6em}

    % -------------------- Row 2: tLPIPS (lower is better) --------------------
    \begin{subfigure}[b]{0.24\textwidth}
        \centering
        \includegraphics[width=\textwidth]{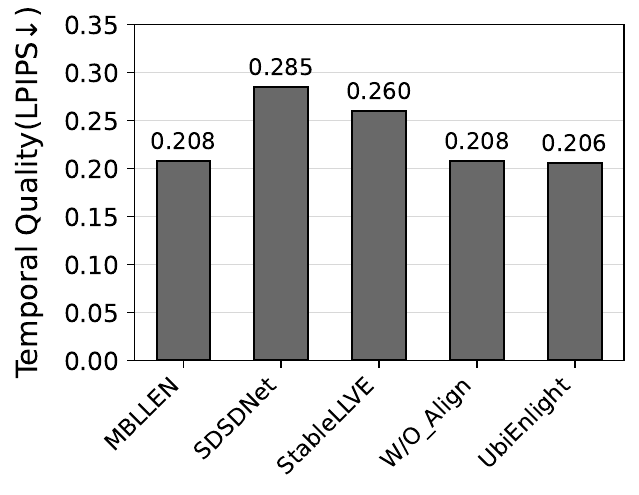}
        \caption{Moving cam, moving obj}
        \label{fig:mm_tlpips}
    \end{subfigure}\hfill
    \begin{subfigure}[b]{0.24\textwidth}
        \centering
        \includegraphics[width=\textwidth]{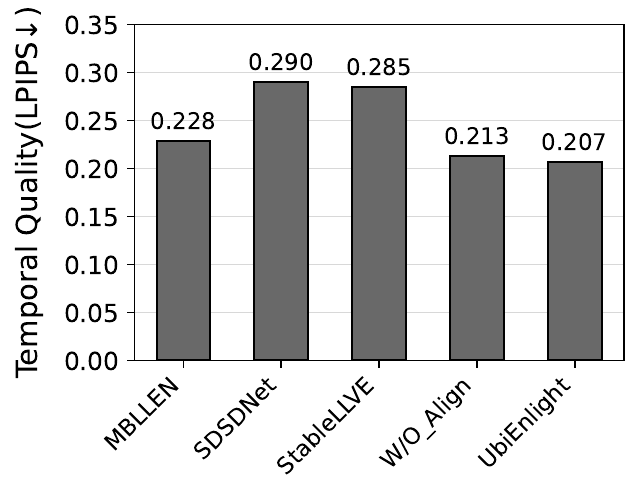}
        \caption{Moving cam, static obj}
        \label{fig:ms_tlpips}
    \end{subfigure}\hfill
    \begin{subfigure}[b]{0.24\textwidth}
        \centering
        \includegraphics[width=\textwidth]{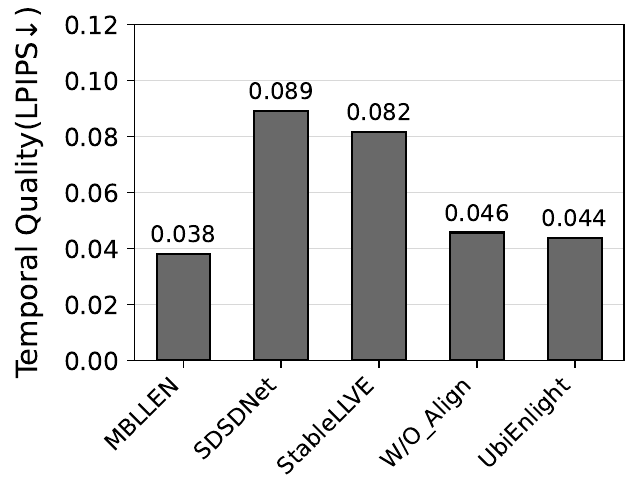}
        \caption{Static cam, moving obj}
        \label{fig:sm_tlpips}
    \end{subfigure}\hfill
    \begin{subfigure}[b]{0.24\textwidth}
        \centering
        \includegraphics[width=\textwidth]{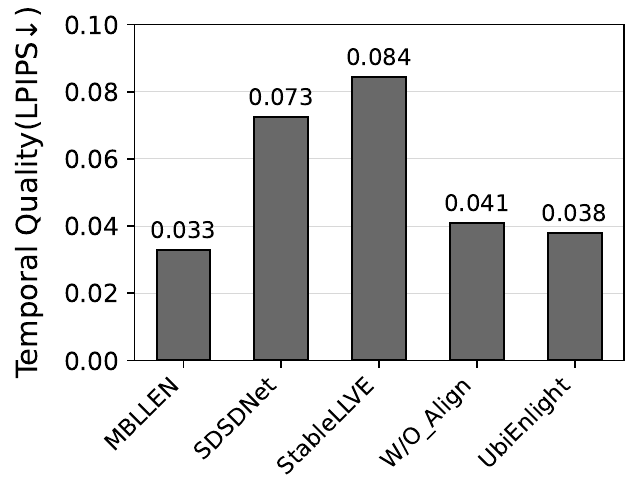}
        \caption{Static cam, static obj}
        \label{fig:ss_tlpips}
    \end{subfigure}

    \vspace{0.6em}

    % -------------------- Row 3: TSSIM (higher is better) --------------------
    \begin{subfigure}[b]{0.24\textwidth}
        \centering
        \includegraphics[width=\textwidth]{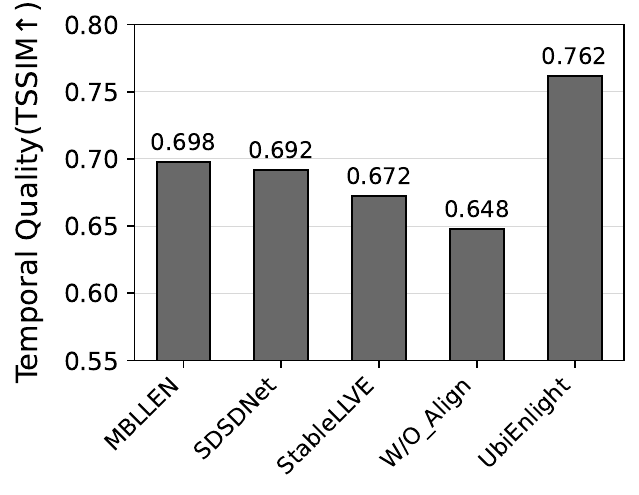}
        \caption{Moving cam, moving obj}
        \label{fig:mm_tssim}
    \end{subfigure}\hfill
    \begin{subfigure}[b]{0.24\textwidth}
        \centering
        \includegraphics[width=\textwidth]{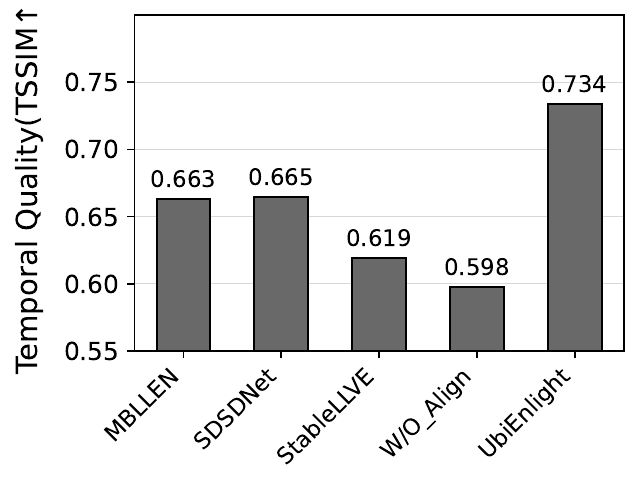}
        \caption{Moving cam, static obj}
        \label{fig:ms_tssim}
    \end{subfigure}\hfill
    \begin{subfigure}[b]{0.24\textwidth}
        \centering
        \includegraphics[width=\textwidth]{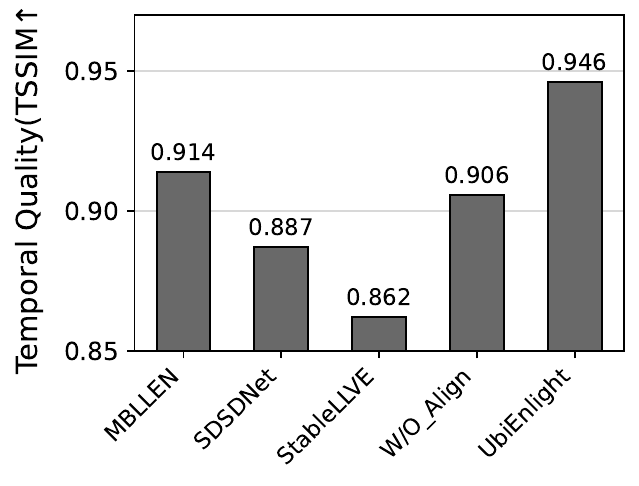}
        \caption{Static cam, moving obj}
        \label{fig:sm_tssim}
    \end{subfigure}\hfill
    \begin{subfigure}[b]{0.24\textwidth}
        \centering
        \includegraphics[width=\textwidth]{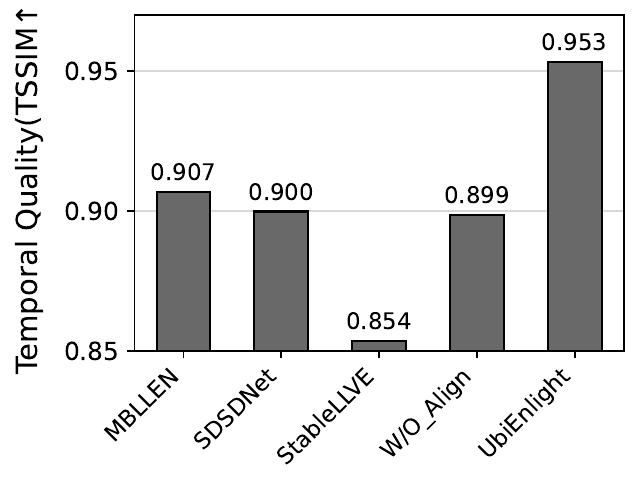}
        \caption{Static cam, static obj}
        \label{fig:ss_tssim}
    \end{subfigure}

    \caption{Temporal stability quantified by NIQE, tLPIPS, and TSSIM, in four mobile scenarios. Scenario~A: moving camera and moving objects \{\subref{fig:mm_niqe},\subref{fig:mm_tlpips},\subref{fig:mm_tssim}\}, Scenario~B: moving camera and static objects \{\subref{fig:ms_niqe},\subref{fig:ms_tlpips},\subref{fig:ms_tssim}\}, Scenario~C: static camera and moving objects \{\subref{fig:sm_niqe},\subref{fig:sm_tlpips},\subref{fig:sm_tssim}\}, and Scenario~D: static camera and static objects \{\subref{fig:ss_niqe},\subref{fig:ss_tlpips},\subref{fig:ss_tssim}\}. A low NIQE and tLPIPS and a high TSSIM indicate good perceptual quality and temporal stability.}
    \label{fig:comparison_grid}
    \vspace{-3mm}
\end{figure*}

\subsection{System Performance under Real-world Diversity and Dynamics}
We evaluate \sysname at the system level, focusing on end-to-end robustness and efficiency under diverse mobile conditions.

\subsubsection{Performance under Diverse Camera/Object Motion}
We evaluate \sysname on our self-collected \textit{UbiVideo} dataset, covering four representative mobile scenarios with different combinations of camera and object motion. 
We report NIQE (lower is better) for perceptual quality and temporal LPIPS / TSSIM (lower / higher is better) for temporal consistency. Results are shown in \figref{fig:comparison_grid}.
\textit{First,} \sysname maintains strong perceptual quality, achieving the lowest NIQE across all four scenarios (\figref{fig:mm_niqe}--\ref{fig:ss_niqe}). Although \textbf{W/O\_Align} is marginally better in a few cases, the difference is small.
\textit{Second,} \sysname consistently improves temporal consistency. It achieves the lowest temporal LPIPS in all moving-camera scenarios and remains best or near-best in static-camera settings (\figref{fig:mm_tlpips}--\figref{fig:ss_tlpips}), indicating reduced flicker and perceptual jitter.
\textit{Third,} \sysname achieves the highest TSSIM across all scenarios (\figref{fig:mm_tssim}--\figref{fig:ss_tssim}). Compared with \textbf{W/O\_Align}, it consistently improves both temporal LPIPS and TSSIM, demonstrating the effectiveness of the Spectrum-Conditioned Diffusion Transformer in stabilizing streaming enhancement.

\textbf{Summary.} Across diverse camera and object motion patterns, \sysname provides the best overall balance of perceptual quality and temporal stability, enabling robust real-time low-light video enhancement.

\subsubsection{Performance over Dynamic Energy Supply}
We evaluate \sysname under dynamic energy constraints during continuous streaming. According to the current battery level, the controller adaptively adjusts the sampling-step budget ($\theta_s$), attention reuse ratio ($\theta_a$), and keyframe refresh rate ($\theta_f$) to balance enhancement quality and system cost. Experiments are conducted on UbiVideo using Google Pixel~7 (CPU, device~1) and Jetson AGX Orin (GPU, device~2) with battery levels ranging from \(\sim\)20\% to 90\%. We report NIQE (lower is better), latency/frame, and energy/frame.
Results are summarized in \tabref{tab:energy_cpu} and \tabref{tab:energy_gpu}. As energy becomes available, \sysname progressively allocates more computation by increasing the denoising budget ($\theta_s$), reducing attention reuse ($\theta_a$), and refreshing keyframes more frequently ($\theta_f$). For example, on the GPU, $\theta_s$ increases from 4 (18\%) to 18 (94\%), $\theta_a$ decreases from $3/4$ to $1/3$, and $\theta_f$ increases from 0 to 1, enabling higher reconstruction quality. Similar adaptation is observed on the CPU, where $\theta_s$ increases from 2 (20\%) to 10 (96\%).

\textbf{Summary.} Across both platforms, \sysname continuously adapts $(\theta_s,\theta_a,\theta_f)$ according to battery availability, improving visual quality when energy is sufficient while maintaining stable latency and energy efficiency. In our continuous-use setting on Google Pixel~7, UbiEnlight depletes the battery in \(2\)$\sim$\(4.5\) hours. 
This duration is specific to the phone-class continuous-use evaluation and should not be interpreted as a uniform result across different platforms.

\begin{table}[t]
  \centering
    \small
 \caption{UbiEnlight's performance under diverse mobile energy budgets with the CPU on device 1.}
  \label{tab:energy_cpu}
  \renewcommand{\arraystretch}{1} % ↑ 行高，关键
  \resizebox{0.65\textwidth}{!}{%
  \begin{tabular}{|c|c|c|c|c|c|c|c|c|}
    \hline
    \multirow{2}{*}{\textbf{No.}} &
    \multirow{2}{*}{\makecell[c]{\textbf{Diverse energy}\\ \textbf{supply}}} &
    \multirow{2}{*}{\makecell[c]{\textbf{Power}\\ \textbf{(W)}}} &
    \multirow{2}{*}{\makecell[c]{\textbf{Energy per}\\ \textbf{frame (mJ)}}} &
    \multirow{2}{*}{\makecell[c]{\textbf{NIQE}\\ \textbf{($\downarrow$)}}} &
    \multirow{2}{*}{\makecell[c]{\textbf{Latency}\\ \textbf{per frame (ms)}} }&
    \multicolumn{3}{c|}{\makecell[c]{\textbf{Optimal}\\ \textbf{Hyperparameters}}} \\
    \cline{7-9}
    & & & & & &
    \textbf{$\theta_s$} & \textbf{$\theta_a$} & \textbf{$\theta_f$} \\
    \hline
    \textbf{1} & 20\% & 3 & 312.76 & 5.2841 & 76.47  & 2  & 1 & 0   \\
    \hline
    \textbf{2} & 43\% & 3.2 & 331.78 & 5.2607 & 103.68 & 6 & 1/3 & 0 \\
    \hline
    \textbf{3} & 68\% & 3.5 & 420.18 &  5.2558 & 120.05 & 10 & 1/2 & 1/5 \\

    \hline
    \textbf{4} & 74\% & 3.7 & 472.60 & 5.2368& 127.73 & 4 & 1/2 & 3/5 \\
    \hline
    \textbf{5} & 96\% & 4.2 & 639.88 & 5.1859 & 159.97 & 10 & 1/5 & 1   \\
    \hline
  \end{tabular}%
  }
\end{table}

\begin{table}[t]
  \centering
   \caption{UbiEnlight's performance under diverse mobile energy budgets with the GPU on device 2.}
  \label{tab:energy_gpu}
  \small
  \renewcommand{\arraystretch}{1} % ↑ 行高，关键
  \resizebox{0.6\textwidth}{!}{%
  \begin{tabular}{|c|c|c|c|c|c|c|c|c|}
    \hline
    \multirow{2}{*}{\textbf{No.}} &
    \multirow{2}{*}{\makecell[c]{\textbf{Diverse energy}\\ \textbf{supply}}} &
    \multirow{2}{*}{\makecell[c]{\textbf{Power}\\ \textbf{(W)}}} &
    \multirow{2}{*}{\makecell[c]{\textbf{Energy per}\\ \textbf{frame (J)}}} &
    \multirow{2}{*}{\makecell[c]{\textbf{NIQE}\\ \textbf{($\downarrow$)}}} &
    \multirow{2}{*}{\makecell[c]{\textbf{Latency}\\ \textbf{per frame (ms)}}} &
    \multicolumn{3}{|c|}{\makecell[c]{\textbf{Optimal}\\ \textbf{hyperparameters}}} \\
    \cline{7-9}
    & & & & & &
    \textbf{$\theta_s$} & \textbf{$\theta_a$} & \textbf{$\theta_f$} \\
    \hline
    \textbf{1} & 18\% & 6.21 & 0.8379 & 4.7725 & 66.16  & 4  & 3/4 & 0   \\
    \hline
    \textbf{2} & 33\% & 6.54 & 0.8473 & 4.7708 & 68.33 & 10 & 1/5 & 1/5 \\
    \hline
    \textbf{3} & 58\% & 6.79 & 0.8691 & 4.7561 & 70.49 & 16 & 1/4 & 1/5 \\
    \hline
    \textbf{4} & 78\% & 6.84 & 0.8951 & 4.7530 & 73.31 & 12 & 1/4 & 1/2 \\
    \hline
    \textbf{5} & 94\% & 6.88 & 0.9144 & 4.7509 & 75.51 & 18 & 1/3 & 1   \\
    \hline
  \end{tabular}%
  }
\end{table}

\subsubsection{Performance across Diverse Video Frame Rates}
\begin{figure*}[t]
    \centering
    \begin{subfigure}[b]{0.24\textwidth}
        \centering
        \includegraphics[width=\textwidth]{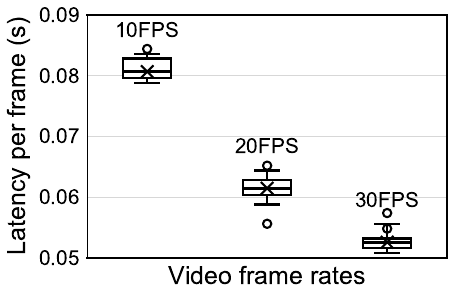}
        \caption{Latency per frame}
        \label{fig:fps_latency_frame}
    \end{subfigure}\hfill
    \begin{subfigure}[b]{0.25\textwidth}
        \centering
        \includegraphics[width=\textwidth]{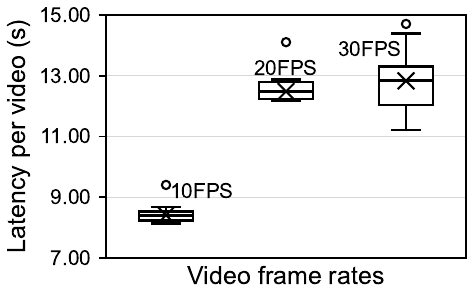}
        \caption{Latency per video}
        \label{fig:fps_latency_video}
    \end{subfigure}\hfill
    \begin{subfigure}[b]{0.24\textwidth}
        \centering
        \includegraphics[width=\textwidth]{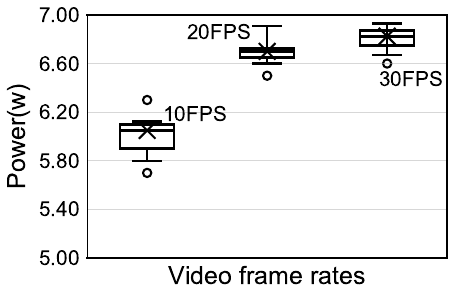}
        \caption{Power}
        \label{fig:fps_power}
    \end{subfigure}\hfill
    \begin{subfigure}[b]{0.24\textwidth}
        \centering
        \includegraphics[width=\textwidth]{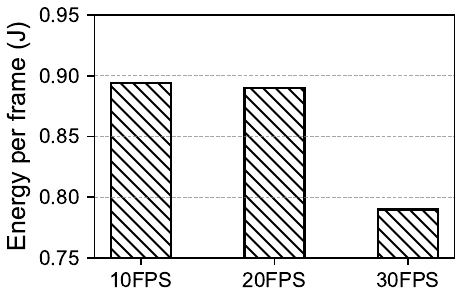}
        \caption{Energy cost per frame}
        \label{fig:fps_energy_frame}
    \end{subfigure}

    \caption{\sysname’s performance across diverse inputs with varying video frame rates (FPS), \emph{i.e.}, 10 FPS, 20FPS, and 30FPS.}
    \label{fig:fps_adaptation}
\end{figure*}

We distinguish \emph{input frame rate} from \emph{enhancement throughput}. The former is the source video rate, whereas the latter is the number of enhanced frames produced per second under the selected runtime configuration, which may be lower than the input rate on resource-constrained devices.
We evaluate whether \sysname maintains responsive streaming enhancement under varying input frame rates (10, 20, and 30 FPS), which commonly arise from different camera settings, scene complexity, and runtime contention. Experiments are conducted on 20 LoLi-Phone clips (10\,s each) using Jetson AGX Orin (GPU, device~2). In adaptive mode, \sysname dynamically adjusts sampling steps, attention reuse, and frame caching according to the input rate. We report per-frame latency, clip-level processing time, power, and energy/frame (\figref{fig:fps_adaptation}).
Results show that \sysname scales efficiently with increasing input rates. As FPS increases from 10 to 30, per-frame latency decreases from 0.081\,s to 0.053\,s, indicating more aggressive computation reuse under higher temporal redundancy. Although clip-level processing time increases with more input frames, its growth remains sublinear (\figref{fig:fps_latency_video}). Meanwhile, power rises only modestly (5.9\,W$\rightarrow$6.8\,W), and the 30\,FPS setting achieves the lowest energy/frame (\figref{fig:fps_energy_frame}).

\textbf{Summary.} \sysname adapts effectively to diverse input frame rates, sustaining responsive streaming while reducing per-frame latency and energy/frame through adaptive computation reuse.

% \textbf{Summary.} \sysname adapts effectively to changing frame rates, preserving real-time behavior with bounded delay, sublinear clip-latency growth, and improved energy/frame at high FPS under only modest power increase.

\subsubsection{Performance on Diverse Mobile Platforms}
\begin{figure}[t]
    \centering
    \includegraphics[width=0.6\linewidth]{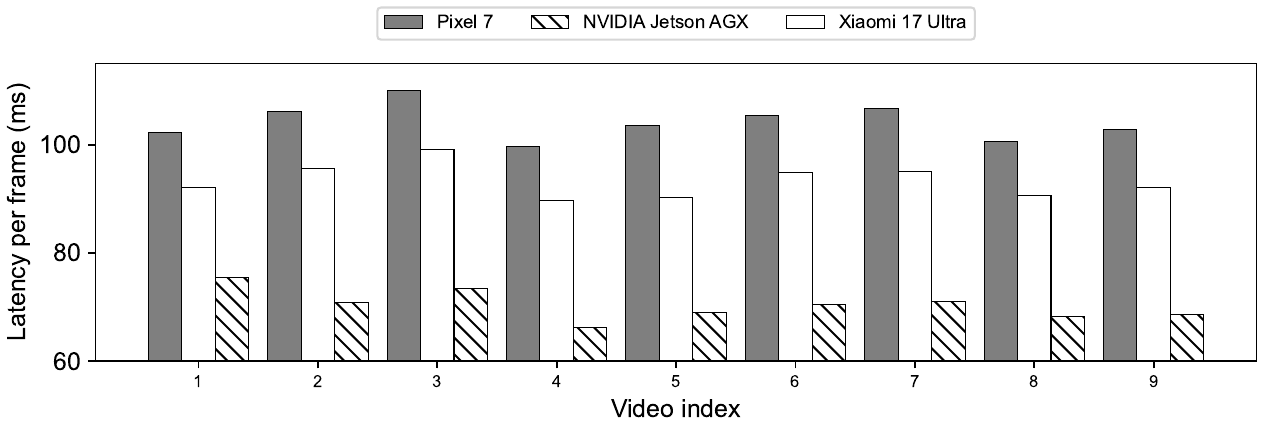}
    \vspace{-3mm}
    \caption{Performance across three typical camera-embedded platforms.}
    \vspace{-3mm}
    \label{fig:latency_comparison}
\end{figure}

We evaluate \sysname on three representative camera-enabled platforms with different compute budgets and deployment stacks: Google Pixel~7 (CPU; NNAPI, device~1), NVIDIA Jetson AGX Orin (GPU; PyTorch, device~2), and Xiaomi~17 Ultra (CPU; PyTorch Mobile, device~3). 
Using eight 30\,s UbiVideo clips, we measure per-frame enhancement latency (Fig.~\ref{fig:latency_comparison}).
Across all platforms, \sysname consistently delivers low-latency diffusion enhancement. On the phone-class platforms (Pixel~7 and Xiaomi~17 Ultra), selected enhanced frames are processed in \(90\)$\sim$\(100\) ms, while Jetson AGX Orin further reduces latency to \(\sim66\) ms, enabling higher effective throughput. Here, ``near-real-time'' refers to maintaining a bounded end-to-end display latency through the adaptive runtime controller, rather than enhancing every input frame at the raw input frame rate. These results demonstrate stable and practical deployment across heterogeneous mobile and edge platforms.

\subsection{Ablation Studies}
Unless otherwise stated, each ablation is conducted on the platform that best exposes the effect of the mechanism under study. We use Jetson AGX Orin (GPU;device 2) for ablations that require stable measurement over a wider quality--latency operating range, and Google Pixel~7 (CPU;device 1)) for ablations that emphasize lightweight runtime efficiency on a practical mobile device. Therefore, these studies are intended to validate the trend of each control knob rather than to compare absolute runtime values across different hardware platforms.

\subsubsection{Impact of Keyframe Selection and Frame Caching}
\begin{figure}[t]
    \centering
    \begin{subfigure}[b]{0.26\linewidth}
        \centering
        \includegraphics[width=\linewidth]{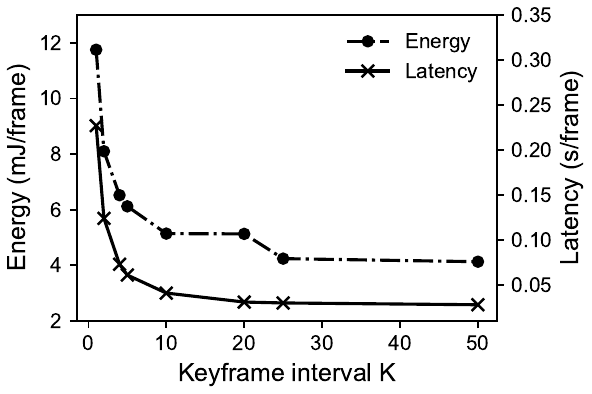}
        \caption{ keyframe interval $K$}
        \label{fig:keyframe_latency_energy}
    \end{subfigure}\hspace{3mm}
    \begin{subfigure}[b]{0.23\linewidth}
        \centering
        \includegraphics[width=\linewidth]{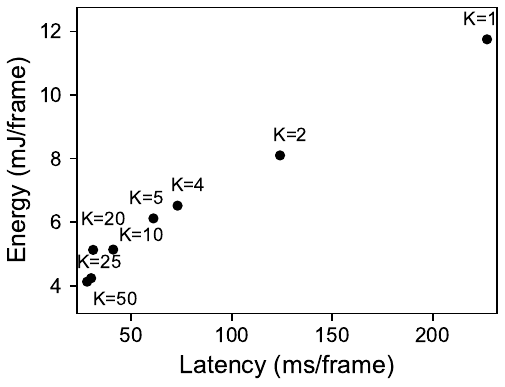}
        \caption{Latency--energy tradeoff}
        \label{fig:keyframe_pareto}
    \end{subfigure}\hspace{3mm}
    \begin{subfigure}[b]{0.31\linewidth}
        \centering
        \includegraphics[width=\linewidth]{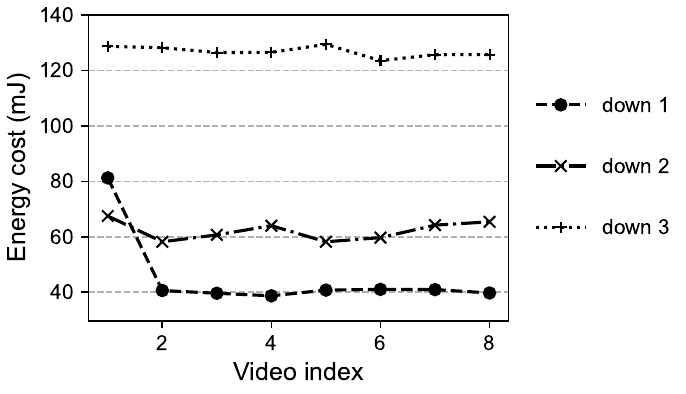} 
        \caption{Impact of frame resolution}
        \label{fig:frame_resolution}
    \end{subfigure}
    \caption{Micro-benchmark of key system parameters. }
    \label{fig:keyframe_cache_resolution}
\end{figure}

We test the efficiency of \sysname's keyframe selection and frame caching. Using LoLi-Phone, we vary the keyframe interval \(K\) from enhancing every frame (\(K{=}1\)) to enhancing one keyframe every \(K\) frames, while keeping all other settings fixed. Experiments are conducted on Jetson AGX Orin (GPU; device~2), reporting per-frame latency and energy (\figref{fig:keyframe_cache_resolution}).
As shown in \figref{fig:keyframe_latency_energy}, increasing \(K\) substantially reduces both latency and energy, with the largest gains at small intervals before gradually saturating. The latency--energy trade-off (\figref{fig:keyframe_pareto}) further shows that intermediate intervals (\eg \(K{=}10\)--\(50\)) achieve most of the efficiency benefits, while larger intervals provide only marginal improvements.
\textbf{Summary.} Keyframe selection and frame caching provide an effective runtime knob for reducing latency and energy while preserving streaming efficiency.

\subsubsection{Impact of Frame Resolution}
We study how input resolution affects \sysname’s on-device energy cost on Google Pixel~7 (CPU; device~1) using Ubivideo. \figref{fig:frame_resolution} reports per-frame energy under three settings: full resolution, down-sampled to \(1/2\), and down-sampled to \(1/3\).
Results show a monotonic trend, \ie higher resolutions consistently incur higher energy across all clips, while down-sampling substantially reduces per-frame energy. \textbf{Summary}.
This indicates that resolution scaling provides a simple, effective knob to tune \sysname’s energy demand under resource constraints.

% ---------------- Figure 1: sampling steps ----------------
\begin{figure*}[t]
  \centering
  \newcommand{\subW}{0.32\linewidth}
  \newcommand{\subH}{2.8cm}
  \makebox[\textwidth][c]{%
    \begin{subfigure}[t]{\subW}
      \centering
      \includegraphics[height=\subH,width=\linewidth,keepaspectratio]{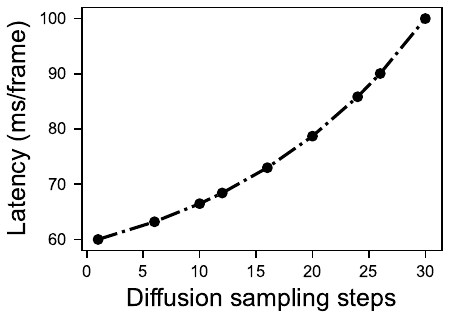}
      \caption{Latency \textit{vs.}\ sampling steps}
      \label{fig:ab_step_latency}
    \end{subfigure}\hspace{0.00\linewidth}
    \begin{subfigure}[t]{\subW}
      \centering
      \includegraphics[height=\subH,width=\linewidth,keepaspectratio]{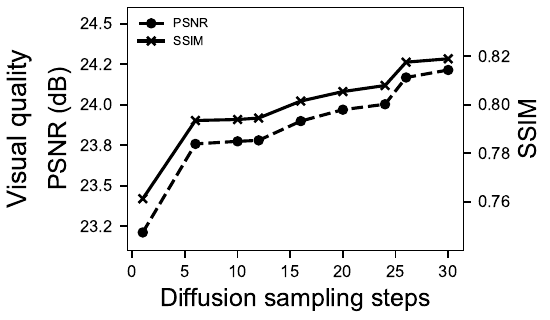}
      \caption{Quality \textit{vs.}\ sampling steps}
      \label{fig:ab_step_quality}
    \end{subfigure}
  }
  \caption{Benchmark of diffusion sampling steps on system performance.}
  \label{fig:ablation_steps}
\end{figure*}

\subsubsection{Impact of Diffusion Sampling Steps}
We evaluate the impact of diffusion sampling steps on the quality--latency trade-off of \sysname. Using DID, we vary the sampling-step budget while keeping all other components fixed, and measure per-frame latency and PSNR/SSIM on NVIDIA Jetson AGX Orin (GPU; device~2). Results are shown in \figref{fig:ablation_steps}.
Latency increases monotonically with the number of sampling steps (\figref{fig:ab_step_latency}), from \(\sim\)60 ms/frame to \(\sim\)100 ms/frame, with a steeper increase at higher step counts. Meanwhile, reconstruction quality improves with diminishing returns (\figref{fig:ab_step_quality}): PSNR increases from 23 dB to 24 dB and SSIM from 0.7 to \(\sim\)0.8, while the incremental gains gradually diminish.
\textbf{Summary.} Sampling steps provide an effective quality-latency control knob. We therefore adopt 10$\sim$20 steps by default to balance reconstruction quality and real-time throughput.

\subsubsection{Impact of Attention Reuse}

\begin{figure*}[t]
  \centering
  \newcommand{\subW}{0.32\linewidth}
  \newcommand{\subH}{2.8cm}
  \makebox[\textwidth][c]{%
    \begin{subfigure}[t]{\subW}
      \centering
      \includegraphics[height=\subH,width=\linewidth,keepaspectratio]{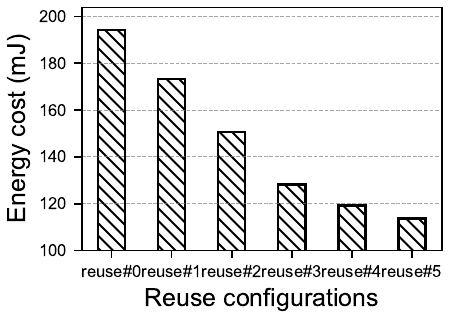}
      \caption{Energy vs.\ attention reuse}
      \label{fig:ab_reuse_energy}
    \end{subfigure}\hspace{0.01\linewidth}
    \begin{subfigure}[t]{\subW}
      \centering
      \includegraphics[height=\subH,width=\linewidth,keepaspectratio]{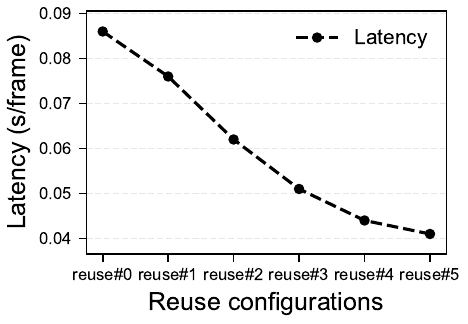}
      \caption{Latency vs.\ attention reuse}
      \label{fig:ab_reuse_latency}
    \end{subfigure}\hspace{0.01\linewidth}
    \begin{subfigure}[t]{\subW}
      \centering
      \includegraphics[height=\subH,width=\linewidth,keepaspectratio]{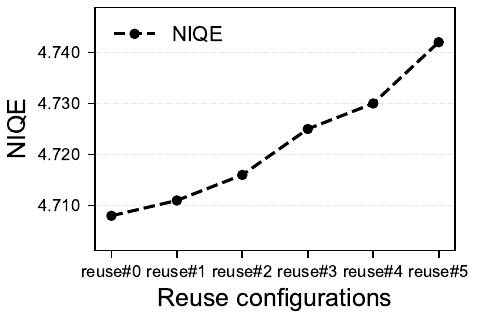}
      \caption{NIQE vs.\ attention reuse}
      \label{fig:ab_reuse_niqe}
    \end{subfigure}
  }
  \caption{Benchmark of attention reuse on \sysname's efficiency and visual quality, \ie (a) per-frame energy cost, (b) per-frame latency, and (c) visual quality under different reuse configurations.}
  \vspace{-3mm}
  \label{fig:ablation_reuse}
\end{figure*}

We evaluate attention reuse as a way to remove redundant computation across similar neighboring diffusion steps and frames.
On Google Pixel~7 (CPU; device~1), we fix the diffusion budget to 20 steps and compare six policies: \textit{reuse\#0} recomputes attention at every step, while \textit{reuse\#1--\#5} reuse cached attention outputs every 2/4/6/8/10 steps.
We also report NIQE on LoLi-Phone to measure perceptual quality.
As shown in \figref{fig:ablation_reuse}, attention reuse improves efficiency with little quality loss.
Energy decreases from 194.1\,mJ to 113.6\,mJ per frame, achieving a 41.5\% reduction.
Latency drops from 86\,ms/frame to 41\,ms/frame, yielding a 52.3\% reduction and a 2.1$\times$ speedup.
Meanwhile, NIQE changes only slightly from 4.708 to 4.742, suggesting negligible perceptual degradation.
\textbf{Summary.}
Attention reuse removes redundant diffusion computation and substantially improves on-device latency and energy efficiency while preserving visual quality.

% We evaluate the benefit of reusing attention outputs across diffusion steps, motivated by the high similarity of attention activations between neighboring steps/frames, which makes full recomputation largely redundant.
% We run \sysname on Google Pixel~7 (CPU; device~1) with a fixed diffusion setting of 30 sampling steps, and compare six reuse policies: \textit{reuse\#0} recomputes attention at every step, while \textit{reuse\#1--\#5} reuse cached attention outputs every 2/4/6/8/10 steps, respectively. Since \sysname already achieves competitive quality without reuse (\figref{fig:comparison_grid}), we focus on latency and energy. Results are shown in \ref{fig:ablation_reuse}.
% \textit{First,} attention reuse reduces energy monotonically (\figref{fig:ab_reuse_energy}): per-frame energy drops from 290\,mJ (reuse\#0) to 170\,mJ (reuse\#5), a 41\% reduction.
% \textit{Second,} reuse also cuts latency (\figref{fig:ab_reuse_latency}) from 130\,ms/frame (reuse\#0) to 60\,ms/frame (reuse\#5), a 54\% reduction (\(\sim\)2.2$\times$ speedup).
% \textbf{Summary.} Reusing attention across diffusion steps is an effective mechanism to remove redundant computation, substantially improving latency and energy efficiency on-device.

\subsubsection{Impact of Spectrum Conditioning}
We evaluate the effect of spectrum conditioning by comparing the full model with two variants under two representative scenarios: moving camera with moving objects and static camera with static objects. Specifically, \textit{UbiEnlight w/o Spectrum Prior} removes the amplitude and phase priors, while \textit{UbiEnlight w/o Align} removes the second-stage Diffusion Transformer.
As shown in Table~\ref{tab:spectrum_ablation}, removing the spectrum prior causes the largest performance degradation, increasing NIQE to 9.214/8.736 and reducing TSSIM to 0.621/0.772 across the two scenarios. This indicates that spectrum conditioning provides stable guidance for diffusion by jointly leveraging amplitude cues for illumination adjustment and phase cues for structural anchoring, rather than relying on diffusion alone. Removing the second-stage Diffusion Transformer mainly reduces temporal stability, while the full model achieves the highest TSSIM (0.762/0.953) and the best NIQE in the static scenario.
\textbf{Summary.}
Spectrum conditioning provides stable guidance for diffusion-based low-light video enhancement, with amplitude improving illumination restoration and phase preserving structural consistency under motion and illumination changes.

% ---------------- Figure 2: attention reuse ----------------
% \begin{figure*}[t]
%   \centering
%   \newcommand{\subW}{0.32\linewidth}
%   \newcommand{\subH}{2.8cm}
%   \makebox[\textwidth][c]{%
%     \begin{subfigure}[t]{\subW}
%       \centering
%       \includegraphics[height=\subH,width=\linewidth,keepaspectratio]{figure/energy_reuse.pdf}
%       \caption{Energy vs.\ attention reuse}
%       \label{fig:ab_reuse_energy}
%     \end{subfigure}\hspace{-0.02\linewidth}
%     \begin{subfigure}[t]{\subW}
%       \centering
%       \includegraphics[height=\subH,width=\linewidth,keepaspectratio]{figure/latency_reuse.pdf}
%       \caption{Latency vs.\ attention reuse}
%       \label{fig:ab_reuse_latency}
%     \end{subfigure}
%   }
%   \caption{Benchmark of attention reuse on \sysname's efficiency, \ie (a) per-frame energy cost under different reuse configurations and (b) per-frame latency under different reuse configurations.}
%   \vspace{-3mm}
%   \label{fig:ablation_reuse}
% \end{figure*}

% ---------------- tab  ----------------
\begin{table}[t]
\centering
\tiny
\caption{Ablation study of the diffusion-stage design under two representative scenarios.}
\label{tab:spectrum_ablation}
\begin{tabular}{|c|p{1.7cm}|p{1.7cm}|p{1.7cm}|p{1.7cm}|}
\hline
\multirow{2}{*}{Method}
& \multicolumn{2}{c|}{Moving cam., moving obj.}
& \multicolumn{2}{c|}{Static cam., static obj.} \\
\cline{2-5}
& \centering NIQE$\downarrow$ 
& \centering TSSIM$\uparrow$ 
& \centering NIQE$\downarrow$ 
& \centering\arraybackslash TSSIM$\uparrow$ \\
\hline
UbiEnlight w/o Spectrum Prior
& \centering 9.214 & \centering 0.621 & \centering 8.736 & \centering\arraybackslash 0.772 \\
\hline
UbiEnlight w/o Align
& \centering \textcolor{red}{5.278} & \centering 0.648 & \centering 4.271 & \centering\arraybackslash 0.899 \\
\hline
UbiEnlight
& \centering 5.484 & \centering \textcolor{red}{0.762} & \centering \textcolor{red}{4.136} & \centering\arraybackslash \textcolor{red}{0.953} \\
\hline
\end{tabular}
\end{table}

\begin{figure*}[t]
    \centering
    \subfloat[Perceived degradations]{
        \includegraphics[width=0.26\textwidth]{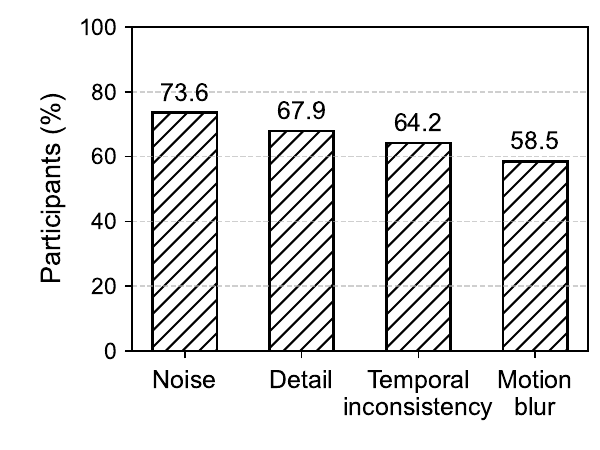}
        \label{fig:user_deg}
    }
    \hspace{3mm}
    \subfloat[Most important factor]{
        \includegraphics[width=0.26\textwidth]{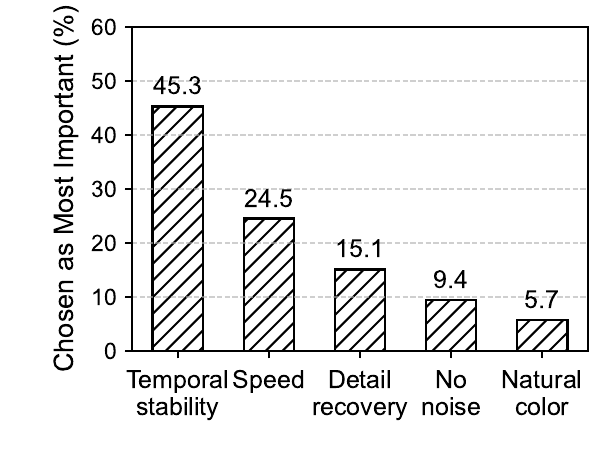}
        \label{fig:user_factor}
    }
    \hspace{3mm}
    \subfloat[Latency acceptance]{
        \includegraphics[width=0.26\textwidth]{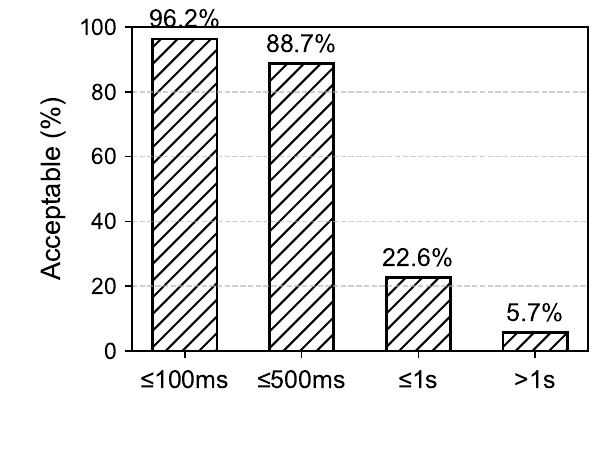}
        \label{fig:user_latency}
    }
    \caption{User study results with 53 participants.}
    \vspace{-3mm}
    \label{fig:userstudy}
\end{figure*}

\begin{table}[t]
\centering
\tiny
\setlength{\tabcolsep}{5pt}
\caption{Subjective evaluation of three operating modes.}
\label{tab:user_pref_modes}
\begin{tabular}{|c|c|c|c|c|c|}
\hline
\textbf{Mode} &
\makecell[c]{\textbf{Temporal}\\\textbf{Score}} &
\makecell[c]{\textbf{Responsiveness}\\\textbf{Score}} &
\makecell[c]{\textbf{Detail}\\\textbf{Score}} &
\makecell[c]{\textbf{Perceptual Utility}\\\(\boldsymbol{U_{\text{pref}}}\)} &
\makecell[c]{\textbf{Preference}\\\textbf{Ratio}} \\
\hline
\makecell[c]{\textbf{Fixed High-Quality}} &
4.16 & 2.23 & \textbf{4.68} & 3.37 & 5/20 (25.0\%) \\
\hline
\makecell[c]{\textbf{Fixed Low-Cost}} &
2.84 & \textbf{4.51} & 2.37 & 2.96 & 2/20 (10.0\%) \\
\hline
\makecell[c]{\textbf{Adaptive Mode}} &
\textbf{4.57} & 4.42 & 4.19 & \textbf{3.92} & \textbf{13/20 (65.0\%)} \\
\hline
\end{tabular}
\end{table}

\subsection{User Study}

We conduct a two-part user study to understand user requirements for low-light mobile video enhancement and evaluate whether \sysname's adaptive design improves perceived usability. 
The study includes (i) a 53-participant survey on user needs and (ii) a 20-participant comparison of three operating modes.

\begin{figure*}[t]
    \centering
    \includegraphics[width=0.5\textwidth]{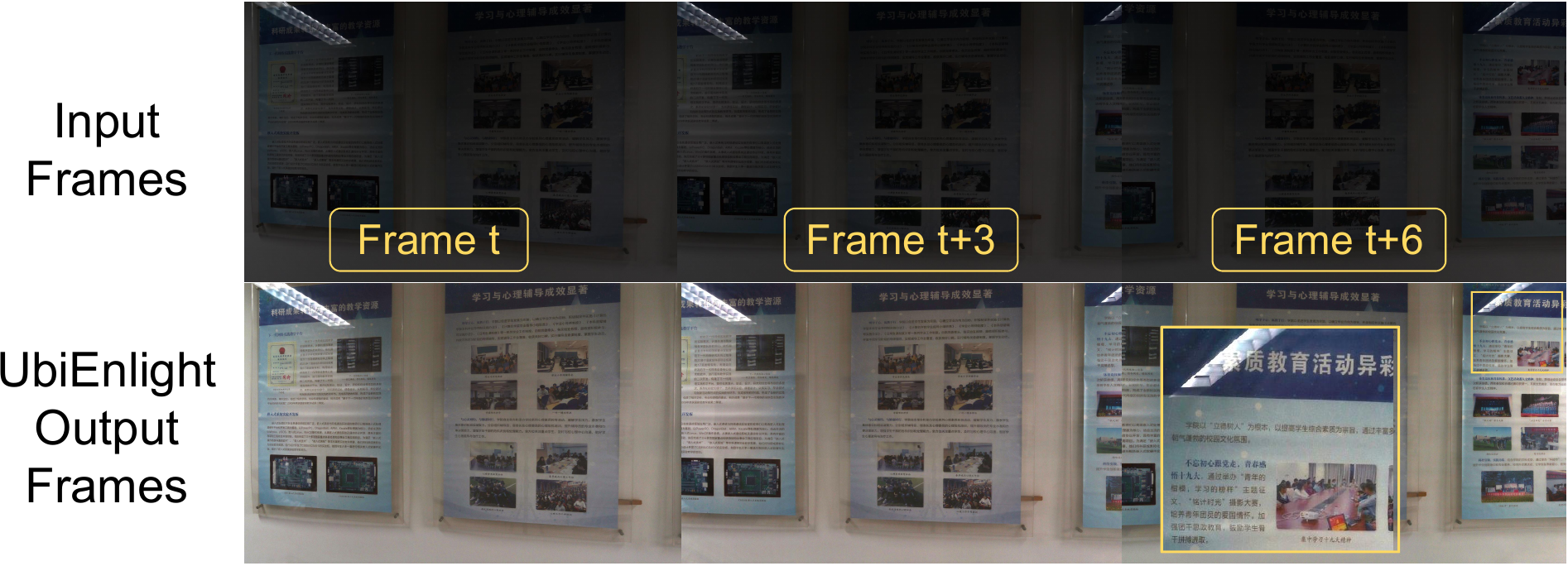}
    \caption{\textbf{Qualitative results on continuous low-light video frames.} UbiEnlight improves dark-region visibility and text details at Frame~$t$, $t{+}3$, and $t{+}6$, while maintaining temporally consistent illumination.}
    \vspace{-3mm}
    \label{fig:visual_base}
\end{figure*}

\textbf{User requirements.}
We first surveyed 53 participants (age 21--68; 29 male/24 female) who regularly capture or view low-light videos on mobile or wearable devices. The survey covered usage scenarios, perceived degradations, and preferences regarding latency, energy consumption, and adaptive behavior.
Results (\figref{fig:userstudy}) show that low-light video enhancement is both common and safety-critical. Participants frequently encountered low-light scenarios, including night commuting (69.8\%), home monitoring (60.4\%), and outdoor activities (52.8\%), while reporting visible noise (73.6\%), detail loss (67.9\%), temporal inconsistency (64.2\%), and motion blur (58.5\%). These degradations affected usability (90.6\%), safety (84.9\%), and viewing comfort (81.1\%). Users consistently prioritized temporal stability (45.3\%) over latency (24.5\%) and detail (15.1\%); 71.7\% considered flicker more objectionable than minor frame artifacts, 96.2\% accepted delays below 100\,ms, and 79.2\% preferred on-device enhancement for responsiveness and privacy. These findings directly motivate the user-centric adaptive design of \sysname.
Visual examples (\figref{fig:visual_base}) further illustrate that \sysname improves dark-region visibility while maintaining stable illumination and reducing flicker across consecutive frames.

\textbf{Adaptive mode evaluation.}
We further recruited 20 participants (age 22--65; 11 male/9 female) from the original cohort to compare three operating modes: \textit{Fixed High-Quality}, \textit{Fixed Low-Cost}, and \textit{Adaptive Mode}. Participants evaluated 12 representative low-light clips (5--8\,s) covering dynamic illumination, fast motion, handheld shake, and static high-detail scenes.
To reduce ordering bias, both video clips and operating modes were randomized and counterbalanced across participants. As this study aims to validate perceptual benefits rather than provide a fully powered statistical analysis, we report descriptive subjective scores, preference ratios, and normalized perceptual utility.
Participants rated temporal stability, responsiveness, detail preservation, and overall preference on five-point scales. Based on the preference survey, we define a preference-weighted perceptual utility
$U_{\text{pref}}=0.453S+0.245R+0.151D$,
where \(S\), \(R\), and \(D\) denote normalized ratings for temporal stability, responsiveness, and detail preservation. We further define the unit-energy perceptual utility
$\mathrm{UVE}=\frac{U_{\text{pref}}}{\mathrm{Energy}}$
to measure perceived utility per unit energy.
Table~\ref{tab:user_pref_modes} summarizes the results. The \textit{Adaptive Mode} achieves the highest overall utility (\(U_{\text{pref}}=3.92\)), outperforming \textit{Fixed High-Quality} (3.37) and \textit{Fixed Low-Cost} (2.96), while receiving the highest user preference (65.0\%). Using measured energy on Google Pixel~7, it also improves \(\mathrm{UVE}\) by 61.3\% and 32.7\% over the two fixed modes, respectively.

\textbf{Summary.}
The results show that \sysname adaptively allocates computation toward the aspects users value most, \eg temporal stability, responsiveness, and usable detail, thereby improving perceived quality without sacrificing energy efficiency.

%
% \begin{figure*}[t]
%     \centering
%     \subfloat[Perceived degradations]{
%         \includegraphics[height=0.18\textheight]{figure/userstudy_degradation.pdf}
%         \label{fig:user_deg}
%     }
%     \hfill
%     \subfloat[Most important factor]{
%         \includegraphics[height=0.18\textheight]{figure/userstudy_priority.pdf}
%         \label{fig:user_factor}
%     }
%     \hfill
%     \subfloat[Latency acceptance]{
%         \includegraphics[height=0.18\textheight]{figure/latency_acceptance_bar.pdf}
%         \label{fig:user_latency}
%     }
%     \caption{User study results on (a) perceived degradations, (b) priority factors, and (c) latency acceptance.}
%     \vspace{-6pt}
%     \label{fig:userstudy_three}
% \end{figure*}

\subsection{Real-world Case Study}
\label{sec:case-study}

\begin{figure*}[t]
    \centering
    % Use a fixed height to keep all subfigures visually aligned.
    \subfloat[Museum/Night-blind]{
        \includegraphics[height=2.4cm]{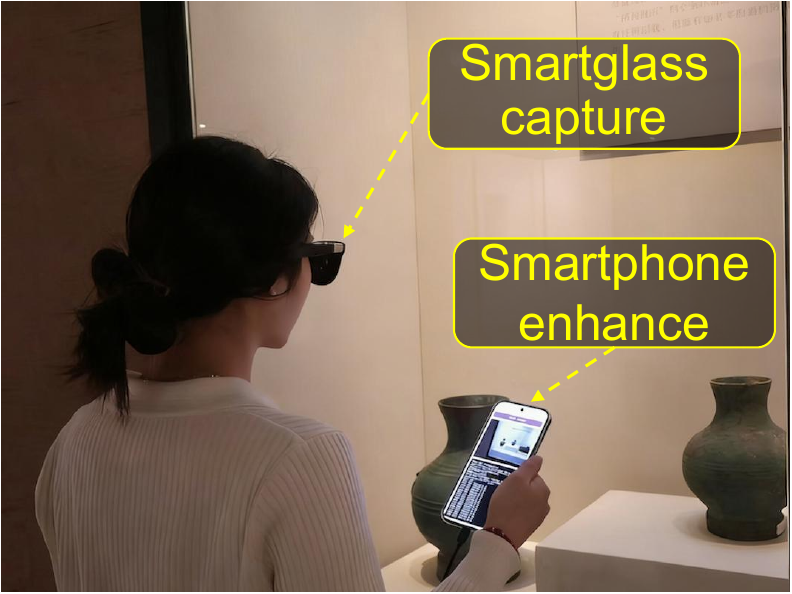}
    }\hfill
    \subfloat[In-vehicle capture]{
        \includegraphics[height=2.4cm]{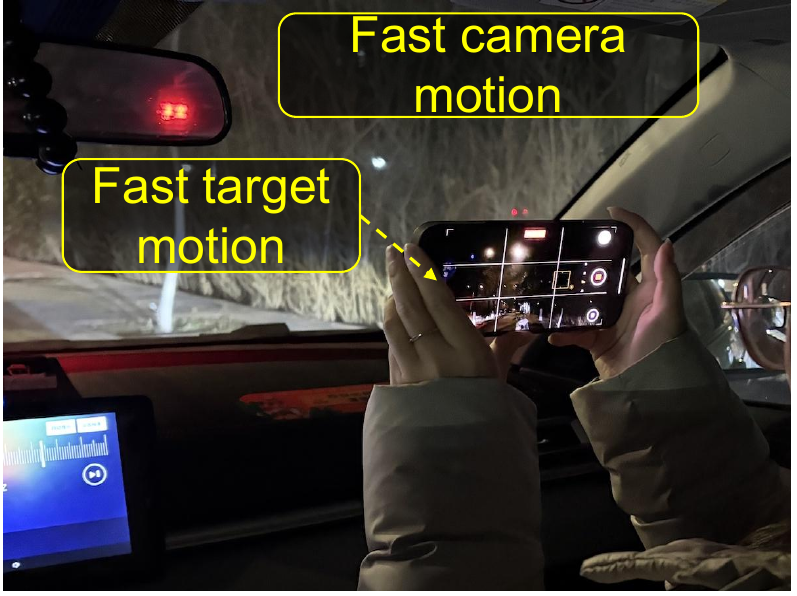}
    }\hfill
    \subfloat[Night e-bike ]{
        \includegraphics[height=2.4cm]{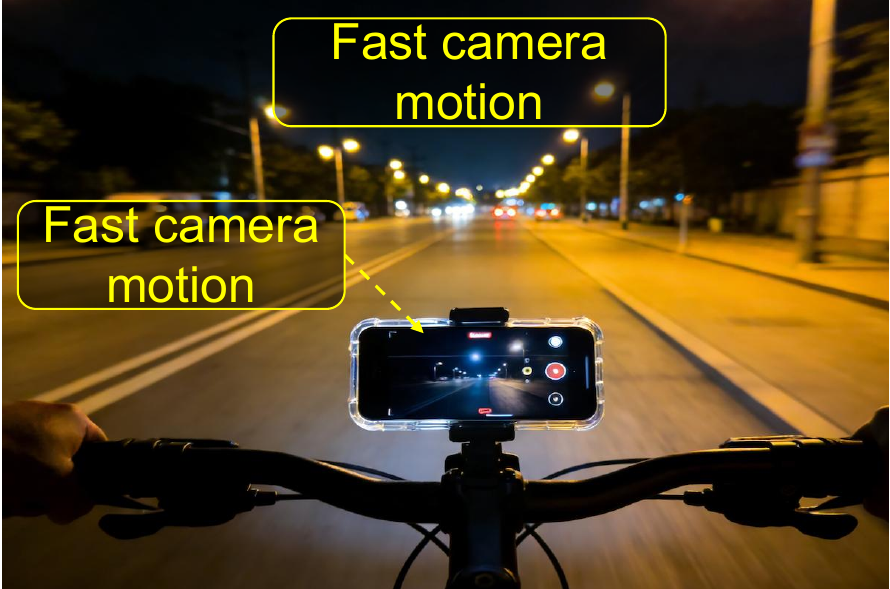}
    }\hfill
    \subfloat[Handheld walk/run ]{
        \includegraphics[height=2.4cm]{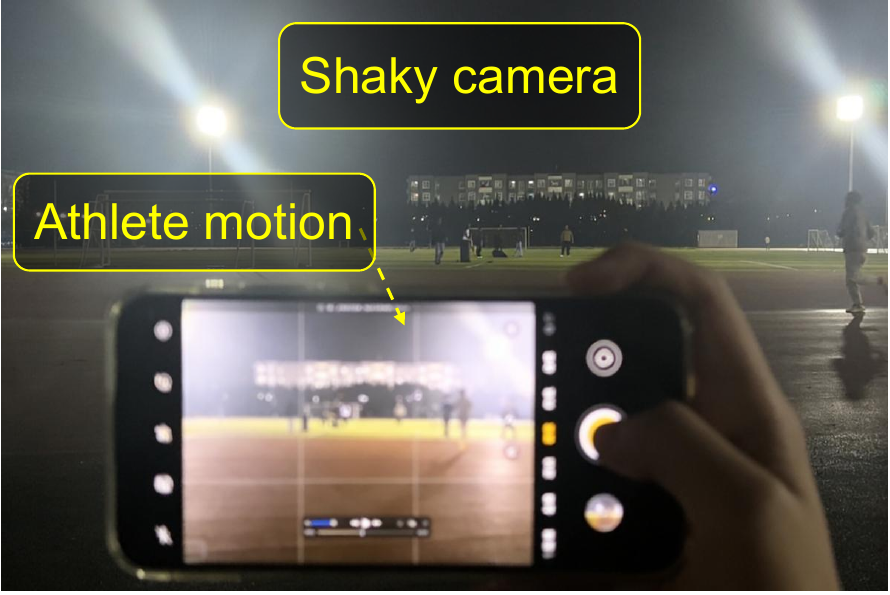}
    }
    \caption{Case study scenarios for real-world low-light videos with non-stationary illumination and motion.}
    \label{fig:scenes}
    \vspace{-3mm}
\end{figure*}

\begin{figure*}[t]
    \centering

    \subfloat[Museum]{
        \begin{minipage}[t]{0.2\textwidth}
            \centering
            \includegraphics[width=\linewidth]{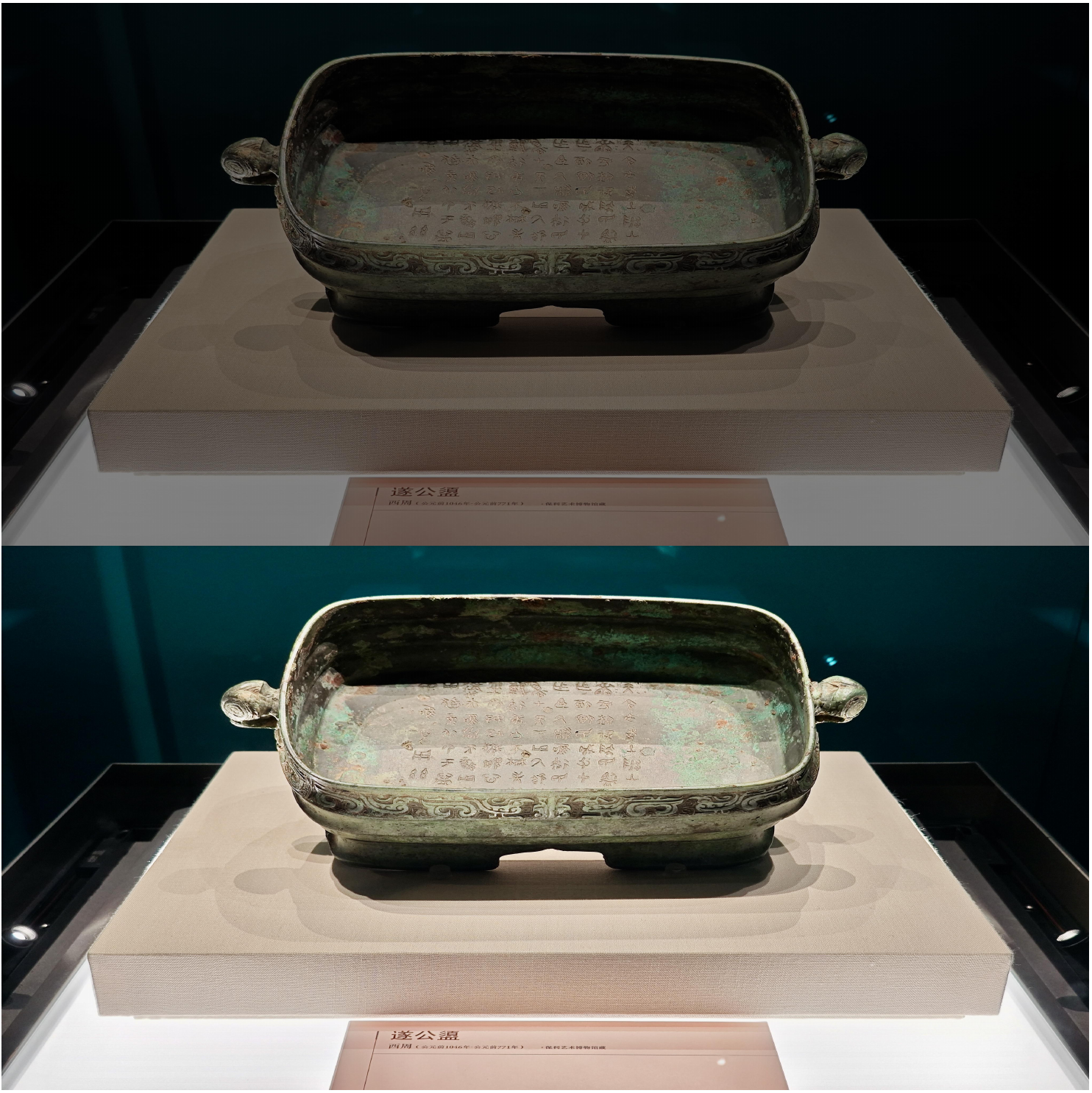}
        \end{minipage}
    }\hfill
    \subfloat[In-vehicle]{
        \begin{minipage}[t]{0.2\textwidth}
            \centering
            \includegraphics[width=\linewidth]{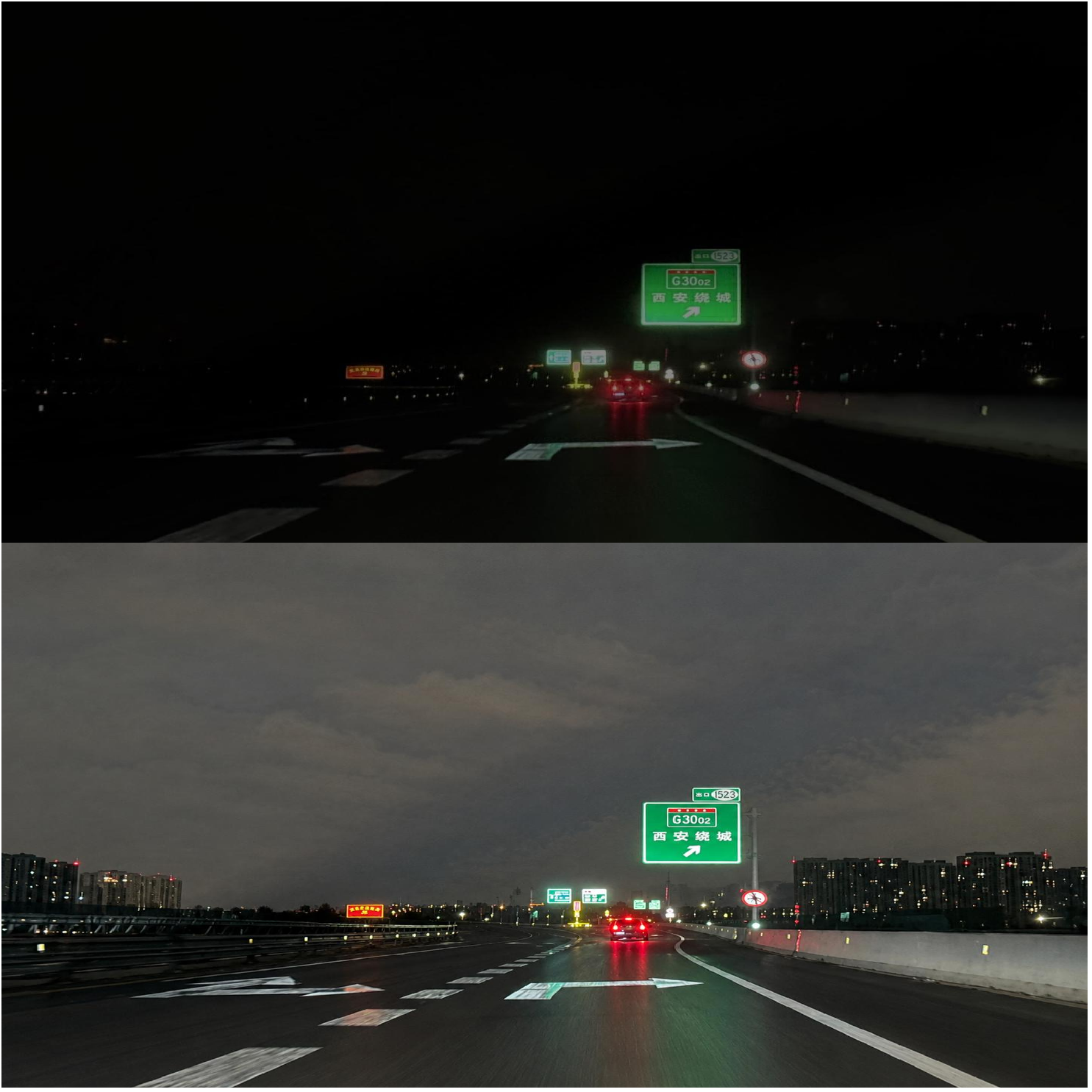}
        \end{minipage}
    }\hfill
    \subfloat[E-bike]{
        \begin{minipage}[t]{0.2\textwidth}
            \centering
            \includegraphics[width=\linewidth]{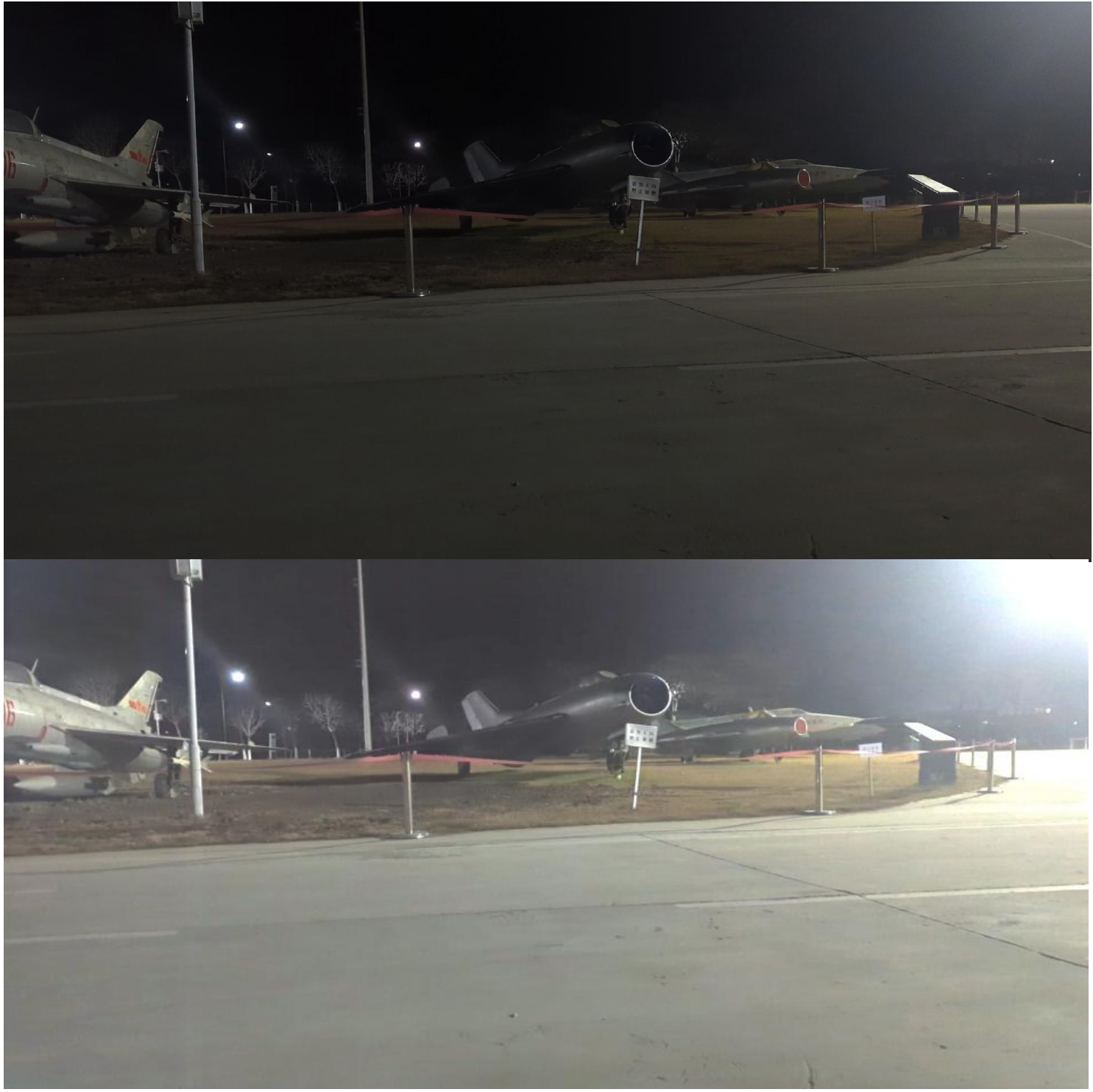}
        \end{minipage}
    }\hfill
    \subfloat[Walk/run]{
        \begin{minipage}[t]{0.2\textwidth}
            \centering
            \includegraphics[width=\linewidth]{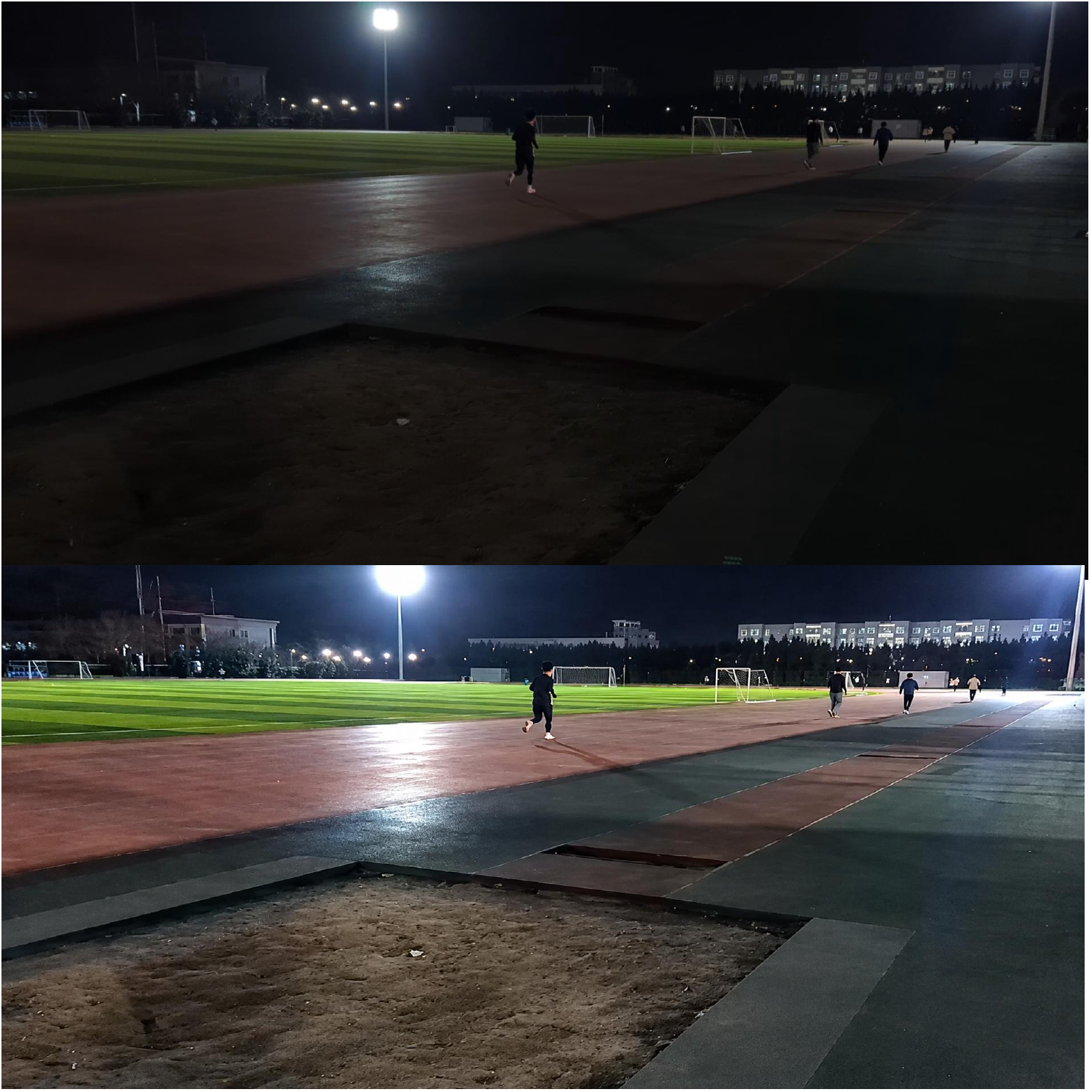}
        \end{minipage}
    }

    \caption{Qualitative enhancement results for the four real-world case-study scenarios in Fig.~\ref{fig:scenes}. Each example compares the low-light  input and the corresponding output, demonstrating improved visibility and structural consistency across museum/smartglass, in-vehicle, night e-bike, and handheld walk/run scenarios.}
    \label{fig:case_results}
    \vspace{-2mm}
\end{figure*}

To stress-test \sysname under real-world low-light videos with mixed, non-stationary illumination and motion, we conduct in-the-wild case studies on mobile and wearable devices. As shown in \figref{fig:scenes}, we curate four representative scenarios with increasing temporal complexity, each recorded as a 2-hour continuous video to evaluate long-term temporal stability and online adaptation. 
Fig.~\ref{fig:case_results} presents qualitative results, showing that \sysname consistently improves dark-region visibility while preserving structural consistency across diverse mobile and wearable low-light scenes. Together, these scenarios cover representative challenges in everyday use, including illumination variation, abrupt scene transitions, and camera motion.
\begin{itemize}
    \item \textbf{Museum/Heritage Smartglass Capture (wearable, perception-sensitive).}
    We capture dim, no-flash exhibits using XREAL One Pro AR glasses with smartphone-assisted enhancement (Xiaomi~17 Ultra), emulating a night-blind user wearing smart glasses. Repeated \emph{stop-pan-stop} motion makes flicker and exposure drift readily perceptible on near-eye displays.
    \item \textbf{Parking Garage Entrance/Exit (abrupt illumination transitions).}
    Using a Pixel~7, we record continuous entry and exit sequences with rapid dark--bright--dark transitions caused by lamps, reflective surfaces, and headlights, challenging exposure consistency.
    \item \textbf{Night E-bike Riding (fast motion + dynamic lighting).}
    A smartphone mounted on an e-bike records night scenes with rapidly changing illumination, vibration, rolling-shutter artifacts, and motion blur, representing a demanding real-time enhancement scenario.
    \item \textbf{Handheld Walking/Running (camera shake).}
    We record handheld outdoor videos under low light, where strong camera shake and stable illumination isolate motion-induced flicker and structural drift.
\end{itemize}

% yundongchangjing, guangzhaoinahua，为什么光照变化，场景变换，为什么会变换呢，骑电动车，光线变化频率大；深海；地下车库，case持续时间的视频；建议几个问题的user study；
% yanjing canguanbowuguan,拍视频，照片

\begin{figure*}[t]
    \centering
    \subfloat[Temporal consistency over time]{
        \includegraphics[width=0.3\textwidth]{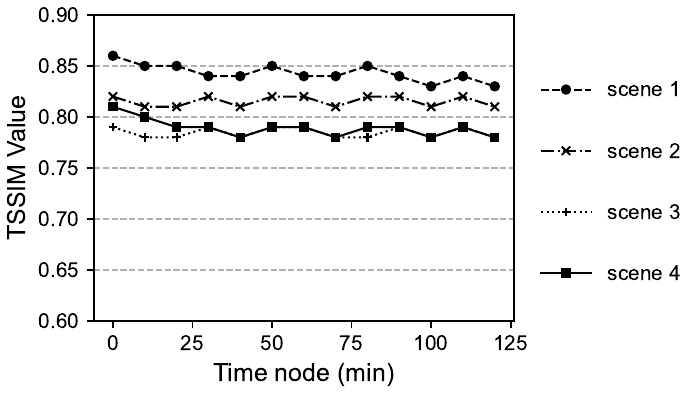}
        \label{fig:tssim_time}
    }
    \hfill
    \subfloat[Perceptual quality over time]{
        \includegraphics[width=0.3\textwidth]{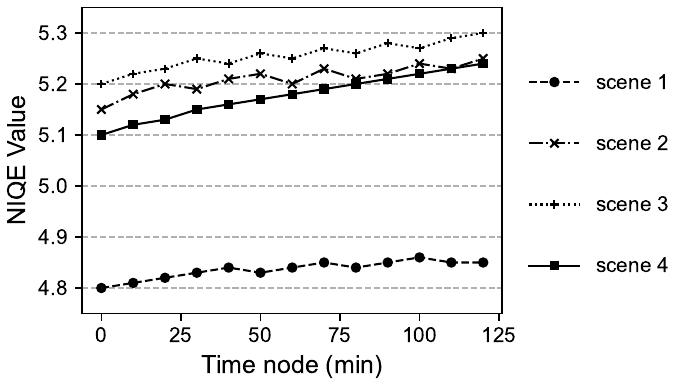}
        \label{fig:niqe_time}
    }
    \hfill
    \subfloat[Per-frame latency over time]{
        \includegraphics[width=0.3\textwidth]{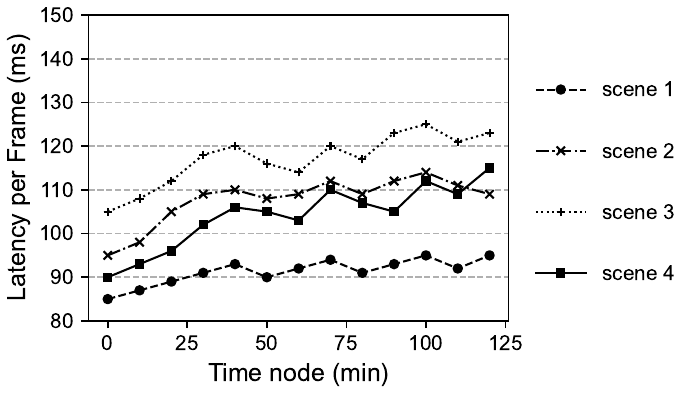}
        \label{fig:frame_latency}
    }
    \caption{Long-stream stability in in-the-wild low-light videos.
    Time-series of TSSIM, NIQE, and per-frame latency over \emph{2-hour} streams across four cases with non-stationary mixed illumination, fast/shaky motion, and extreme low light; the smartglasses case (scene 1) highlights benefits for users \textit{with night-blindness}.}
    \label{fig:userstudy_three}
    \vspace{-6mm}
\end{figure*}

We evaluate time-series trends of temporal consistency (TSSIM), no-reference visual quality (NIQE), and per-frame latency (\figref{fig:userstudy_three}).
\textit{First,} \sysname maintains \emph{consistent visual quality} across diverse dynamics: NIQE stays within a narrow band over time for each scenario (Fig.~\ref{fig:niqe_time}).
\textit{Second,} \sysname achieves \emph{stable temporal consistency} over the full 2-hour streams: TSSIM curves are largely flat with only small fluctuations, indicating suppressed flicker and no long-term drift (Fig.~\ref{fig:tssim_time}).
\textit{Third,} \sysname sustains \emph{bounded real-time latency} without systematic growth, suggesting robust online adaptation that does not destabilize throughput (Fig.~\ref{fig:frame_latency}).
Unlike the controlled battery-budget experiment in Section~5.3.2, this latency trend reflects real in-the-wild runtime variation rather than battery level alone.
Latency may increase in harder segments with fast motion, abrupt illumination changes, or fewer reuse/caching opportunities, where the controller allocates more computation to preserve stability and perceptual quality.
Thus, the trend does not contradict the battery-level experiment; it shows that long-running performance is jointly shaped by scene dynamics, adaptation decisions, and system conditions.
\textbf{Summary.}
Across four 2-hour in-the-wild scenarios, \sysname maintains high visual quality, stable temporal consistency, and bounded real-time latency under mixed illumination, fast motion, shaky motion, and extreme low-light.
The smartglasses case study further suggests practical benefits for users with night blindness.

%% file: body/related.tex
\section{RELATED WORK}

\subsection{User-Centric Mobile Visual Sensing Application}
Smartphones and wearables have become \emph{video-centric} sensing platforms for human-environment interaction, enabled by high-quality cameras, connectivity, and on-device AI.
Prior work mainly spans \textit{media creation} (photography/short video~\cite{widyo2023postmedia,hasinoff2016burstmedia}), \textit{interactive perception} (video communication~\cite{videocommu, karis2016improvingvideo}, AR navigation~\cite{gerstweiler2015AR,ARCore2018learn,applicationVR}), and \textit{safety/assistive sensing} (assistive perception~\cite{assistance2018wearable,assistanceli2022sensing}, mobile robotics~\cite{robot2019fast,robot2023safety}, smart-home monitoring~\cite{home2014vision,home2020vision}).
These applications run over \textit{long streams}, are directly perceived by users, and must meet strict on-device latency/energy budgets.
\sysname targets robust long-duration processing by explicitly modeling perceptual tolerance, temporal stability, and runtime variability.

\subsection{Low-light Image/Video Enhancement}
We group prior work into \textbf{three types} and position \sysname accordingly.
\textbf{First, frame-centric enhancement.}
Videos are processed frame-by-frame using physical priors (\eg Retinex~\cite{1977retinex}, 1977) or learning-based single-image models~\cite{2017llcnn,2017lowcnn,Retinex-Net,zero-Dce,zero-dce++,ExCNet,uretinex,2023fourllie}.
Curve-based methods (Zero-DCE~\cite{zero-Dce} and its accelerated variant Zero-DCE++~\cite{zero-dce++}) estimate image-adaptive exposure curves with (little or) no paired supervision, enabling lightweight deployment.
These methods often deliver strong per-frame fidelity with lightweight models~\cite{zero-dce++,MBLLEN,Review2025}, but independent optimization readily induces flicker/jitter and motion-related artifacts.
\textbf{Second, frame-centric enhancement + post-hoc temporal modeling.}
Frame-centric pipelines are augmented with explicit temporal constraints (\eg flow-based warping, recurrent propagation, or short-term temporal modules) to suppress flicker.
A representative milestone is~\cite{zhang2021learning} (2021), which introduces inter-frame consistency constraints to learn temporal stability for low-light video enhancement even from \emph{single-image} training data.
While such constraints improve coherence with minimal changes to image backbones, they are \emph{fragile} when alignment becomes unreliable (low SNR, blur, fast motion) and incur high latency/energy overhead.
\textbf{Third, video-native spatiotemporal restoration.}
Videos are optimized \emph{directly} with video-centric data, models, and objectives, evolving from
\textit{(III-A)} paired dynamic-video supervision~\cite{smoid, SDSD}, to \textit{(III-B)} pushing stronger spatiotemporal guidance~\cite{DID}, and further to
\textit{(III-C)} unpaired/generative learning~\cite{zhu2024unrolled}.
Diffusion models~\cite{2024wavedmdiff,2023diffretinex,2024fourierdiff,2023exposurediffusion} belong to \textit{Stage III-C} (generative priors): strong perceptual quality, but typically too heavy for real-time on-device use due to iterative sampling.
By exploiting cross-frame redundancy, video-native methods make temporal stability largely an \emph{endogenous} property; however, III-C must still balance consistency with efficiency for real-time deployment.
This taxonomy clarifies two key questions:
\begin{itemize}
    \item \textit{Q1: why video-native beats frame-wise + post-hoc consistency.}
    Low-light degradation and motion/illumination changes are \emph{spatiotemporally coupled}, so enforcing consistency \emph{after} independent per-frame restoration often leaves residual flicker and motion artifacts.
    Video-native methods restore jointly across frames, turning consistency from an external constraint into an \emph{endogenous} outcome.
    \item \textit{Q2: our leap within Stage III-C.}
    We move beyond reliance on paired dynamic-video supervision (costly, narrow coverage, and weaker generalization under dynamic lighting/fast motion) toward \emph{unpaired/generative} video-native learning that still yields \emph{stable} spatiotemporal enhancement in real scenes, while remaining \emph{resource-aware} on ubiquitous devices.
\end{itemize}

\subsection{Adaptive Vision Tasks on Mobile Devices}
Mobile vision increasingly adopts \textit{adaptive computation} to balance perceptual quality with latency and energy under dynamic runtime conditions.
Prior work mainly falls into two lines.
\textit{Architectural adaptivity} (\eg early-exit~\cite{early2017deepsense,early2016branchynet}, MoE~\cite{moe2022switch,yi2025edgemoe}) skips computation via conditional execution, but is often \textit{runtime-agnostic} (battery/thermals/contention), limiting long-duration mobile use.
\textit{System-level adaptivity} (\eg scheduling~\cite{ye2025schedule}, content-aware control~\cite{chin2019adascale}) adjusts runtime execution, but typically treats models as black boxes and lacks \textit{perceptual-quality control}.
\sysname closes this gap by coupling \textit{human perceptual tolerance} with \textit{runtime feedback} to scale enhancement cost and sustain always-on mobile video processing.

%% file: body/conclud.tex
% \section{Conclusion}
% We present \sysname, an on-device real-time low-light video enhancement system that makes low-light capture \textit{stable and usable}, not merely aesthetic—under tight compute and energy budgets, benefiting safety- and accessibility-critical scenarios.
% \sysname couples a spectrum-guided diffusion transformer that corrects illumination while anchoring structure/identity for reduced flicker, with a user perception-aware, resource-aware controller that adapts sampling depth, attention reuse, and caching online.
% Implemented end-to-end on 手机 and 智能眼镜, \sysname achieves consistently superior temporal stability and up to 99.68\% latency reduction over strong baselines, enabling practical real-time 视频流 增强 on 动态混杂照度与用户运动，包括快速移动、极端低照度. 这对普通使用和夜盲症用户都很重要。
% % Limitations include heuristic control signals that may not fully reflect individual preferences.
% Future work will explore learned scheduling policies grounded in user perception, and more efficient diffusion mechanisms for long-duration, energy-constrained deployment.

\section{CONCLUSION}
We presented \sysname, a real-time on-device low-light video enhancement system that makes diffusion-based enhancement practical for mobile interactive vision.
Rather than pursuing visual quality alone, \sysname jointly optimizes enhancement fidelity, temporal stability, and interactive responsiveness through a perception-aware and resource-aware design.
By tightly coupling diffusion inference with runtime resource adaptation, \sysname enables stable real-time enhancement under highly dynamic illumination and user motion on commodity smartphones and smartglasses.
More broadly, this work demonstrates that diffusion models can become practical building blocks for mobile interactive vision when algorithm design is co-optimized with user perception and system constraints.
More effort and insight are needed in perception-driven scheduling, adaptive generative inference, and resource-aware foundation models for always-on mobile intelligence.